\documentclass[10pt,twocolumn,letterpaper]{article}

\usepackage[pagenumbers]{wacv}

\usepackage{iftex}
\ifPDFTeX\usepackage[accsupp]{axessibility}\fi

\usepackage{microtype}

\usepackage{mathtools}
\usepackage{amsthm}
\usepackage{bm}

\usepackage{float}        %
\usepackage{dblfloatfix}  %
\usepackage{array}
\usepackage{multirow}
\usepackage{makecell}
\usepackage{pifont}
\usepackage{siunitx}
\usepackage{rotating}
\usepackage{tabularx}
\usepackage{setspace}
\usepackage[export]{adjustbox}

\theoremstyle{plain}

\theoremstyle{definition}

\theoremstyle{remark}

\usepackage{xr-hyper}

\definecolor{wacvblue}{rgb}{0.21,0.49,0.74}
\usepackage[pagebackref,breaklinks,colorlinks,allcolors=wacvblue]{hyperref}

\def\wacvPaperID{974} %
\def\confName{WACV}
\def\confYear{2027}

\title{Image Deblending with Diffusion Models}

\author{Dan Botchan \qquad Ohad Fried \qquad Yacov Hel-Or\\
Reichman University, Herzliya, Israel\\
{\tt\small dan.botchan@post.runi.ac.il}
}

\definecolor{darkred}{rgb}{0.7,0.1,0.1}
\definecolor{darkgreen}{rgb}{0.1,0.6,0.1}
\definecolor{cyan}{rgb}{0.7,0.0,0.7}
\definecolor{otherblue}{rgb}{0.1,0.4,0.8}
\definecolor{maroon}{rgb}{0.76,.13,.28}
\definecolor{burntorange}{rgb}{0.81,.33,0}

\newif\ifdraft
\draftfalse

\ifdraft
  \newcommand{\dan}[1]{{\color{magenta}[\textbf{Dan:} #1]}}
  \newcommand{\ohad}[1]{{\color{burntorange}[\textbf{Ohad:} #1]}}
  \newcommand{\toky}[1]{{\color{cyan}[\textbf{Toky:} #1]}}

\else
  \newcommand{\dan}[1]{}
  \newcommand{\ohad}[1]{}
  \newcommand{\toky}[1]{}

\fi

\newcommand{\suppcref}[1]{\cref{#1} of the supplementary material}

\newif\ifincludesupp \includesupptrue

\ifincludesupp\else
\fi

\begin{document}
\maketitle
{\let\thefootnote\relax\footnotetext{\scriptsize \copyright\ 2027 IEEE. Personal use of this material is permitted. Permission from IEEE must be obtained for all other uses, in any current or future media, including reprinting/republishing this material for advertising or promotional purposes, creating new collective works, for resale or redistribution to servers or lists, or reuse of any copyrighted component of this work in other works. Accepted to the 2027 IEEE/CVF Winter Conference on Applications of Computer Vision (WACV).}}

\begin{abstract}
Image deblending is the task of decomposing a composite image back into its source images, which is an ill-posed problem, as a single blended image can be decomposed into multiple plausible source layers. While diffusion models are often used to solve inverse problems, they typically address single-output reconstruction problems. In this work, we formulate deblending as a joint inverse problem, employing diffusion posterior sampling to simultaneously reconstruct two source images, demonstrating that a single diffusion prior is expressive enough to disentangle two full images at once rather than just one. Drawing inspiration from human perception, where semantic understanding helps disentangle visual inconsistencies, we propose a training-free framework that leverages text prompts to constrain this solution space. We address the blind deblending task through a two-stage architecture: text-guided latent diffusion performs semantic separation using prompts as high-level cues, followed by a pixel-space refinement to recover fine details. Stochastic sampling produces multiple plausible decompositions that can be ranked without ground truth. Extensive experiments show that our training-free method achieves state-of-the-art results on unseen domains and extends to spatially-varying alpha maps, while remaining competitive with supervised baselines on standard scalar blending benchmarks.

\end{abstract}

\section{Introduction}
\label{sec:introduction}

A widely used technique for image composition in digital editing is alpha-blending, which combines multiple images using spatial weight coefficient maps, also known as alpha maps, that determine the contribution of each source image at every pixel. For two color images, it is formulated as:

\begin{equation}
\label{eq:alpha_blending}
y=\alpha\odot I_{1}+(1-\alpha)\odot I_{2}
\end{equation} 

where $y$ denotes the blended image, $I_{1}, I_{2}\in\mathbb{R}^{H\times W\times3}$ are two RGB images, $\alpha\in[0,1]^{H\times W}$ is the alpha map and $\odot$ represents pixel-wise multiplication. While forming $y$ is straightforward, the inverse problem of recovering $I_{1}, I_{2}$, and $\alpha$, which we refer to as \textit{alpha-deblending}, requires recovering three unknowns from a single observation, which is fundamentally ill-posed.

\begin{figure}[t]
    \centering
    \scriptsize
    \makebox[0.24\columnwidth][c]{\textbf{Source 1}}\hfill
    \makebox[0.24\columnwidth][c]{\textbf{Source 2}}\hfill
    \makebox[0.24\columnwidth][c]{\textbf{$\alpha$ map}}\hfill
    \makebox[0.24\columnwidth][c]{\textbf{Blended input}}\\[-1pt]
    \begin{subfigure}[b]{0.24\columnwidth}
        \centering
        \includegraphics[frame,width=0.98\linewidth]{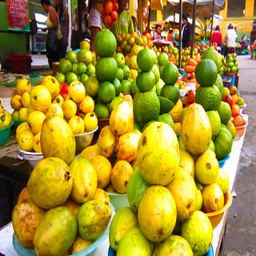}
        \caption{}
    \end{subfigure}\hfill
    \begin{subfigure}[b]{0.24\columnwidth}
        \centering
        \includegraphics[frame,width=0.98\linewidth]{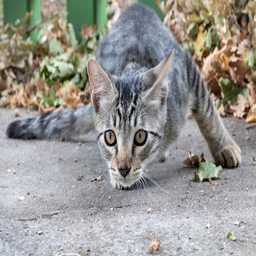}
        \caption{}
    \end{subfigure}\hfill
    \begin{subfigure}[b]{0.24\columnwidth}
        \centering
        \includegraphics[frame,width=0.98\linewidth]{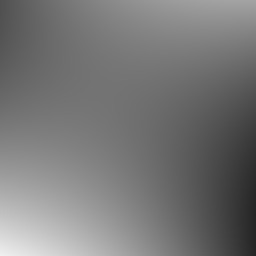}
        \caption{}
    \end{subfigure}\hfill
    \begin{subfigure}[b]{0.24\columnwidth}
        \centering
        \includegraphics[frame,width=0.98\linewidth]{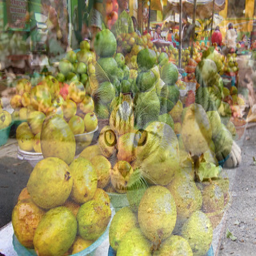}
        \caption{}
    \end{subfigure}\\[0.15cm]
    
    \raisebox{2pt}{\makebox[0.24\columnwidth][c]{\textbf{Recovered 1}}}\hfill
    \raisebox{2pt}{\makebox[0.24\columnwidth][c]{\textbf{Recovered 2}}}\hfill
    \raisebox{2pt}{\makebox[0.24\columnwidth][c]{\textbf{Recovered $\alpha$}}}\hfill
    \raisebox{2pt}{\makebox[0.24\columnwidth][c]{\textbf{Reconstruction}}}\\[-1pt]
    \begin{subfigure}[b]{0.24\columnwidth}
        \centering
        \includegraphics[frame,width=0.98\linewidth]{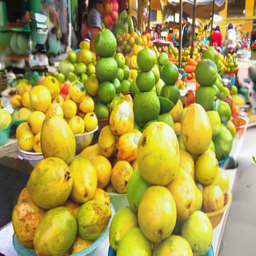}
        \caption{}
    \end{subfigure}\hfill
    \begin{subfigure}[b]{0.24\columnwidth}
        \centering
        \includegraphics[frame,width=0.98\linewidth]{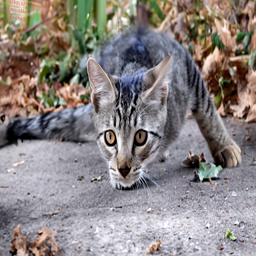}
        \caption{}
    \end{subfigure}\hfill
    \begin{subfigure}[b]{0.24\columnwidth}
        \centering
        \includegraphics[frame,width=0.98\linewidth]{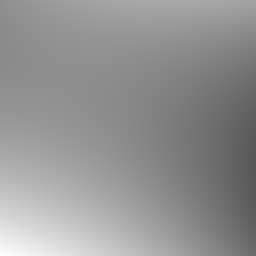}
        \caption{}
    \end{subfigure}\hfill
    \begin{subfigure}[b]{0.24\columnwidth}
        \centering
        \includegraphics[frame,width=0.98\linewidth]{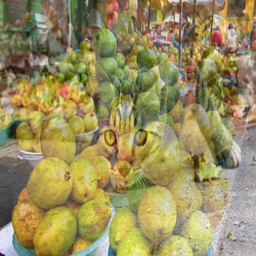}
        \caption{}
    \end{subfigure}
    
    \normalsize
    \caption{\textbf{A single blended image is separated into two sources and a spatially varying $\alpha$-map, without training.}
    Top: (a,b) ground-truth sources, (c) ground-truth $\alpha$-map, (d) blended input $y$.
    Bottom: our recovered sources (e,f), recovered $\alpha$-map (g), and the re-blended reconstruction $\tilde{y}$ (h).
    }
    {\scriptsize{Prompts: ``\textit{Lemons on the shelf in the supermarket}'', ``\textit{A tiger facing the camera.''}}
    }
    \label{fig:teaser_fig}
\end{figure}

Despite this mathematical ambiguity, humans can often identify the constituent images from inconsistencies in perspective, object boundaries, lighting, or semantic context. We use semantic cues to mirror this process, supplied either through lightweight user descriptions or automatically from the blend by a vision--language model, and treat text guidance as a substitute for the task-specific supervision other methods require. Early approaches sought to encode such priors explicitly through mathematical models \citep{farid1999separating,levin2002learning, sarel2004separating, levin2007user} while modern supervised methods learn them implicitly from data \cite{zou2020DAD, han2022BIDeN, su2024compositional, hong2024differ, montenegro2025LCD}.

In parallel, a different line of work has demonstrated that pretrained diffusion models can be adapted to solve inverse problems through posterior sampling, achieving strong results on tasks such as deblurring, super-resolution, inpainting, compressed MRI, and spectral CT decomposition \cite{choi2021ilvr, chung2022come, chung2022dps, chung2022mcg,dontas2024blind,jiang2025specPDS}. This training-free paradigm is appealing for image decomposition for two reasons: it generalizes across image domains without task-specific training data, and text-conditioned diffusion models allow human semantic guidance to be injected directly through prompts \cite{dontas2024blind}. 

To the best of our knowledge, existing image-domain training-free diffusion methods focus on single-output inverse problems and have not been adapted to multi-output alpha-deblending, where multiple source images must be recovered from a single observation.

We extend the posterior diffusion sampling paradigm to simultaneously denoise multiple images while enforcing consistency with the blending equation (\cref{fig:teaser_fig}), showing that a single pretrained diffusion prior is expressive enough to disentangle two full images jointly, rather than being limited to reconstructing one output at a time as in prior work.

We evaluate three settings: known scalar $\alpha$, unknown scalar $\alpha$, and unknown spatially varying $\alpha$. Without task-specific training, our method achieves state-of-the-art results on unseen domains, the best SSIM and LPIPS on spatially varying blends, and remains competitive with supervised baselines on known-scalar benchmarks.

Moreover, unlike deterministic approaches that commit to a single output, our stochastic sampler generates multiple plausible decompositions: when one sample fails, resampling offers another chance at a correct solution.

Our main contributions are:
\begin{itemize}
    \item We introduce a training-free framework for blind alpha-deblending, jointly recovering source images and alpha maps from a single blended observation.
    \item We propose a two-stage process combining text-guided latent diffusion with pixel-space refinement to leverage both semantic priors and fine-grained detail recovery.
    \item We introduce a ground-truth-free candidate-selection strategy and compare PickScore with CLIP similarity for selecting high-quality candidates among multiple generations.
\end{itemize}

{\dimen0=\parindent\raggedright\parindent=\dimen0\Urlmuskip=0mu\relax\def\UrlBreaks{\do\/}\def\UrlBigBreaks{}\def\UrlNoBreaks{\do\-} Code is available at~\url{https://github.com/DanBotchan/image-deblending-diffusion}.\par}

\section{Related Work}
\label{sec:related_work}

The task of separating a composite image into its constituent source images is a fundamental problem in computer vision and image processing that manifests across domains including reflection separation \citep{levin2004separating, farid1999separating, zhang2018singleimagereflectionseparation} and layer separation \citep{li2014single, su2024compositional, zou2020DAD}. The field began with approaches requiring multiple captures of the same scene \citep{farid1999separating, sarel2004separating} or relying on hand-crafted priors such as sparsity constraints \citep{levin2007user} and smoothness assumptions \citep{li2014single}.

As deep learning advanced, the Double-DIP method \citep{gandelsman2019doubleDIP} extended the Deep Image Prior \citep{ulyanov2018DIP} framework for training-free decomposition through per-image optimization of coupled networks. Deep Adversarial Decomposition (DAD) \citep{zou2020DAD} introduced adversarial losses to ensure realistic and independent layers. This approach was subsequently extended with GIP \citep{zou2021GIP}, which shifted focus to learning degradation parameters rather than directly predicting clean images. Building on these ideas, the Blind Image Decomposition Network (BIDeN) \citep{han2022BIDeN} generalized the framework to handle decomposition into $N$ sources using multi-scale encoders with $N$ decoder heads. BIDeN also introduced comprehensive benchmark datasets across multiple decomposition tasks which we adopt for evaluation. Supervised methods achieve strong in-distribution performance; however, they require domain-specific training and assume spatially-constant blending coefficients.

Diffusion models \citep{dickestein2015deepunsupervised, ho2020DDPM} and score-based generative models \citep{song2019generative, song2021score} have also been applied to image decomposition. Decomp Diffusion \citep{su2024compositional} decomposes scenes into compositional concepts and L-DiffER \citep{hong2024differ} introduced text-guided reflection removal. Both approaches require training diffusion models specifically for their respective decomposition tasks. Closer to our setting, \citet{montenegro2025LCD} adapted Cold Diffusion \citep{bansal2023cold} to decompose superimposed images. However, this approach is limited to scenarios where the $\alpha$ map is known in advance and spatially constant. DiffDecompose \citep{wang2025diffdecompose} trains a diffusion Transformer conditioned on the observed image, semantic prompts, and blending type. In contrast, our method uses a frozen text-to-image prior without decomposition-specific training and jointly recovers a blind, spatially varying $\alpha$ map.

\textbf{Diffusion Posterior Sampling for Inverse Problems}
Diffusion models have been successfully applied as generative priors for training-free inverse problem solving through posterior sampling (\cref{sec:preliminaries}) \citep{chung2022mcg, chung2022dps, chung2022come, bansal2023universal}, in both pixel and latent space \citep{rout2023solving, bansal2023universal, dontas2024blind}. These methods achieve strong results on tasks such as super-resolution, inpainting, and deblurring. However, image-domain formulations generally focus on single-output problems, recovering a single unknown image from one observation. Outside image deblending, multi-source diffusion separation has been studied in audio and signal processing using a learned joint multi-source prior \citep{mariani2024multisource} or separately learned source priors in a training-free posterior sampler \citep{zhang2025bayesian}.

Posterior sampling has also been extended to blind inverse problems, where degradation parameters are unknown and must be jointly optimized alongside image reconstruction \citep{dontas2024blind}, but these methods still formulate the output as one target image rather than multiple source images. In this work, we adapt alternating optimization to jointly recover multiple source images and spatially-varying alpha maps from a single blended observation.

Text-to-image generation commonly produces several plausible candidates, motivating automatic ranking criteria. PickScore \citep{kirstain2023pickapic} is trained from pairwise user preferences for text-to-image generations. Although it was not trained for deblending, we use it as a ground-truth-free selector and compare it with CLIP similarity in \suppcref{appendix:selector_ablation}.

\section{Preliminaries}
\label{sec:preliminaries}
We first outline diffusion posterior sampling for general single-output inverse problems. We then extend this framework in \cref{subsec:method_joint_source_sampling} to jointly recover multiple source images.

In a single-output blind inverse problem, we aim to recover an unknown signal $x$ from a measurement $y$ when either the measurement operator $A_\phi$ itself or its parameters $\phi$ are unknown. The forward process is described as:
\begin{equation}
\label{eq:blind_inverse_problem}
y = A_\phi(x) + n,
\end{equation}
where $n$ denotes additive noise. Inverse problems are often fundamentally ill-posed due to information loss from both $A_\phi$ and $n$, meaning that multiple distinct signals $x$ can yield identical measurements $y$.

To resolve this ill-posedness in image reconstruction inverse problems, we leverage generative models as learned priors that constrain the solution space to natural images. Diffusion models \citep{ho2020DDPM} and score-based models \citep{song2021score} provide such a prior by capturing the image distribution through the learned score function $\nabla_{x_t}\log p(x_t)$, where $x_t$ denotes a noisy version of the image at diffusion timestep $t$, with $x_0 = x$ representing the clean image (\suppcref{eq:forward_diffusion}). In practice, the score is given by a noise-prediction network $\epsilon_\theta$, with $\nabla_{x_t}\log p(x_t) \approx -\epsilon_\theta(x_t,t)/\sqrt{1-\bar{a}_t}$. We can leverage the conditional score function $\nabla_{x_t}\log p(x_t|y)$ \citep{chung2022come, chung2022dps, chung2022mcg} to sample from the posterior distribution. Using Bayes' rule, the conditional score can be decomposed as:
\begin{equation}
\label{eq:conditional_score_function}
\nabla_{x_t}\log p(x_t|y) = \nabla_{x_t}\log p(y|x_t) + \nabla_{x_t}\log p(x_t),
\end{equation}
where $\nabla_{x_t}\log p(y)$ is omitted since $y$ is constant with respect to $x_t$.
This separates the conditional score into two components: the prior score $\nabla_{x_t}\log p(x_t)$, provided by the pretrained diffusion model, and the log-likelihood gradient $\nabla_{x_t}\log p(y|x_t)$, which favors estimates that reproduce the observation $y$ under the forward model $A_\phi$. This agreement between the predicted and observed measurements is the data-consistency condition. More generally, the prior score can be conditioned on auxiliary information $c$, such as a text prompt, yielding $\nabla_{x_t}\log p(x_t|c)$, which is directly available from pretrained text-to-image diffusion models. For readability, we omit this conditioning from the equations throughout this section and simply write $\nabla_{x_t}\log p(x_t)$, with the understanding that it can be replaced by $\nabla_{x_t}\log p(x_t|c)$ wherever text guidance is used.

The main challenge lies in computing the likelihood gradient $\nabla_{x_t}\log p(y|x_t)$. During reverse diffusion, we only have access to noisy samples $x_t$, but the measurement $y$ was generated from a clean image $x_0$ via the forward process in \cref{eq:blind_inverse_problem}. Computing $p(y|x_t)$ would require marginalizing over all possible clean images consistent with $x_t$: $p(y|x_t) = \int p(y|x_0)p(x_0|x_t)dx_0$, which is intractable. 

To address this, \citet{chung2022dps} propose approximating $p(y|x_t)$ using an estimate of the clean image. Specifically, they estimate $\hat{x}_0$ from the noisy sample $x_t$ using Tweedie's formula \citep{efron2011tweedie}:
\begin{equation}
\label{eq:tweedies_formula}
\hat{x}_0 := \mathbb{E}[x_0|x_t] = \frac{1}{\sqrt{\bar{a}_t}}(x_t + (1-\bar{a}_t)\nabla_{x_t}\log p(x_t)),
\end{equation}
where $\bar{a}_t$ is the cumulative noise coefficient at timestep $t$. By approximating $p(y|x_t) \approx p(y|\hat{x}_0)$, we obtain a tractable form. Under the Gaussian measurement noise assumption from \cref{eq:blind_inverse_problem}, the likelihood becomes:
\begin{equation}
p(y|\hat{x}_0) \propto \exp\left(-\frac{1}{2\sigma_y^2}\|y - A_\phi(\hat{x}_0)\|^2\right),
\end{equation}
yielding the likelihood gradient:
\begin{equation}
\label{eq:likelihood_approx}
\nabla_{x_t}\log p(y|x_t) \approx -\lambda \cdot \nabla_{x_t} \|y - A_\phi(\hat{x}_0)\|^2,
\end{equation}
where $\lambda$ is a scaling factor controlling the data-consistency strength. The resulting gradient reduces the discrepancy between the predicted measurement $A_\phi(\hat{x}_0)$ and the observation $y$.

Using this conditional score in the reverse diffusion process combines the learned image prior with measurement consistency. Different initial noise samples $x_T \sim \mathcal{N}(0,I)$ produce distinct posterior samples, which we later exploit to generate multiple plausible decompositions. We next extend this framework to jointly recover two source images and a spatially varying alpha map.

\section{Method}
\label{sec:method}
We present a training-free framework that leverages pretrained diffusion models to solve the blind alpha-deblending problem (\cref{subsec:method_problem_formulation}). Our approach extends posterior sampling to jointly recover multiple source images through simultaneous denoising (\cref{subsec:method_joint_source_sampling}), constrained by a polynomial parameterization of the alpha map that ensures smooth spatial blending patterns (\cref{subsec:method_alpha_map_parameterization}). We employ a two-stage pipeline (\cref{subsec:method_two_stage_approach}): Stage 1 performs text-guided semantic separation in latent space, using text as an explicit, controllable substitute for the implicit prior supervised methods learn from labeled data, while Stage 2 refines the results in pixel space to enhance fine-grained details. Throughout, we exploit the stochasticity of the initial noise sample to generate multiple plausible decompositions, offering a practical advantage over deterministic methods that provide no recourse when a single output fails to separate the sources correctly.
\begin{figure*}[t]
  \centering
  \includegraphics[width=\linewidth]{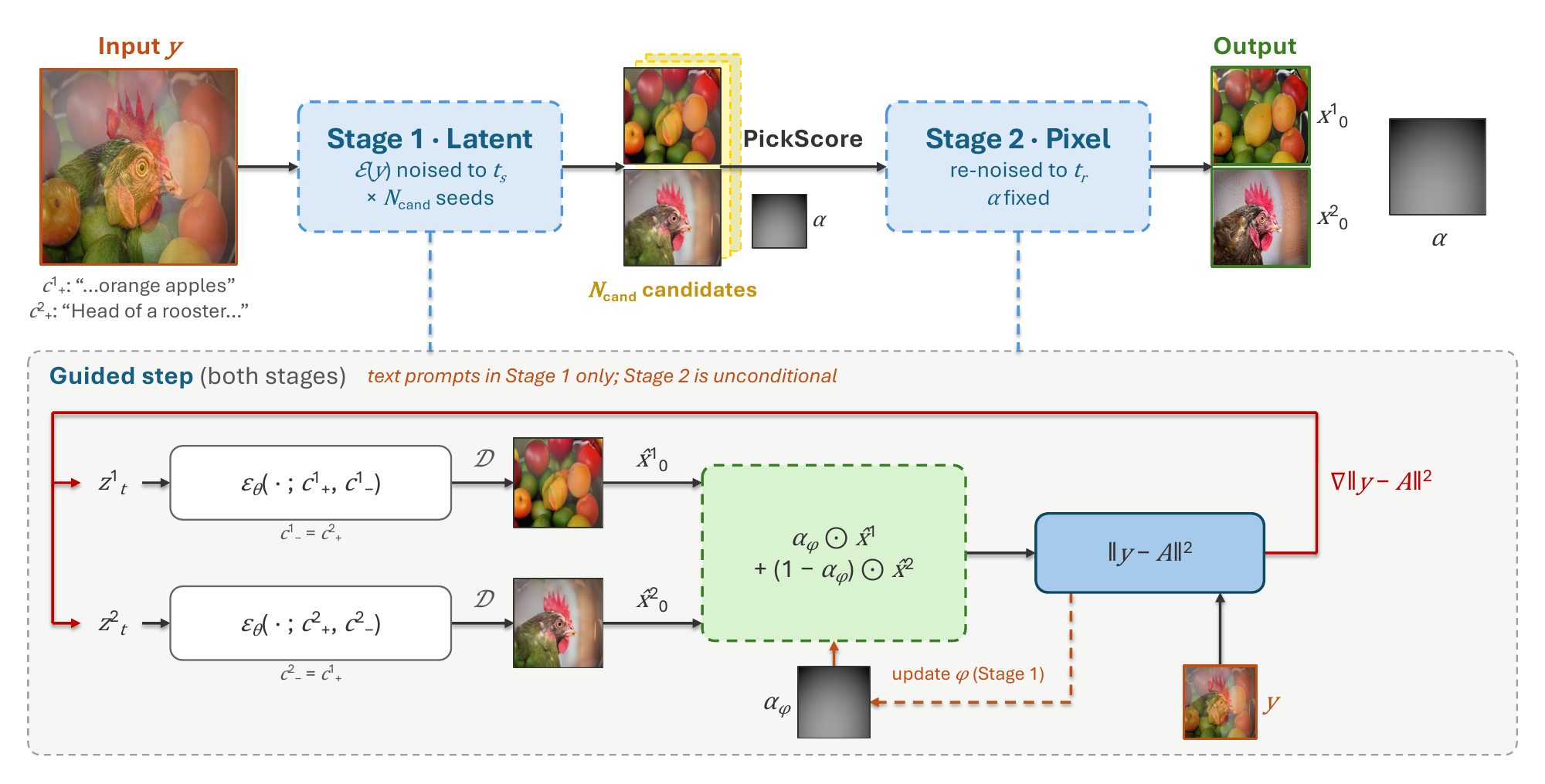}
    \caption{\textbf{Overview of our two-stage training-free blind deblending framework.} \emph{Top:} Stage~1 separates the blended image $y$ in latent space, starting from $\mathcal{E}(y)$ noised to $t_s$ and guided by prompts $c^1_+, c^2_+$. It runs from $N_{\mathrm{cand}}$ seeds, and PickScore selects one candidate pair by scoring each output against its prompt. Stage~2 re-noises that pair to $t_r$ and refines it in pixel space without prompts, keeping $\alpha$ fixed. \emph{Bottom:} the guided step shared by both stages. Each source is denoised by a pretrained model and decoded ($\mathcal{D}$; identity in Stage~2). In Stage~1 the model is conditioned on cross-negative prompts ($c^1_- = c^2_+$, $c^2_- = c^1_+$). The estimates are re-blended with the polynomial $\alpha_\phi$, and the gradient of $\Vert y-A\Vert^2$ updates both sources; in Stage~1, $\phi$ is also updated every $k$ steps. Prompts: ``Close-up image of green, red and orange apples'' and ``Head of a rooster, side view''.}
  \label{fig:method_overview}
\end{figure*}

\subsection{Problem Formulation}
\label{subsec:method_problem_formulation}
Given a blended observation $y \in \mathbb{R}^{H \times W \times 3}$, we seek to recover both source images $x^1_0, x^2_0 \in \mathbb{R}^{H \times W \times 3}$ and the spatially-varying alpha map $\alpha \in [0, 1]^{H \times W}$ that combined them. This extends the single-output blind inverse problem in \cref{eq:blind_inverse_problem}, with unknown parameter $\phi = \alpha$, to a setting with two unknown source images recovered jointly from a single observation. The alpha blending forward process is
\begin{equation}
\label{eq:blending_forward}
y = A(x^1_0, x^2_0, \alpha) = \alpha \odot x^1_0 + (1 - \alpha) \odot x^2_0,
\end{equation}
where $\odot$ denotes element-wise multiplication. The additive noise term is omitted because the blending operation is deterministic and noiseless.

\subsection{Joint Source Sampling}
\label{subsec:method_joint_source_sampling}
We perform joint inference over both sources by sampling from the posterior distribution $p(x_t^1, x_t^2 \mid y)$, where $x_t^1, x_t^2$ represent the source images at diffusion timestep $t$. Following the framework in \cref{sec:preliminaries}, we decompose the joint conditional score using Bayes' theorem:
\begin{multline}
\label{eq:joint_score_function}
\nabla_{x_t^1,x_t^2}\log p(x_t^1, x_t^2 \mid y) = \\
\nabla_{x_t^1,x_t^2} \log p(y \mid x_t^1, x_t^2) + \nabla_{x_t^1,x_t^2} \log p(x_t^1, x_t^2),
\end{multline}
Here, $\nabla_{x_t^1,x_t^2} f := (\nabla_{x_t^1}f,\nabla_{x_t^2}f)^\top$ stacks the gradients with respect to the two separate image variables; it does not require a network that takes two images as input. We evaluate the same pretrained single-image diffusion model independently for each source, while the sources are coupled through the blending likelihood $p(y\mid x_t^1,x_t^2)$. In \cref{eq:joint_score_function}, the first term is the likelihood score and the second is the prior score.
We assume source images to be independent: $p(x_t^1, x_t^2) = p(x_t^1)p(x_t^2)$, since the two sources are independently generated. Under this assumption, the joint prior decomposes as:
\begin{equation}
\label{eq:dual_source_independent_prior}
\nabla_{x_t^1, x_t^2} \log p(x_t^1, x_t^2) = \begin{pmatrix} \nabla_{x_t^1} \log p(x_t^1) \\ \nabla_{x_t^2} \log p(x_t^2) \end{pmatrix},
\end{equation}
enabling independent score computation for each source.
For the likelihood term, we apply the measurement consistency constraint. Following the approximation in \cref{eq:likelihood_approx}, with the constant $\lambda$ replaced by a time-dependent schedule $s(t)$, the likelihood gradient is:
\begin{multline}
\label{eq:likelihood_score_function}
\nabla_{x_t^1,x_t^2} \log p(y|x_t^1,x_t^2) = \\
-s(t) \nabla_{x_t^1,x_t^2} \|y - A(\hat{x}_0^1, \hat{x}_0^2, \alpha)\|^2,
\end{multline}
where $\hat{x}_0^1, \hat{x}_0^2$ are the Tweedie estimates from $x_t^1, x_t^2$, and $s(t) > 0$ is a time-dependent weighting schedule that controls the gradient strength (detailed in \suppcref{appendix:implementation_details}). Because the blending residual is a differentiable scalar function of both source estimates, backpropagation through $A(\hat{x}_0^1,\hat{x}_0^2,\alpha)$ directly yields one likelihood gradient for each source.
Combining these components, the complete joint conditional score becomes:
\begin{multline}
\label{eq:complete_joint_score}
\nabla_{x_t^1, x_t^2}\log p(x_t^1, x_t^2 | y) = \\\begin{pmatrix} \nabla_{x_t^1} \log p(x_t^1) + \nabla_{x_t^1} \log p(y | x_t^1, x_t^2) \\ \nabla_{x_t^2} \log p(x_t^2) + \nabla_{x_t^2} \log p(y | x_t^1, x_t^2) \end{pmatrix}
\end{multline}
This decomposition enables joint inference through two parallel reverse diffusion processes: each source is guided by its own prior score, while the shared likelihood term couples them so that their blend matches $y$.

\subsection{Alpha Map Reparameterization}
\label{subsec:method_alpha_map_parameterization}
\begin{figure}[t]
    \centering
    \captionsetup[subfigure]{aboveskip=1pt}
    \begin{subfigure}[b]{0.24\columnwidth}
        \centering
        \includegraphics[frame,width=0.98\linewidth]{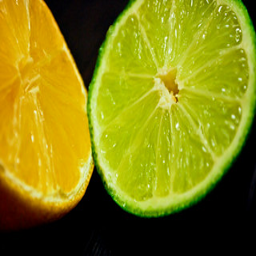}
        \caption{}
    \end{subfigure}\hfill
    \begin{subfigure}[b]{0.24\columnwidth}
        \centering
        \includegraphics[frame,width=0.98\linewidth]{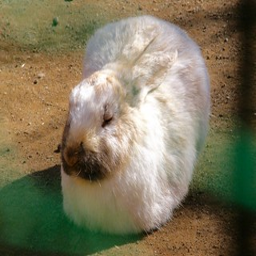}
        \caption{}
    \end{subfigure}\hfill
    \begin{subfigure}[b]{0.24\columnwidth}
        \centering
        \includegraphics[frame,width=0.98\linewidth]{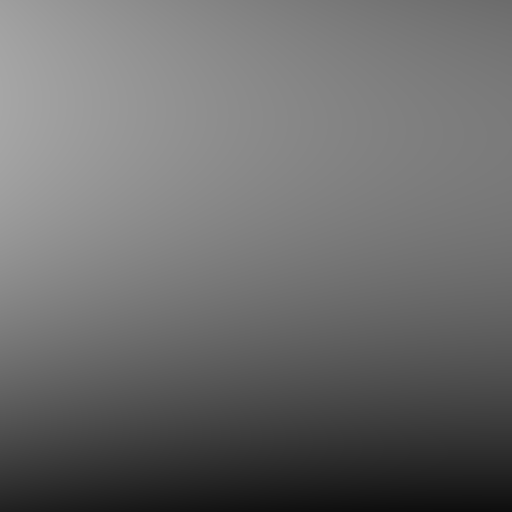}
        \caption{}
    \end{subfigure}\hfill
    \begin{subfigure}[b]{0.24\columnwidth}
        \centering
        \includegraphics[frame,width=0.98\linewidth]{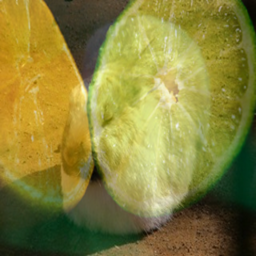}
        \caption{}
    \end{subfigure}\\[0.15cm]
    
    \begin{subfigure}[b]{0.24\columnwidth}
        \centering
        \includegraphics[frame,width=0.98\linewidth]{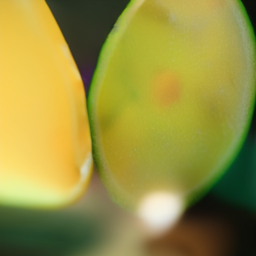}
        \caption{}
    \end{subfigure}\hfill
    \begin{subfigure}[b]{0.24\columnwidth}
        \centering
        \includegraphics[frame,width=0.98\linewidth]{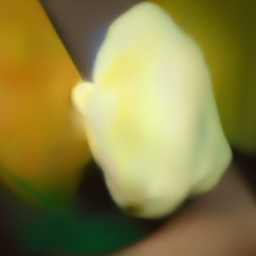}
        \caption{}
    \end{subfigure}\hfill
    \begin{subfigure}[b]{0.24\columnwidth}
        \centering
        \includegraphics[frame,width=0.98\linewidth]{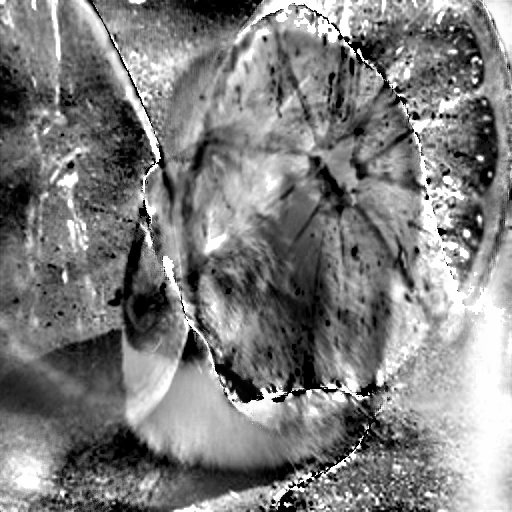}
        \caption{}
    \end{subfigure}\hfill
    \begin{subfigure}[b]{0.24\columnwidth}
        \centering
        \includegraphics[frame,width=0.98\linewidth]{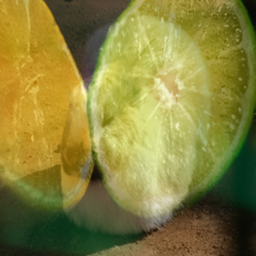}
        \caption{}
    \end{subfigure}
    
    \caption{\textbf{Failure of unconstrained pixel-wise $\alpha$.}
    (a,b) source images; (c) ground-truth $\alpha$-map; (d) blended input. (e-h) Result of optimizing a free per-pixel $\alpha$-map jointly with the sources. The blended result (h) matches the input (d), but the predicted map (g) clearly exhibits ``edge stealing,'' where high-frequency details leak into the mixing mechanism, demonstrating that spatial optimization fails without proper constraints.
    {\scriptsize{Prompts: ``\textit{A close up photo of a lemon cut in half}'', ``\textit{A white bunny sleeping}.''}}
    }
    \label{fig:full_spatial_alpha}
\end{figure}

Initial experiments with pixel-wise $\alpha$-maps revealed a failure mode: the optimization assigned high-frequency edge content to $\alpha$ while the recovered sources $x_0^1$ and $x_0^2$ remained blurred (\cref{fig:full_spatial_alpha}).

With $H \times W$ free parameters in a pixelwise alpha map (one per pixel), $\alpha$ can directly represent edges rather than serving as a blending mask. To address this, we constrain $\alpha$ to smooth spatial patterns by parameterizing it as a second-order polynomial:
\begin{equation} 
\label{eq:polynomial}
\alpha_{\phi}(\bar{u}, \bar{v}) = a + b\bar{u} + c\bar{v} + d\bar{u}\bar{v} + e\bar{u}^2 + f\bar{v}^2
\end{equation}
where $u \in \{0, \ldots, W-1\}$ and $v \in \{0, \ldots, H-1\}$ are pixel coordinates, $\bar{u} = u/(W-1)$ and $\bar{v} = v/(H-1)$ are their normalized values in $[0,1]$, and $\phi = \{a, b, c, d, e, f\}$ are the polynomial coefficients. This reduces the parameter space from $\mathbb{R}^{HW}$ to $\mathbb{R}^6$ while allowing spatially constant, linear gradients, and radial blending patterns. Second-order terms provide sufficient flexibility for common compositing scenarios while preventing high-frequency edge encoding.

We introduce unconstrained parameters $\tilde{\phi}$ and transform them to constrain $\alpha \in [0,1]$ at the four image corners. This reparameterization encourages smooth interior values near the valid range, but it does not mathematically bound every interior pixel. The transformation details are in \suppcref{appendix:alpha_reparameterization}.

\subsection{Two-Stage Decomposition Pipeline}
\label{subsec:method_two_stage_approach}
We address image decomposition using a coarse-to-fine pipeline that separates semantic decomposition from high-fidelity refinement (see \cref{fig:method_overview}). Latent diffusion provides effective text-guided semantic separation but can lose image details through VAE compression~\citep{rombach2022stable,bansal2023universal,he2024manifold}. Stage 1 therefore establishes a coarse separation in latent space using text guidance, and Stage 2 refines the selected decomposition with a pixel-space diffusion prior (see ablation in \suppcref{appendix:two_stage_ablation}).

Both stages employ the same core sampling procedure described in~\cref{subsec:method_joint_source_sampling}. To enhance sampling stability, we adopt the ``Time-Travel'' resampling strategy~\citep{lugmayr2022repaint,dontas2024blind}, performing iterative refinements at each diffusion timestep.
Additionally, in Stage~1 we dynamically optimize the alpha-map parameters $\tilde{\phi}$ every $k=5$ timesteps. At each optimization step, we freeze the current Tweedie estimates $\{\hat{x}_0^1(x_t^1), \hat{x}_0^2(x_t^2)\}$ and optimize $\alpha_\phi$ (\cref{eq:polynomial}).

In Stage 1, we establish coarse semantic structure using Stable Diffusion's latent diffusion model. Following CCDF~\citep{chung2022come}, we encode the blended image $y$ with the VAE encoder $\mathcal{E}$ and duplicate $\mathcal{E}(y)$ to initialize two latent branches $z_t^1, z_t^2$, noised to timestep $t_s < T$ (timestep selection ablated in \suppcref{appendix:timestep_ablation}). This preserves the overall spatial layout of the mixture while introducing initial diversity between the branches. To enforce semantic separation, we employ \textit{cross-negative prompting} where each source is conditioned on its own text prompt ($c^1_+$ or $c^2_+$) while using the opposite prompt as the negative condition in Classifier-Free Guidance~\citep{ho2022CFG} of $\epsilon_\theta$ with scale $\omega$ (i.e., $c^1_- = c^2_+$ and $c^2_- = c^1_+$). This contrastive guidance suppresses features belonging to the opposing source, encouraging clean semantic factorization (see ablation in \suppcref{appendix:text_guidance}).

Although sampling occurs in latent space, we enforce data consistency in pixel space following the universal guidance approach~\citep{bansal2023universal}: at each timestep, we decode the predicted latents $\hat{z}_0^1, \hat{z}_0^2$ to pixel space with the VAE decoder $\mathcal{D}$, giving $\hat{x}_0^1, \hat{x}_0^2$, and apply the reconstruction loss from~\cref{subsec:method_joint_source_sampling}. Finally, to leverage the stochastic nature of diffusion sampling, we generate $N_{\mathrm{cand}}=3$ decomposition candidates using different random seeds. Because the two output branches are unordered, we score both assignments between the two outputs and prompts using PickScore~\citep{kirstain2023pickapic}, retain the higher-scoring assignment, and select the candidate with the largest mean score. Selection uses only the Stage 1 outputs and prompts, without ground truth, and Stage 2 refines the selected pair. PickScore is an imperfect proxy trained for text-to-image preference rather than deblending quality; we therefore compare it with a permutation-invariant CLIP selector in \suppcref{appendix:selector_ablation}. In an interactive setting, a human user could instead visually select the best candidate among the generated samples. The effect of the candidate budget is evaluated in \suppcref{appendix:candidate_budget}.

In Stage 2, we refine the selected coarse estimates using a pretrained pixel-space diffusion model. Here $z_t = x_t$ and $\mathcal{D}$ is the identity. We add moderate noise to the coarse estimates up to timestep $t_r$ of the Stage~2 schedule ($t_r/T_r=350/600$, a lower relative noise level than $t_s/T_s=0.8$ in Stage~1), preserving the established semantic layout while allowing the model to synthesize high-frequency details. We then sample using the same joint sampling procedure, keeping $\alpha$ fixed at its Stage~1 estimate. This stage recovers fine-grained textures, corrects VAE compression artifacts, and encourages the final sources to blend consistently with the observed mixture.

\section{Experiments}
\label{sec:experiments}
We benchmark our method against five image decomposition methods: Deep Adversarial Decomposition (DAD) \cite{zou2020DAD}, the inverse problem method from \cite{zou2021GIP} (GIP), the Blind Image Decomposition Network (BIDeN) \cite{han2022BIDeN}, the cold diffusion \cite{bansal2023cold} approach from \cite{montenegro2025LCD} (LCD), and the training-free DoubleDIP method \cite{gandelsman2019doubleDIP}. Unlike our method, none of these baselines make use of text guidance; we detail this distinction and how it is handled in our evaluation in \cref{subsec:exp_setup}. Our evaluation examines three scenarios of increasing complexity, with each scenario relaxing assumptions about the blending $\alpha$ map: from known scalar values, to unknown scalars, and finally to unknown spatially-varying $\alpha$. The key advantage of our method is its ability to generalize across domains without requiring task-specific retraining. To demonstrate this, we also evaluate on the unseen Vehicles and Flowers domains (\cref{tab:baseline_results}, Right).
\subsection{Experimental Setup}
\label{subsec:exp_setup}
Existing image decomposition methods typically operate under restrictive assumptions regarding the mixing mechanism. Most approaches are limited to scalar blending factors, with some further constrained by the requirement that this scalar value be known a priori. In contrast, our method generalizes to spatially-varying $\alpha$-maps without requiring prior knowledge of the mixing scenario. To benchmark against current state-of-the-art methods while rigorously testing these limitations, we design our evaluation around three distinct scenarios of increasing complexity:
\begin{itemize}
    \item \textbf{Known scalar:} $\alpha = 0.5$, spatially constant, known.
    \item \textbf{Unknown scalar:} $\alpha \sim \mathcal{U}[0.2, 0.8]$, spatially constant, unknown.
    \item \textbf{Spatially-varying:} $\alpha$-maps sampled via \cref{eq:polynomial}, spatially varying, unknown.
\end{itemize}
We evaluate using the BIDeN benchmark \cite{han2022BIDeN}, constructing 300 test pairs from the \textbf{Fruits} and \textbf{Animals} domains. To strictly evaluate domain generalization, we additionally construct 100 random test pairs from two unseen domains: \textbf{Vehicles} and \textbf{Flowers}, which none of the supervised baselines were trained on.
Our baseline comparison covers both supervised and training-free paradigms. We evaluate four supervised methods: DAD \cite{zou2020DAD}, GIP \cite{zou2021GIP}, BIDeN \cite{han2022BIDeN}, and LCD \cite{montenegro2025LCD}. Additionally, we compare against the training-free DoubleDIP method \cite{gandelsman2019doubleDIP}. To ensure a fair comparison despite the limitations of the supervised architectures, we adapt their training procedures where possible:
\begin{itemize}
    \item For the \textbf{known scalar} scenario, we use the official pretrained BIDeN model and train DAD, GIP, and LCD on the BIDeN training set (Fruits and Animals). DoubleDIP is applied directly as it is training-free.
    \item For the \textbf{unknown scalar} scenario, we retrain DAD and BIDeN on an expanded $\alpha$ range of $[0.2, 0.8]$.
    \item For the \textbf{spatially-varying} scenario, we retrain DAD and BIDeN on pairs blended with polynomial $\alpha$-maps.
\end{itemize}
Note that GIP and LCD cannot be adapted to unknown or spatially-varying $\alpha$ values---GIP requires an inverse formula dependent on scalar $\alpha$; LCD uses $\alpha$ as initial diffusion timestep. Therefore, they are evaluated only in the known scalar scenario. Among the baselines, DoubleDIP is the only method that permits drawing multiple stochastic samples, so we evaluate it with the same three-sample budget used by our method and select among its samples with PickScore (\suppcref{appendix:baseline_oversampling}).
\begin{table*}[!t]
\centering
\setlength{\tabcolsep}{2pt} %
\scriptsize %
\newcolumntype{Y}{>{\centering\arraybackslash}X} 
\caption{Quantitative evaluation on seen (Fruits, Animals; 300 pairs) and unseen (Vehicles, Flowers; 100 pairs) domains. Sources are matched to the ground truth by minimum MSE. TF: training-free. Bold: best per scenario, domain and metric. LCD and GIP support only a known scalar $\alpha$. Ours and DoubleDIP select one of three candidates with PickScore; ours selects without ground truth, while DoubleDIP's outputs are first put in ground-truth order (\suppcref{appendix:baseline_oversampling}). Only ours uses text prompts (\cref{subsec:exp_setup}).}
\begin{tabularx}{\linewidth}{llcYYYYYY YYYYYY}
\toprule
\multicolumn{3}{c}{} & \multicolumn{6}{c}{\textbf{In-Distribution (Seen Domains)}} & \multicolumn{6}{c}{\textbf{Generalization (Unseen Domains)}} \\
\cmidrule(lr){4-9} \cmidrule(lr){10-15}

 & & & \multicolumn{3}{c}{\textbf{Fruits}} & \multicolumn{3}{c}{\textbf{Animals}} & \multicolumn{3}{c}{\textbf{Vehicles}} & \multicolumn{3}{c}{\textbf{Flowers}} \\
\cmidrule(lr){4-6} \cmidrule(lr){7-9} \cmidrule(lr){10-12} \cmidrule(lr){13-15}

\textbf{Scenario} & \textbf{Method} & \textbf{TF} & \textbf{PSNR}$\uparrow$ & \textbf{SSIM}$\uparrow$ & \textbf{LPIPS}$\downarrow$ & \textbf{PSNR}$\uparrow$ & \textbf{SSIM}$\uparrow$ & \textbf{LPIPS}$\downarrow$ & \textbf{PSNR}$\uparrow$ & \textbf{SSIM}$\uparrow$ & \textbf{LPIPS}$\downarrow$ & \textbf{PSNR}$\uparrow$ & \textbf{SSIM}$\uparrow$ & \textbf{LPIPS}$\downarrow$ \\
\midrule

\multirow{6}{*}{\makecell[l]{\textbf{Known}\\\textbf{Scalar}}}
& DAD & \ding{55} & \num{17.47} & \num{0.618} & \num{0.342} & \num{17.42} & \num{0.646} & \num{0.426} & \num{11.09} & \num{0.432} & \num{0.503} & \num{11.45} & \num{0.387} & \num{0.522} \\
& GIP & \ding{55} & \num{17.77} & \num{0.588} & \num{0.380} & \num{17.61} & \num{0.587} & \num{0.493} & \num{14.07} & \num{0.509} & \num{0.508} & \num{14.09} & \num{0.482} & \num{0.504} \\
& BIDeN & \ding{55} & \num{20.10} & \num{0.737} & \num{0.186} & \num{19.90} & \num{0.715} & \num{0.233} & \num{15.44} & \num{0.552} & \num{0.385} & \num{15.69} & \num{0.542} & \num{0.379} \\
& LCD & \ding{55} & \bfseries\num{23.04} & \bfseries\num{0.812} & \num{0.132} & \bfseries\num{23.06} & \num{0.795} & \num{0.249} & \num{17.21} & \num{0.689} & \num{0.293} & \num{17.19} & \num{0.655} & \num{0.298} \\
& DoubleDIP & \checkmark & \num{14.60} & \num{0.535} & \num{0.484} & \num{14.57} & \num{0.506} & \num{0.587} & \num{13.70} & \num{0.480} & \num{0.510} & \num{13.71} & \num{0.437} & \num{0.525} \\
& \textbf{Ours} & \checkmark & \num{21.36} & \num{0.804} & \bfseries\num{0.131} & \num{21.36} & \bfseries\num{0.823} & \bfseries\num{0.153} & \bfseries\num{20.23} & \bfseries\num{0.817} & \bfseries\num{0.121} & \bfseries\num{20.42} & \bfseries\num{0.806} & \bfseries\num{0.117} \\
\midrule

\multirow{4}{*}{\makecell[l]{\textbf{Unknown}\\\textbf{Scalar}}}
& DAD & \ding{55} & \num{17.87} & \num{0.627} & \num{0.344} & \num{17.52} & \num{0.639} & \num{0.427} & \num{14.90} & \num{0.527} & \num{0.467} & \num{15.79} & \num{0.543} & \num{0.432} \\
& BIDeN & \ding{55} & \bfseries\num{20.07} & \num{0.742} & \num{0.198} & \bfseries\num{19.86} & \num{0.723} & \num{0.252} & \num{14.52} & \num{0.495} & \num{0.464} & \num{15.65} & \num{0.538} & \num{0.399} \\
& DoubleDIP & \checkmark & \num{14.84} & \num{0.536} & \num{0.500} & \num{14.55} & \num{0.496} & \num{0.603} & \num{13.47} & \num{0.438} & \num{0.570} & \num{14.47} & \num{0.469} & \num{0.491} \\
& \textbf{Ours} & \checkmark & \num{20.03} & \bfseries\num{0.758} & \bfseries\num{0.174} & \num{19.54} & \bfseries\num{0.769} & \bfseries\num{0.206} & \bfseries\num{18.90} & \bfseries\num{0.771} & \bfseries\num{0.154} & \bfseries\num{20.06} & \bfseries\num{0.788} & \bfseries\num{0.126} \\
\midrule

\multirow{4}{*}{\makecell[l]{\textbf{Spatially-}\\\textbf{Varying}}}
& DAD & \ding{55} & \num{16.62} & \num{0.613} & \num{0.362} & \num{17.28} & \num{0.665} & \num{0.394} & \num{14.50} & \num{0.555} & \num{0.430} & \num{14.86} & \num{0.525} & \num{0.441} \\
& BIDeN & \ding{55} & \bfseries\num{18.21} & \num{0.692} & \num{0.247} & \bfseries\num{18.65} & \num{0.690} & \num{0.278} & \num{14.68} & \num{0.528} & \num{0.422} & \num{15.01} & \num{0.521} & \num{0.406} \\
& DoubleDIP & \checkmark & \num{13.59} & \num{0.488} & \num{0.538} & \num{13.83} & \num{0.443} & \num{0.650} & \num{12.97} & \num{0.410} & \num{0.574} & \num{13.14} & \num{0.385} & \num{0.578} \\
& \textbf{Ours} & \checkmark & \num{17.91} & \bfseries\num{0.715} & \bfseries\num{0.210} & \num{18.19} & \bfseries\num{0.750} & \bfseries\num{0.213} & \bfseries\num{17.81} & \bfseries\num{0.755} & \bfseries\num{0.167} & \bfseries\num{18.35} & \bfseries\num{0.750} & \bfseries\num{0.158} \\
\bottomrule
\end{tabularx}

\label{tab:baseline_results}
\end{table*}

We assess decomposition quality using PSNR, SSIM~\cite{wang2004ssim}, and LPIPS~\cite{zhang2018lpips}. Metrics are computed separately for each recovered source. To resolve the inherent ordering ambiguity in blind decomposition (i.e., identifying which output corresponds to which ground truth), we compute MSE for both possible assignments and select the permutation that minimizes the total MSE before calculating the final metrics.
Unlike the baselines, our method uses semantic information from text guidance, so the main comparison with manually written prompts is not strictly information-matched. We write these prompts from the blended image alone, without access to the ground-truth sources, and describe only visually inferable features (examples in \cref{fig:full_spatial_alpha,fig:teaser_fig}). To quantify this dependence, \suppcref{appendix:text_guidance} evaluates prompt-free sampling with $\omega=0$, and \suppcref{appendix:auto_prompts} evaluates a fully automatic alternative in which a vision--language model generates both prompts from the blended image. In the main evaluation, we generate $N_{\mathrm{cand}}=3$ candidates and apply permutation-invariant PickScore selection before Stage 2; candidate-budget and selector ablations are reported in \suppcref{appendix:candidate_budget,appendix:selector_ablation}.

\subsection{Experimental Results}
\label{subsec:results}
Quantitative results for all three scenarios are summarized in \cref{tab:baseline_results}, with corresponding qualitative comparisons shown in \cref{fig:combined-comparison} (with additional results in the supplementary material).

\begin{figure*}[tb]
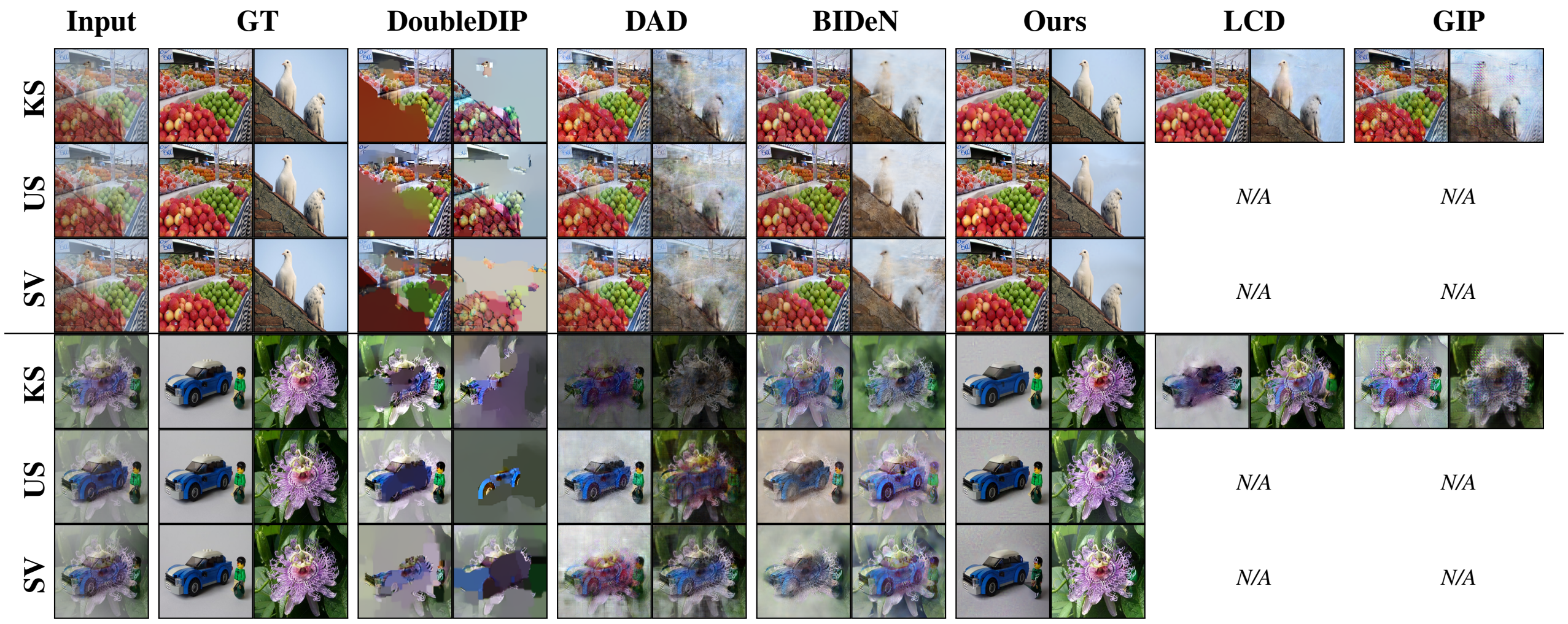

\centering
\small
\renewcommand{\arraystretch}{0.75}

\newcommand{\sourceAA}{market}
\newcommand{\sourceBA}{pigeons}
\newcommand{\mixNameA}{\sourceAA-\sourceBA}

\newcommand{\sourceAB}{lego}
\newcommand{\sourceBB}{purple-flower}
\newcommand{\mixNameB}{\sourceAB-\sourceBB}

\newcommand{\framedimage}[1]{\includegraphics[width=0.057\textwidth, valign=m, frame]{#1}}

\newcommand{\vertlabel}[1]{\rotatebox[origin=c]{90}{\textbf{#1}}}

\newcommand{\notapplicable}{\multicolumn{2}{c}{\scriptsize\textit{N/A}}}

\begin{tabular}{
  c @{\hskip 0.1cm}
  c @{\hskip 0.1cm}
  c@{}c @{\hskip 0.1cm}
  c@{}c @{\hskip 0.1cm}
  c@{}c @{\hskip 0.1cm}
  c@{}c @{\hskip 0.1cm}
  c@{}c @{\hskip 0.1cm}
  c@{}c @{\hskip 0.1cm}
  c@{}c
}

& \textbf{Input} & \multicolumn{2}{c}{\textbf{GT}} & \multicolumn{2}{c}{\textbf{DoubleDIP}} & \multicolumn{2}{c}{\textbf{DAD}} & \multicolumn{2}{c}{\textbf{BIDeN}} & \multicolumn{2}{c}{\textbf{Ours}} & \multicolumn{2}{c}{\textbf{LCD}} & \multicolumn{2}{c}{\textbf{GIP}} \\[0.1cm]

\vertlabel{KS}
  & \framedimage{figures/4_experiments/baseline_evaluation/scenario1_images/input/\mixNameA.png}
  & \framedimage{figures/4_experiments/baseline_evaluation/scenario1_images/ground_truth/\sourceAA.png}
  & \framedimage{figures/4_experiments/baseline_evaluation/scenario1_images/ground_truth/\sourceBA.png}
  & \framedimage{figures/4_experiments/baseline_evaluation/scenario1_images/DoubleDIP/\sourceAA.png}
  & \framedimage{figures/4_experiments/baseline_evaluation/scenario1_images/DoubleDIP/\sourceBA.png}
  & \framedimage{figures/4_experiments/baseline_evaluation/scenario1_images/DAD/\sourceAA.png}
  & \framedimage{figures/4_experiments/baseline_evaluation/scenario1_images/DAD/\sourceBA.png}
  & \framedimage{figures/4_experiments/baseline_evaluation/scenario1_images/BIDeN/\sourceAA.png}
  & \framedimage{figures/4_experiments/baseline_evaluation/scenario1_images/BIDeN/\sourceBA.png}
  & \framedimage{figures/4_experiments/baseline_evaluation/scenario1_images/Ours/\sourceAA.png}
  & \framedimage{figures/4_experiments/baseline_evaluation/scenario1_images/Ours/\sourceBA.png}
  & \framedimage{figures/4_experiments/baseline_evaluation/scenario1_images/LCD/\sourceAA.png}
  & \framedimage{figures/4_experiments/baseline_evaluation/scenario1_images/LCD/\sourceBA.png}
  & \framedimage{figures/4_experiments/baseline_evaluation/scenario1_images/GIP/\sourceAA.png}
  & \framedimage{figures/4_experiments/baseline_evaluation/scenario1_images/GIP/\sourceBA.png} \\

\vertlabel{US}
  & \framedimage{figures/4_experiments/baseline_evaluation/scenario2_images/input/\mixNameA.png}
  & \framedimage{figures/4_experiments/baseline_evaluation/scenario2_images/ground_truth/\sourceAA.png}
  & \framedimage{figures/4_experiments/baseline_evaluation/scenario2_images/ground_truth/\sourceBA.png}
  & \framedimage{figures/4_experiments/baseline_evaluation/scenario2_images/DoubleDIP/\sourceAA.png}
  & \framedimage{figures/4_experiments/baseline_evaluation/scenario2_images/DoubleDIP/\sourceBA.png}
  & \framedimage{figures/4_experiments/baseline_evaluation/scenario2_images/DAD/\sourceAA.png}
  & \framedimage{figures/4_experiments/baseline_evaluation/scenario2_images/DAD/\sourceBA.png}
  & \framedimage{figures/4_experiments/baseline_evaluation/scenario2_images/BIDeN/\sourceAA.png}
  & \framedimage{figures/4_experiments/baseline_evaluation/scenario2_images/BIDeN/\sourceBA.png}
  & \framedimage{figures/4_experiments/baseline_evaluation/scenario2_images/Ours/\sourceAA.png}
  & \framedimage{figures/4_experiments/baseline_evaluation/scenario2_images/Ours/\sourceBA.png}
  & \notapplicable
  & \notapplicable
  \\

\vertlabel{SV}
  & \framedimage{figures/4_experiments/baseline_evaluation/scenario3_images/input/\mixNameA.png}
  & \framedimage{figures/4_experiments/baseline_evaluation/scenario3_images/ground_truth/\sourceAA.png}
  & \framedimage{figures/4_experiments/baseline_evaluation/scenario3_images/ground_truth/\sourceBA.png}
  & \framedimage{figures/4_experiments/baseline_evaluation/scenario3_images/DoubleDIP/\sourceAA.png}
  & \framedimage{figures/4_experiments/baseline_evaluation/scenario3_images/DoubleDIP/\sourceBA.png}
  & \framedimage{figures/4_experiments/baseline_evaluation/scenario3_images/DAD/\sourceAA.png}
  & \framedimage{figures/4_experiments/baseline_evaluation/scenario3_images/DAD/\sourceBA.png}
  & \framedimage{figures/4_experiments/baseline_evaluation/scenario3_images/BIDeN/\sourceAA.png}
  & \framedimage{figures/4_experiments/baseline_evaluation/scenario3_images/BIDeN/\sourceBA.png}
  & \framedimage{figures/4_experiments/baseline_evaluation/scenario3_images/Ours/\sourceAA.png}
  & \framedimage{figures/4_experiments/baseline_evaluation/scenario3_images/Ours/\sourceBA.png}
  & \notapplicable
  & \notapplicable
  \\
  \hline

\vertlabel{KS}
  & \framedimage{figures/4_experiments/generalization/scenario1_images/input/\mixNameB.png}
  & \framedimage{figures/4_experiments/generalization/scenario1_images/ground_truth/\sourceAB.png}
  & \framedimage{figures/4_experiments/generalization/scenario1_images/ground_truth/\sourceBB.png}
  & \framedimage{figures/4_experiments/generalization/scenario1_images/DoubleDIP/\sourceAB.png}
  & \framedimage{figures/4_experiments/generalization/scenario1_images/DoubleDIP/\sourceBB.png}
  & \framedimage{figures/4_experiments/generalization/scenario1_images/DAD/\sourceAB.png}
  & \framedimage{figures/4_experiments/generalization/scenario1_images/DAD/\sourceBB.png}
  & \framedimage{figures/4_experiments/generalization/scenario1_images/BIDeN/\sourceAB.png}
  & \framedimage{figures/4_experiments/generalization/scenario1_images/BIDeN/\sourceBB.png}
  & \framedimage{figures/4_experiments/generalization/scenario1_images/Ours/\sourceAB.png}
  & \framedimage{figures/4_experiments/generalization/scenario1_images/Ours/\sourceBB.png}
  & \framedimage{figures/4_experiments/generalization/scenario1_images/LCD/\sourceAB.png}
  & \framedimage{figures/4_experiments/generalization/scenario1_images/LCD/\sourceBB.png}
  & \framedimage{figures/4_experiments/generalization/scenario1_images/GIP/\sourceAB.png}
  & \framedimage{figures/4_experiments/generalization/scenario1_images/GIP/\sourceBB.png} \\

\vertlabel{US}
  & \framedimage{figures/4_experiments/generalization/scenario2_images/input/\mixNameB.png}
  & \framedimage{figures/4_experiments/generalization/scenario2_images/ground_truth/\sourceAB.png}
  & \framedimage{figures/4_experiments/generalization/scenario2_images/ground_truth/\sourceBB.png}
  & \framedimage{figures/4_experiments/generalization/scenario2_images/DoubleDIP/\sourceAB.png}
  & \framedimage{figures/4_experiments/generalization/scenario2_images/DoubleDIP/\sourceBB.png}
  & \framedimage{figures/4_experiments/generalization/scenario2_images/DAD/\sourceAB.png}
  & \framedimage{figures/4_experiments/generalization/scenario2_images/DAD/\sourceBB.png}
  & \framedimage{figures/4_experiments/generalization/scenario2_images/BIDeN/\sourceAB.png}
  & \framedimage{figures/4_experiments/generalization/scenario2_images/BIDeN/\sourceBB.png}
  & \framedimage{figures/4_experiments/generalization/scenario2_images/Ours/\sourceAB.png}
  & \framedimage{figures/4_experiments/generalization/scenario2_images/Ours/\sourceBB.png}
  & \notapplicable
  & \notapplicable
  \\

\vertlabel{SV}
  & \framedimage{figures/4_experiments/generalization/scenario3_images/input/\mixNameB.png}
  & \framedimage{figures/4_experiments/generalization/scenario3_images/ground_truth/\sourceAB.png}
  & \framedimage{figures/4_experiments/generalization/scenario3_images/ground_truth/\sourceBB.png}
  & \framedimage{figures/4_experiments/generalization/scenario3_images/DoubleDIP/\sourceAB.png}
  & \framedimage{figures/4_experiments/generalization/scenario3_images/DoubleDIP/\sourceBB.png}
  & \framedimage{figures/4_experiments/generalization/scenario3_images/DAD/\sourceAB.png}
  & \framedimage{figures/4_experiments/generalization/scenario3_images/DAD/\sourceBB.png}
  & \framedimage{figures/4_experiments/generalization/scenario3_images/BIDeN/\sourceAB.png}
  & \framedimage{figures/4_experiments/generalization/scenario3_images/BIDeN/\sourceBB.png}
  & \framedimage{figures/4_experiments/generalization/scenario3_images/Ours/\sourceAB.png}
  & \framedimage{figures/4_experiments/generalization/scenario3_images/Ours/\sourceBB.png}
  & \notapplicable
  & \notapplicable
  \\

\end{tabular}

\caption{\textbf{Qualitative alpha-deblending comparison} for Known Scalar (\textbf{KS}), Unknown Scalar (\textbf{US}), and Spatially Varying (\textbf{SV}) mixing (rows) on a seen (Market + Pigeons) and an unseen (Lego + Purple Flower) pair, chosen by eye as representative examples; further examples are in the supplement. Each method shows its two recovered sources side by side; \textbf{GT} gives the two ground-truth sources. LCD and GIP apply only to KS, and \emph{N/A} marks inapplicable methods. Baselines leave residual traces of one source in the other, which our method largely avoids.}
\label{fig:combined-comparison}
\end{figure*}

Our strongest gains occur on domains absent from task-specific deblending training. In the \textbf{Known Scalar} setting, for example, LCD decreases from 23.06 dB PSNR on Animals to 17.21 dB on Vehicles, and DAD decreases from 17.42 to 11.09 dB, whereas our method changes from 21.36 to 20.23 dB. Across Vehicles and Flowers, our method achieves the best PSNR, SSIM, and LPIPS in all three blending scenarios. These results are consistent with better domain transfer from the pretrained diffusion priors, without task-specific retraining.

The same trend holds as assumptions about the blending mechanism are relaxed. In the \textbf{Unknown Scalar} and \textbf{Spatially-Varying} settings, our method obtains the best SSIM and LPIPS in every domain; on the unseen domains, it also obtains the best PSNR. Unlike the supervised baselines, which are retrained for each supported setting, our formulation optimizes the scalar or polynomial $\alpha$-map jointly with the source images at inference time.

On the seen domains in the \textbf{Known Scalar} setting, LCD attains the highest PSNR and our method remains competitive: it obtains the lowest LPIPS on both domains and the highest SSIM on Animals. LCD and BIDeN degrade more substantially on Vehicles and Flowers, while our method remains comparatively stable. Under the same three-sample PickScore selection budget, DoubleDIP remains in the 13--15 dB PSNR range across the evaluated settings, compared with about 18--21 dB for our method (\suppcref{appendix:baseline_oversampling}).

\paragraph{Ablation Study.}
Stage~2 raises mean PSNR over Stage~1 by 2.40--3.97\,dB, with consistent SSIM and LPIPS gains (\suppcref{appendix:two_stage_ablation}). Intermediate noising works best: $t_s=80$ and $t_r=350$ give the strongest results among the tested starting timesteps, balancing preserved mixture structure against room for separation and refinement (\suppcref{appendix:timestep_ablation}). Raising Stage~2 Time-Travel recurrences from eight to ten gains only 0.13\,dB for 3.5 extra minutes, so we use eight in both stages (\suppcref{appendix:recurrence_ablation}). On spatially varying blends, cross-negative prompting at $\omega=2$ improves mean PSNR by 2.45\,dB over standard CFG and by 3.58\,dB over prompt-free sampling (\suppcref{appendix:text_guidance}).

PickScore improves the aggregate mean over CLIP in all 18 domain--scenario--metric comparisons, although this comparison is exploratory because the selector was adopted after inspecting these sets (\suppcref{appendix:selector_ablation}). With the true $\alpha$ supplied, going from one to 16 candidates gains 1.90 and 2.12\,dB on known and unknown scalar blends but gives no consistent spatially varying gain, so we use three (\suppcref{appendix:candidate_budget}). A vision--language model can also write both prompts from the mixture alone, recovering about half of the PSNR gain of manual prompts over prompt-free sampling (\suppcref{appendix:auto_prompts}).

Our method records the lowest LPIPS in every domain and blending scenario (\cref{tab:baseline_results}). In \cref{fig:combined-comparison}, the supervised baselines frequently leave ghosting along high-contrast edges, tinting from the other layer's colors, and cross-contamination where the images overlap, and these artifacts intensify as the blending assumptions are relaxed. Our separations preserve object boundaries and texture with markedly less bleed, suggesting that the pretrained prior supplies plausible content where the forward model alone is under-determined.

\section{Conclusion}
\label{sec:conclusion}
We presented a training-free framework for blind alpha deblending that jointly recovers two source images and the blending parameters using pretrained diffusion priors. It generalizes to domains absent from task-specific deblending training and handles spatially varying alpha maps without retraining, reusing the same single-image diffusion model for both source branches. Its stochastic sampler also provides alternative decompositions when an individual sample fails.

\paragraph{Limitations.}
Our method is intended for offline rather than real-time use. Generating three Stage~1 candidates and refining the selected candidate takes 50.6--71.0 minutes per decomposition, excluding selector overhead. Text guidance also makes the main comparison with the baselines not strictly information-matched. The sampler can operate without prompts at $\omega=0$, and our automated-prompt experiment shows that both prompts can be generated from the blended observation alone (\suppcref{appendix:text_guidance,appendix:auto_prompts}). In addition, PickScore was trained for ordinary text-to-image preferences and is an imperfect proxy for deblending quality. Our polynomial alpha parameterization favors smooth blending maps and constrains their corner values, but does not strictly bound every interior value or represent arbitrary high-frequency masks; our spatially varying benchmark is drawn from the same quadratic family, and smooth mattes outside it degrade performance gradually (\suppcref{appendix:alpha_out_of_family}). Finally, the current formulation is limited to two source layers. Future work includes reducing runtime, possibly by unifying the two stages, extending the formulation to $K>2$ layers, and learning prompt representations and deblending-specific candidate selectors.

\section*{Acknowledgements}
This work was partially supported by the Israel Science Foundation (ISF) grant no.\ 1574/21.

{
    \small
    \bibliographystyle{ieeenat_fullname}
    \bibliography{references}
}

\ifincludesupp
\appendix
\maketitlesupplementary
\renewcommand{\thesection}{\Alph{section}}
\renewcommand{\thesubsection}{\thesection.\arabic{subsection}}
\setcounter{section}{0}
\section{Preliminaries}
\label{appendix:preliminaries}

\subsection{Denoising Diffusion Probabilistic Models (DDPM)}
\label{appendix:diffusion_models}

Denoising Diffusion Probabilistic Models~\citep{dickestein2015deepunsupervised,ho2020DDPM} are generative models that learn to reverse a fixed noising process. The forward process gradually corrupts data by adding Gaussian noise:
\begin{equation}
\label{eq:forward_noise_distribution}
q(x_t \mid x_{t-1}) := \mathcal{N}(x_t; \sqrt{1-\beta_t}\, x_{t-1}, \beta_t I)
\end{equation}
where $\{\beta_t\}_{t=1}^T$ is a fixed variance schedule. Using the reparameterization $x_t = a_t x_{t-1} + \sqrt{\beta_t}\,\varepsilon$ with $a_t := \sqrt{1-\beta_t}$ and $\varepsilon \sim \mathcal{N}(0,I)$, we obtain a closed-form expression for sampling $x_t$ directly from $x_0$:
\begin{equation}
\label{eq:perturbation_noise_distribution}
q(x_t \mid x_0) = \mathcal{N}(x_t; \sqrt{\bar{a}_t}\, x_0, (1-\bar{a}_t)I)
\end{equation}
where $\bar{a}_t := \prod_{i=1}^t a_i$ is the cumulative noise coefficient. Equivalently:
\begin{equation}
\label{eq:forward_diffusion}
x_t = \sqrt{\bar{a}_t}\, x_0 + \sqrt{1-\bar{a}_t}\, \varepsilon, \quad \varepsilon \sim \mathcal{N}(0,I).
\end{equation}

The reverse process $p_\theta$ learns to denoise iteratively from $x_T \sim \mathcal{N}(0,I)$ back to $x_0$. The model predicts the noise $\varepsilon_\theta(x_t, t)$ added during the forward process, yielding the update:
\begin{equation}
\label{eq:reverse_mean}
x_{t-1} = \frac{1}{\sqrt{a_t}}\left(x_t - \frac{\beta_t}{\sqrt{1-\bar{a}_t}}\,\varepsilon_\theta(x_t,t)\right) + \sigma_t \varepsilon
\end{equation}
where $\sigma_t$ controls sampling stochasticity; setting $\sigma_t = 0$ for all $t$ recovers the deterministic DDIM sampler~\citep{song2021ddim}, which we adopt in our method (\cref{sec:preliminaries}), so that all stochasticity in our sampling procedure arises solely from the initial noise $x_T$. The model is trained by minimizing:
\begin{equation}
\label{eq:ddpm_loss}
\mathcal{L}_{\text{simple}} = \mathbb{E}_{t,x_0,\varepsilon}\left[\|\varepsilon - \varepsilon_\theta(x_t,t)\|^2\right].
\end{equation}
\section{Method}
In this appendix, we provide the detailed mathematical formulations supporting the method described in \cref{sec:method}. We first present the derivation of the likelihood gradient for the joint posterior sampling in \cref{appendix:joint_posterior_derivations}, justifying the approximation of the Dirac delta distribution. Subsequently, in \cref{appendix:alpha_reparameterization}, we detail the cascaded reparameterization strategy used to constrain the polynomial alpha map at its four corners.

\label{appendix:method_derivations}
\subsection{Likelihood Gradient Derivation and Decomposition}
\label{appendix:joint_posterior_derivations}
This section derives the likelihood gradient for joint posterior sampling introduced in \cref{subsec:method_joint_source_sampling}. 
Our alpha-blending operator $A$ defined in \cref{eq:blending_forward} is deterministic, unlike standard inverse problems with stochastic measurement noise. However, the forward process operates on clean images $x_0^1, x_0^2$, while sampling occurs in the noisy space $x_t^1, x_t^2$. We address this through the Tweedie approximation (\cref{eq:tweedies_formula}) and model the likelihood as a Dirac delta distribution:
\begin{equation}
\label{eq:app_likelihood_dirac}
p(y|x_t^1,x_t^2) = \delta(y - A(x_t^1,x_t^2, \alpha))
\end{equation}
Since the Dirac delta is not differentiable, we approximate it with a Gaussian distribution of small variance $\sigma_y$, formally corresponding to the limit $\sigma_y \to 0^+$; in practice, we treat $\sigma_y$ (or equivalently, its associated scaling $1/\sigma_y^2$) as a finite hyperparameter controlling the strength of data-consistency guidance, rather than taking the limit literally. Following \cref{eq:tweedies_formula}, we use the Tweedie estimates $\hat{x}_0^1 = \mathbb{E}[x_0^1|x_t^1]$ and $\hat{x}_0^2 = \mathbb{E}[x_0^2|x_t^2]$ to approximate the likelihood:
\begin{multline}
\label{eq:app_likelihood_gaussian}
p(y|x_t^1,x_t^2) \approx p(y|\hat{x}_0^1,\hat{x}_0^2) \propto\\ 
\exp\left(-\frac{1}{2\sigma_y^2}\|y - 
A(\hat{x}_0^1,\hat{x}_0^2,\alpha)\|^2\right),
\end{multline}
where $\|\cdot\|$ denotes the $L_2$ norm and the proportionality constant absorbs the Gaussian normalization factor.
Taking the logarithm yields:
\begin{equation}
\label{eq:app_log_likelihood}
\log p(y|\hat{x}_0^1,\hat{x}_0^2) = -\frac{1}{2\sigma_y^2}\|y - 
A(\hat{x}_0^1,\hat{x}_0^2,\alpha)\|^2 + \text{const},
\end{equation}
where the constant term is independent of the estimated clean sources $\hat{x}_0^1, \hat{x}_0^2$.
Computing the gradient with respect to both noisy sources, the constant term vanishes, yielding:
\begin{equation}
\label{eq:app_likelihood_gradient_full}
\nabla_{x_t^1,x_t^2} \log p(y|\hat{x}_0^1,\hat{x}_0^2) = 
-\frac{1}{2\sigma_y^2} \nabla_{x_t^1,x_t^2} \|y - 
A(\hat{x}_0^1,\hat{x}_0^2, \alpha)\|^2,
\end{equation}
where the gradient is computed via the chain rule through the Tweedie estimates $\hat{x}_0^i(x_t^i)$ from \cref{eq:tweedies_formula}.
In practice, we absorb the variance term $1/(2\sigma_y^2)$ into a time-dependent scaling factor $s(t)$, yielding the final form:
\begin{equation}
\label{eq:app_likelihood_score_final}
\nabla_{x_t^1,x_t^2} \log p(y|\hat{x}_0^1,\hat{x}_0^2) = 
-s(t) \nabla_{x_t^1,x_t^2} \|y - A(\hat{x}_0^1,\hat{x}_0^2, \alpha)\|^2,
\end{equation}
which corresponds to \cref{eq:likelihood_score_function} in the main text. 
The blending operator $A(\hat{x}_0^1, \hat{x}_0^2, \alpha) = \alpha \odot \hat{x}_0^1 + (1-\alpha) \odot \hat{x}_0^2$ enables clean decomposition of the joint gradient. For the likelihood term, we have:
\begin{equation}
\nabla_{\hat{x}_0^1} \|y - A(\hat{x}_0^1, \hat{x}_0^2, \alpha)\|^2 = -2\alpha \odot (y - A(\hat{x}_0^1, \hat{x}_0^2, \alpha)),
\end{equation}
\begin{equation}
\nabla_{\hat{x}_0^2} \|y - A(\hat{x}_0^1, \hat{x}_0^2, \alpha)\|^2 = -2(1-\alpha) \odot (y - A(\hat{x}_0^1, \hat{x}_0^2, \alpha)).
\end{equation}
The gradients with respect to the noisy sources $x_t^1, x_t^2$ are obtained by applying the chain rule through the Tweedie estimates. This decomposition shows that each source is updated proportionally to its contribution weight $\alpha$ (or $1-\alpha$) in the blend.
\subsection{Alpha Map Reparameterization}
\label{appendix:alpha_reparameterization}
The polynomial $\alpha$-map from \cref{eq:polynomial} should ideally satisfy $\alpha(\bar{u},\bar{v}) \in [0, 1]$ over the image domain. Our reparameterization explicitly constrains its values at the four corners. Although this does not guarantee the same bound at every interior location, the low-order polynomial restricts the map to smooth spatial variation.
Evaluating \cref{eq:polynomial} at the four corners yields:
\begin{align}
\alpha(0,0) &= a \\
\alpha(1,0) &= a + b + e \\
\alpha(0,1) &= a + c + f \\
\alpha(1,1) &= a + b + c + d + e + f
\end{align}
We use a cascaded parameterization that constrains these four corner values to $(0,1)$. Interior values are not explicitly bounded. We define a smooth bounded transformation $\tau: \mathbb{R} \to (0,1)$:
\begin{equation}
\tau(x) = \frac{1}{2}\tanh(x) + \frac{1}{2}
\label{eq:tau}
\end{equation}
Each coefficient is constructed through a cascaded parameterization where it depends on previously defined coefficients:
\begin{align}
a &= \tau(\tilde{a}) \label{eq:param_a}\\
b &= -a + \tau(\tilde{b}) \label{eq:param_b}\\
e &= -(a + b) + \tau(\tilde{e}) \label{eq:param_e}\\
c &= -a + \tau(\tilde{c}) \label{eq:param_c}\\
f &= -(a + c) + \tau(\tilde{f}) \label{eq:param_f}\\
d &= -(a + b + c + e + f) + \tau(\tilde{d}) \label{eq:param_d}
\end{align}
By this construction, each corner value simplifies to a single $\tau(\cdot)$ term:
\begin{align}
\alpha(0,0) &= a = \tau(\tilde{a}) \\
\alpha(1,0) &= a + b + e = \tau(\tilde{e}) \\
\alpha(0,1) &= a + c + f = \tau(\tilde{f}) \\
\alpha(1,1) &= a + b + c + d + e + f = \tau(\tilde{d})
\end{align}
Thus, all four corner values lie in $(0,1)$. This construction does not provide a global bound: a second-order polynomial can attain an interior extremum outside the range of its corner values. In our experiments, this corner-constrained, low-dimensional parameterization produced smooth $\alpha$-maps whose interior values typically remained near the valid range. We nevertheless treat the absence of a strict interior bound as a limitation of the parameterization.
\section{Experiments}
This appendix details the experimental setup and specific configurations used to evaluate our method. We first describe the adaptation of the Generative Inverse Problems (GIP) baseline for the alpha-deblending task in \cref{appendix:baseline_models_training}. Subsequently, in \cref{appendix:implementation_details}, we provide a comprehensive breakdown of the hyperparameters, model architectures, and optimization strategies employed in our two-stage pipeline.

\subsection{Baseline Models Training}
\label{appendix:baseline_models_training}

\subsubsection{DoubleDIP}
DoubleDIP~\cite{gandelsman2019doubleDIP} is a training-free method that decomposes images by optimizing two Deep Image Prior (DIP) networks at test time. Each network generates one source component, and their combination is constrained to match the observed mixture.

We use the default architecture from the official implementation with images resized to $256 \times 256$ pixels. For all scenarios, we optimize for 8000 iterations with Adam (learning rate $10^{-3}$) using the reconstruction loss $\|\alpha \cdot \text{DIP}_1(z_1) + (1-\alpha) \cdot \text{DIP}_2(z_2) - y\|^2$ plus the DoubleDIP exclusion loss between the two outputs with weight $0.1$. In the known scalar scenario, $\alpha = 0.5$ is fixed. For the unknown scalar and spatially-varying scenarios, we optimize $\alpha$ jointly with the source images using an additional DIP network of the same architecture, whose output is mapped to $[0.05, 0.95]$ (its center pixel in the scalar scenario) and pulled toward $0.5$ with weight $0.5$ during the first 1000 iterations. We return the iterate whose re-blended output best matches $y$.

\subsubsection{DAD}
Deep Adversarial Decomposition (DAD)~\cite{zou2020DAD} trains two generator networks $G_1$ and $G_2$ to decompose a mixed image $y$ into two source images $\hat{x}_1$ and $\hat{x}_2$. We use the default architecture and hyperparameters from the official implementation with images resized to $256 \times 256$ pixels. Training pairs are randomly sampled from the BIDeN training set (Fruits and Animals domains). For each scenario, we train a 256-resolution U-Net generator from scratch for 200 epochs and keep the best validation checkpoint: for the known scalar scenario with the official mixing, $\alpha \sim \mathcal{U}[0.4, 0.6]$ centered on the evaluated $\alpha=0.5$; for the unknown scalar scenario with $\alpha \sim \mathcal{U}[0.2, 0.8]$; and for the spatially-varying scenario with polynomial $\alpha$-maps (\cref{eq:polynomial}).

\subsubsection{GIP}
The GIP framework~\cite{zou2021GIP} solves inverse problems $y = \mathcal{F}(x; z)$ by training a generator to predict degradation parameters $z = G(y)$, then recovering the source via $x = \mathcal{F}^{-1}(y; z)$. GIP enables unsupervised training through a bidirectional constraint: applying predicted degradations to random clean images should produce realistic degraded observations, enforced by two discriminators ensuring domain consistency.

We reimplement GIP in PyTorch and adapt it to alpha deblending by designating $x_2$ as the degradation parameter to predict: $\hat{x}_2 = G(y)$, and recovering the first source image via $x_1 = (y - (1-\alpha)\hat{x}_2) / \alpha$. This formulation requires $\alpha$ to be known in advance, limiting GIP evaluation to the known scalar scenario only.

\subsubsection{BIDeN}
Blind Image Decomposition Network (BIDeN)~\cite{han2022BIDeN} decomposes mixed images into $N$ source components using a single encoder-decoder network with $N$ decoder branches.
For the known scalar scenario, we use the official pretrained model released by the authors. For the unknown scalar and spatially-varying scenarios, we retrain from scratch using the default architecture with $N=2$ components and the official loss weights and learning rate, with batch size 8 for 1000 epochs (unknown scalar) and 200 epochs (spatially-varying). Training pairs are randomly sampled from the BIDeN training set (Fruits and Animals domains). Images are resized to $256 \times 256$ pixels. For the unknown scalar scenario, $\alpha$ is randomly sampled from $\mathcal{U}[0.2, 0.8]$ during training. For the spatially-varying scenario, blending is performed with polynomial $\alpha$-maps (\cref{eq:polynomial}). The model operates directly on the blended measurement input without requiring knowledge of the mixing mechanism.

\subsubsection{LCD}
Latent Cold Diffusion (LCD)~\cite{montenegro2025LCD} adapts the cold diffusion framework~\cite{bansal2023cold} to image decomposition by treating the blending process as a degradation operator. 
We reimplement LCD~\cite{montenegro2025LCD} on top of the cold diffusion framework~\cite{bansal2023cold}: a U-Net is trained from scratch on the BIDeN training set (Fruits and Animals domains) at $256 \times 256$ to predict one source from mixtures whose mixing weight follows a linear schedule from 0 to 0.8 (47k iterations, learning rate $10^{-4}$). At inference, sampling starts from the observed $\alpha = 0.5$ mixture and applies the LCD sampling procedure from~\cite{montenegro2025LCD}; the held-out Vehicles and Flowers evaluation used its error-removal variant. The method is evaluated only in the known scalar scenario as the degradation operator requires $\alpha$ to be known in advance, preventing adaptation to unknown or spatially-varying $\alpha$ scenarios.
\subsection{Implementation Details}
\label{appendix:implementation_details}
This appendix provides complete implementation specifications for our two-stage $\alpha$-deblending pipeline. Our method operates identically across all three experimental scenarios (known scalar ($\alpha = 0.5$), unknown scalar ($\alpha \in [0.2, 0.8]$), and spatially-varying $\alpha$-maps), with the only difference being whether the $\alpha$ parameters are fixed or optimized during sampling.

\paragraph{Stage 1: Latent Space Decomposition.}
We use pretrained Stable Diffusion v1.4, which operates on $64 \times 64$ latent representations with a VAE downsampling factor of 8 for $512 \times 512$ pixel images. The blended observation $y$ is encoded via the VAE encoder and duplicated for both branches: $z^1_0 = z^2_0 = \mathcal{E}(y)$. Each branch is independently noised to timestep $t_s$ using different noises $\epsilon_k \sim \mathcal{N}(0, I), k \in \{1,2\}$.

We employ DDIM sampling with $T_s$ total timesteps starting from timestep $t_s$. For the known scalar scenario, we use $T_s = 80$ and $t_s = 64$. For scenarios where $\alpha$ is optimized (unknown scalar and spatially-varying), we use $T_s = 100$ and $t_s = 80$ to allow more denoising steps. We apply classifier-free guidance with CFG-scale $\omega = 2.0$.

Data consistency is enforced through multi-scale posterior sampling operating at three spatial scales $\{1.0, 0.5, 0.25\}$ with corresponding loss weights $\{0.25, 0.5, 1.0\}$. At each scale, the reconstruction loss combines L2 norm and LPIPS perceptual loss (VGG-based) weighted by $\lambda_{\text{LPIPS}} = 100$. At each diffusion timestep, we perform $n_{\text{rec}} = 8$ recurrence iterations (ablated in \cref{appendix:recurrence_ablation}), applying gradient-based corrections to the predicted noise $\epsilon_\theta$ only on even-numbered recurrences. The guidance strength follows a three-stage linear schedule: $\gamma_{\text{refine}}=50.0$ for $t < 300$, interpolating to $\gamma_{\text{critical}}=30.0$ at $t=300$, then to $\gamma_{\text{chaotic}}=10.0$ at $t=800$, remaining constant for $t > 800$. This schedule uses stronger guidance at low noise levels (refinement stage) and weaker guidance at high noise levels to avoid artifacts during early denoising, consistent with universal guidance principles.

For each blended image, we generate $N_{\text{cand}} = 3$ candidate decompositions using different random seeds. Because the two output branches are unordered, we score each candidate using the better of the two prompt--image assignments and then select the highest-scoring candidate: $k^*=\arg\max_k\max_{\pi\in S_2}\frac{1}{2}\sum_{i=1}^{2}\operatorname{PickScore}(\tilde{x}^{\pi(i)}_k,c_i^+)$. Selection uses only the Stage~1 outputs and input prompts, without ground-truth images. The selected pair/sample identifier and random seed are preserved in Stage~2, whose declared inputs are required to match the selected Stage~1 output paths exactly.

\paragraph{Stage 2: Pixel Space Refinement.}
We use an unconditional DDPM model from OpenAI's guided-diffusion~\cite{dhariwal2021diffusion}, trained on ImageNet at $256 \times 256$ resolution. The model employs a UNet architecture with 256 base channels, 2 residual blocks per resolution, attention at resolutions $\{32, 16, 8\}$, and 4 attention heads with 64 channels per head. The model predicts both mean and variance ($\Sigma_\theta$) of the reverse process.

We employ 600 evenly-spaced timesteps from the original 1000-step diffusion process using DDIM sampling with $\eta = 0.0$. The selected Stage 1 outputs $\{\hat{x}^1_0, \hat{x}^2_0\}$ are noised to timestep $t_r = 350$ via $x^k_{t_r} = \sqrt{\bar{\alpha}_{t_r}} \hat{x}^k_0 + \sqrt{1-\bar{\alpha}_{t_r}} \epsilon_k$, where $\epsilon_k \sim \mathcal{N}(0, I), k \in \{1,2\}$ are independently sampled. We then denoise from $t_r = 350$ to $t=0$. Starting at $t_r = 350$ of the 600-step schedule, a lower relative noise level than Stage 1 ($350/600$ versus $t_s/T_s=0.8$), ensures we preserve semantic information from Stage 1 while enabling pixel-level refinement. In all scenarios $\alpha$ is kept fixed in Stage 2: the known scalar remains $\alpha=0.5$, and for the optimized scalar and spatially-varying scenarios the $\alpha$ parameters estimated in Stage 1 are used without further optimization.

Data consistency is enforced using the same multi-scale posterior sampling approach as Stage 1, operating directly in pixel space. The multi-scale loss combines L2 reconstruction and LPIPS perceptual loss with $\lambda_{\text{LPIPS}} = 20$ (reduced from Stage 1's 100 to prevent over-correction in pixel space), using the same three scales $\{1.0, 0.5, 0.25\}$ and weights $\{0.25, 0.5, 1.0\}$. We maintain $n_{\text{rec}} = 8$ recurrence iterations per timestep with corrections applied on even-numbered iterations. Between recurrences (except the last), we apply Tweedie-inspired re-noising: $x_t = \sqrt{1-\beta_t} \cdot x_{t-1} + \sqrt{\beta_t} \cdot \epsilon$, where $\beta_t$ is the noise schedule and $\epsilon \sim \mathcal{N}(0, I)$.

The guidance scales are significantly reduced compared to Stage 1: $\gamma_{\text{refine}}=12.0$ for $t < 300$, interpolating to $\gamma_{\text{critical}}=5.0$ at $t=300$, then to $\gamma_{\text{chaotic}}=0.25$ at $t=800$, remaining constant for $t > 800$. These lower scales maintain data consistency while preventing over-correction artifacts in pixel space. Stage 2 uses an unconditional diffusion model without classifier-free guidance or text prompts: the semantic structure established in Stage 1 is preserved through initialization at $t_r = 350$, while the unconditional prior focuses on recovering realistic high-frequency details.

\paragraph{Computational Requirements.}
All experiments are conducted on a single NVIDIA GeForce RTX 3090 using PyTorch with mixed-precision training where applicable. Across the six evaluated settings, the recorded sampling metadata gives 32.1--53.3 minutes to generate all three Stage~1 candidates and 17.8--19.6 minutes to refine the selected candidate in Stage~2, for a canonical sampling total of 50.6--71.0 minutes per input. These totals exclude the unrecorded selector overhead. Random seeds are set for reproducibility, with different seeds used for each candidate generation in Stage~1.

\subsection{Additional Qualitative Results}
\label{appendix:additional_results}

\begin{figure*}[p]
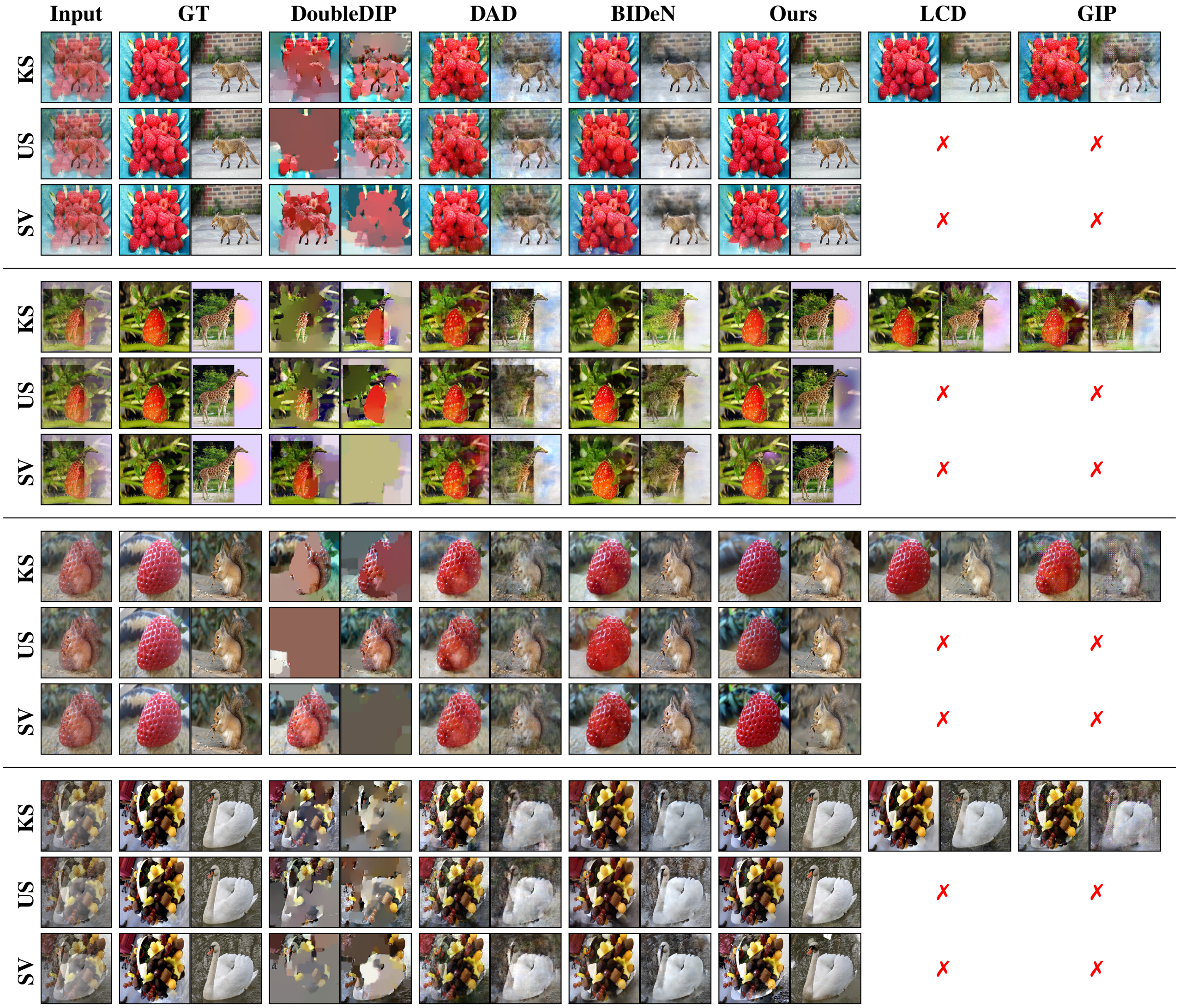
 %
\centering
\small
\renewcommand{\arraystretch}{0.5}

\newcommand{\setOneA}{redberries}
\newcommand{\setOneB}{fox}
\newcommand{\mixOne}{\setOneA-\setOneB}

\newcommand{\setTwoA}{strawberry}
\newcommand{\setTwoB}{girrafe} %
\newcommand{\mixTwo}{\setTwoA-\setTwoB}

\newcommand{\setThreeA}{straweberry2} %
\newcommand{\setThreeB}{squirrel}
\newcommand{\mixThree}{\setThreeA-\setThreeB}

\newcommand{\setFourA}{fruit}
\newcommand{\setFourB}{duck}
\newcommand{\mixFour}{\setFourA-\setFourB}

\newcommand{\framedimage}[1]{\includegraphics[width=0.057\textwidth, valign=m, frame]{#1}}

\newcommand{\vertlabel}[1]{\rotatebox[origin=c]{90}{\textbf{#1}}}

\newcommand{\redx}{\multicolumn{2}{c}{\centering\textcolor{red}{\adjustbox{valign=m}{\ding{55}}}}}

\begin{tabular}{
  c @{\hskip 0.1cm} %
  c @{\hskip 0.1cm} %
  c@{}c @{\hskip 0.1cm} %
  c@{}c @{\hskip 0.1cm} %
  c@{}c @{\hskip 0.1cm} %
  c@{}c @{\hskip 0.1cm} %
  c@{}c @{\hskip 0.1cm} %
  c@{}c @{\hskip 0.1cm} %
  c@{}c             %
}

& \textbf{Input} & \multicolumn{2}{c}{\textbf{GT}} & \multicolumn{2}{c}{\textbf{DoubleDIP}} & \multicolumn{2}{c}{\textbf{DAD}} & \multicolumn{2}{c}{\textbf{BIDeN}} & \multicolumn{2}{c}{\textbf{Ours}} & \multicolumn{2}{c}{\textbf{LCD}} & \multicolumn{2}{c}{\textbf{GIP}} \\[0.1cm]

\vertlabel{KS}
  & \framedimage{figures/4_experiments/baseline_evaluation/scenario1_images/input/\mixOne.png}
  & \framedimage{figures/4_experiments/baseline_evaluation/scenario1_images/ground_truth/\setOneA.png}
  & \framedimage{figures/4_experiments/baseline_evaluation/scenario1_images/ground_truth/\setOneB.png}
  & \framedimage{figures/4_experiments/baseline_evaluation/scenario1_images/DoubleDIP/\setOneA.png}
  & \framedimage{figures/4_experiments/baseline_evaluation/scenario1_images/DoubleDIP/\setOneB.png}
  & \framedimage{figures/4_experiments/baseline_evaluation/scenario1_images/DAD/\setOneA.png}
  & \framedimage{figures/4_experiments/baseline_evaluation/scenario1_images/DAD/\setOneB.png}
  & \framedimage{figures/4_experiments/baseline_evaluation/scenario1_images/BIDeN/\setOneA.png}
  & \framedimage{figures/4_experiments/baseline_evaluation/scenario1_images/BIDeN/\setOneB.png}
  & \framedimage{figures/4_experiments/baseline_evaluation/scenario1_images/Ours/\setOneA.png}
  & \framedimage{figures/4_experiments/baseline_evaluation/scenario1_images/Ours/\setOneB.png}
  & \framedimage{figures/4_experiments/baseline_evaluation/scenario1_images/LCD/\setOneA.png}
  & \framedimage{figures/4_experiments/baseline_evaluation/scenario1_images/LCD/\setOneB.png}
  & \framedimage{figures/4_experiments/baseline_evaluation/scenario1_images/GIP/\setOneA.png}
  & \framedimage{figures/4_experiments/baseline_evaluation/scenario1_images/GIP/\setOneB.png} \\
  \noalign{\vspace{2pt}}

\vertlabel{US}
  & \framedimage{figures/4_experiments/baseline_evaluation/scenario2_images/input/\mixOne.png}
  & \framedimage{figures/4_experiments/baseline_evaluation/scenario2_images/ground_truth/\setOneA.png}
  & \framedimage{figures/4_experiments/baseline_evaluation/scenario2_images/ground_truth/\setOneB.png}
  & \framedimage{figures/4_experiments/baseline_evaluation/scenario2_images/DoubleDIP/\setOneA.png}
  & \framedimage{figures/4_experiments/baseline_evaluation/scenario2_images/DoubleDIP/\setOneB.png}
  & \framedimage{figures/4_experiments/baseline_evaluation/scenario2_images/DAD/\setOneA.png}
  & \framedimage{figures/4_experiments/baseline_evaluation/scenario2_images/DAD/\setOneB.png}
  & \framedimage{figures/4_experiments/baseline_evaluation/scenario2_images/BIDeN/\setOneA.png}
  & \framedimage{figures/4_experiments/baseline_evaluation/scenario2_images/BIDeN/\setOneB.png}
  & \framedimage{figures/4_experiments/baseline_evaluation/scenario2_images/Ours/\setOneA.png}
  & \framedimage{figures/4_experiments/baseline_evaluation/scenario2_images/Ours/\setOneB.png}
  & \redx & \redx \\
  \noalign{\vspace{2pt}}

\vertlabel{SV}
  & \framedimage{figures/4_experiments/baseline_evaluation/scenario3_images/input/\mixOne.png}
  & \framedimage{figures/4_experiments/baseline_evaluation/scenario3_images/ground_truth/\setOneA.png}
  & \framedimage{figures/4_experiments/baseline_evaluation/scenario3_images/ground_truth/\setOneB.png}
  & \framedimage{figures/4_experiments/baseline_evaluation/scenario3_images/DoubleDIP/\setOneA.png}
  & \framedimage{figures/4_experiments/baseline_evaluation/scenario3_images/DoubleDIP/\setOneB.png}
  & \framedimage{figures/4_experiments/baseline_evaluation/scenario3_images/DAD/\setOneA.png}
  & \framedimage{figures/4_experiments/baseline_evaluation/scenario3_images/DAD/\setOneB.png}
  & \framedimage{figures/4_experiments/baseline_evaluation/scenario3_images/BIDeN/\setOneA.png}
  & \framedimage{figures/4_experiments/baseline_evaluation/scenario3_images/BIDeN/\setOneB.png}
  & \framedimage{figures/4_experiments/baseline_evaluation/scenario3_images/Ours/\setOneA.png}
  & \framedimage{figures/4_experiments/baseline_evaluation/scenario3_images/Ours/\setOneB.png}
  & \redx & \redx \\
  \noalign{\vspace{5pt}} \hline \noalign{\vspace{5pt}}

\vertlabel{KS}
  & \framedimage{figures/4_experiments/baseline_evaluation/scenario1_images/input/\mixTwo.png}
  & \framedimage{figures/4_experiments/baseline_evaluation/scenario1_images/ground_truth/\setTwoA.png}
  & \framedimage{figures/4_experiments/baseline_evaluation/scenario1_images/ground_truth/\setTwoB.png}
  & \framedimage{figures/4_experiments/baseline_evaluation/scenario1_images/DoubleDIP/\setTwoA.png}
  & \framedimage{figures/4_experiments/baseline_evaluation/scenario1_images/DoubleDIP/\setTwoB.png}
  & \framedimage{figures/4_experiments/baseline_evaluation/scenario1_images/DAD/\setTwoA.png}
  & \framedimage{figures/4_experiments/baseline_evaluation/scenario1_images/DAD/\setTwoB.png}
  & \framedimage{figures/4_experiments/baseline_evaluation/scenario1_images/BIDeN/\setTwoA.png}
  & \framedimage{figures/4_experiments/baseline_evaluation/scenario1_images/BIDeN/\setTwoB.png}
  & \framedimage{figures/4_experiments/baseline_evaluation/scenario1_images/Ours/\setTwoA.png}
  & \framedimage{figures/4_experiments/baseline_evaluation/scenario1_images/Ours/\setTwoB.png}
  & \framedimage{figures/4_experiments/baseline_evaluation/scenario1_images/LCD/\setTwoA.png}
  & \framedimage{figures/4_experiments/baseline_evaluation/scenario1_images/LCD/\setTwoB.png}
  & \framedimage{figures/4_experiments/baseline_evaluation/scenario1_images/GIP/\setTwoA.png}
  & \framedimage{figures/4_experiments/baseline_evaluation/scenario1_images/GIP/\setTwoB.png} \\
  \noalign{\vspace{2pt}}

\vertlabel{US}
  & \framedimage{figures/4_experiments/baseline_evaluation/scenario2_images/input/\mixTwo.png}
  & \framedimage{figures/4_experiments/baseline_evaluation/scenario2_images/ground_truth/\setTwoA.png}
  & \framedimage{figures/4_experiments/baseline_evaluation/scenario2_images/ground_truth/\setTwoB.png}
  & \framedimage{figures/4_experiments/baseline_evaluation/scenario2_images/DoubleDIP/\setTwoA.png}
  & \framedimage{figures/4_experiments/baseline_evaluation/scenario2_images/DoubleDIP/\setTwoB.png}
  & \framedimage{figures/4_experiments/baseline_evaluation/scenario2_images/DAD/\setTwoA.png}
  & \framedimage{figures/4_experiments/baseline_evaluation/scenario2_images/DAD/\setTwoB.png}
  & \framedimage{figures/4_experiments/baseline_evaluation/scenario2_images/BIDeN/\setTwoA.png}
  & \framedimage{figures/4_experiments/baseline_evaluation/scenario2_images/BIDeN/\setTwoB.png}
  & \framedimage{figures/4_experiments/baseline_evaluation/scenario2_images/Ours/\setTwoA.png}
  & \framedimage{figures/4_experiments/baseline_evaluation/scenario2_images/Ours/\setTwoB.png}
  & \redx & \redx \\
  \noalign{\vspace{2pt}}

\vertlabel{SV}
  & \framedimage{figures/4_experiments/baseline_evaluation/scenario3_images/input/\mixTwo.png}
  & \framedimage{figures/4_experiments/baseline_evaluation/scenario3_images/ground_truth/\setTwoA.png}
  & \framedimage{figures/4_experiments/baseline_evaluation/scenario3_images/ground_truth/\setTwoB.png}
  & \framedimage{figures/4_experiments/baseline_evaluation/scenario3_images/DoubleDIP/\setTwoA.png}
  & \framedimage{figures/4_experiments/baseline_evaluation/scenario3_images/DoubleDIP/\setTwoB.png}
  & \framedimage{figures/4_experiments/baseline_evaluation/scenario3_images/DAD/\setTwoA.png}
  & \framedimage{figures/4_experiments/baseline_evaluation/scenario3_images/DAD/\setTwoB.png}
  & \framedimage{figures/4_experiments/baseline_evaluation/scenario3_images/BIDeN/\setTwoA.png}
  & \framedimage{figures/4_experiments/baseline_evaluation/scenario3_images/BIDeN/\setTwoB.png}
  & \framedimage{figures/4_experiments/baseline_evaluation/scenario3_images/Ours/\setTwoA.png}
  & \framedimage{figures/4_experiments/baseline_evaluation/scenario3_images/Ours/\setTwoB.png}
  & \redx & \redx \\
  \noalign{\vspace{5pt}} \hline \noalign{\vspace{5pt}}

\vertlabel{KS}
  & \framedimage{figures/4_experiments/baseline_evaluation/scenario1_images/input/\mixThree.png}
  & \framedimage{figures/4_experiments/baseline_evaluation/scenario1_images/ground_truth/\setThreeA.png}
  & \framedimage{figures/4_experiments/baseline_evaluation/scenario1_images/ground_truth/\setThreeB.png}
  & \framedimage{figures/4_experiments/baseline_evaluation/scenario1_images/DoubleDIP/\setThreeA.png}
  & \framedimage{figures/4_experiments/baseline_evaluation/scenario1_images/DoubleDIP/\setThreeB.png}
  & \framedimage{figures/4_experiments/baseline_evaluation/scenario1_images/DAD/\setThreeA.png}
  & \framedimage{figures/4_experiments/baseline_evaluation/scenario1_images/DAD/\setThreeB.png}
  & \framedimage{figures/4_experiments/baseline_evaluation/scenario1_images/BIDeN/\setThreeA.png}
  & \framedimage{figures/4_experiments/baseline_evaluation/scenario1_images/BIDeN/\setThreeB.png}
  & \framedimage{figures/4_experiments/baseline_evaluation/scenario1_images/Ours/\setThreeA.png}
  & \framedimage{figures/4_experiments/baseline_evaluation/scenario1_images/Ours/\setThreeB.png}
  & \framedimage{figures/4_experiments/baseline_evaluation/scenario1_images/LCD/\setThreeA.png}
  & \framedimage{figures/4_experiments/baseline_evaluation/scenario1_images/LCD/\setThreeB.png}
  & \framedimage{figures/4_experiments/baseline_evaluation/scenario1_images/GIP/\setThreeA.png}
  & \framedimage{figures/4_experiments/baseline_evaluation/scenario1_images/GIP/\setThreeB.png} \\
  \noalign{\vspace{2pt}}

\vertlabel{US}
  & \framedimage{figures/4_experiments/baseline_evaluation/scenario2_images/input/\mixThree.png}
  & \framedimage{figures/4_experiments/baseline_evaluation/scenario2_images/ground_truth/\setThreeA.png}
  & \framedimage{figures/4_experiments/baseline_evaluation/scenario2_images/ground_truth/\setThreeB.png}
  & \framedimage{figures/4_experiments/baseline_evaluation/scenario2_images/DoubleDIP/\setThreeA.png}
  & \framedimage{figures/4_experiments/baseline_evaluation/scenario2_images/DoubleDIP/\setThreeB.png}
  & \framedimage{figures/4_experiments/baseline_evaluation/scenario2_images/DAD/\setThreeA.png}
  & \framedimage{figures/4_experiments/baseline_evaluation/scenario2_images/DAD/\setThreeB.png}
  & \framedimage{figures/4_experiments/baseline_evaluation/scenario2_images/BIDeN/\setThreeA.png}
  & \framedimage{figures/4_experiments/baseline_evaluation/scenario2_images/BIDeN/\setThreeB.png}
  & \framedimage{figures/4_experiments/baseline_evaluation/scenario2_images/Ours/\setThreeA.png}
  & \framedimage{figures/4_experiments/baseline_evaluation/scenario2_images/Ours/\setThreeB.png}
  & \redx & \redx \\
  \noalign{\vspace{2pt}}

\vertlabel{SV}
  & \framedimage{figures/4_experiments/baseline_evaluation/scenario3_images/input/\mixThree.png}
  & \framedimage{figures/4_experiments/baseline_evaluation/scenario3_images/ground_truth/\setThreeA.png}
  & \framedimage{figures/4_experiments/baseline_evaluation/scenario3_images/ground_truth/\setThreeB.png}
  & \framedimage{figures/4_experiments/baseline_evaluation/scenario3_images/DoubleDIP/\setThreeA.png}
  & \framedimage{figures/4_experiments/baseline_evaluation/scenario3_images/DoubleDIP/\setThreeB.png}
  & \framedimage{figures/4_experiments/baseline_evaluation/scenario3_images/DAD/\setThreeA.png}
  & \framedimage{figures/4_experiments/baseline_evaluation/scenario3_images/DAD/\setThreeB.png}
  & \framedimage{figures/4_experiments/baseline_evaluation/scenario3_images/BIDeN/\setThreeA.png}
  & \framedimage{figures/4_experiments/baseline_evaluation/scenario3_images/BIDeN/\setThreeB.png}
  & \framedimage{figures/4_experiments/baseline_evaluation/scenario3_images/Ours/\setThreeA.png}
  & \framedimage{figures/4_experiments/baseline_evaluation/scenario3_images/Ours/\setThreeB.png}
  & \redx & \redx \\
  \noalign{\vspace{5pt}} \hline \noalign{\vspace{5pt}}

\vertlabel{KS}
  & \framedimage{figures/4_experiments/baseline_evaluation/scenario1_images/input/\mixFour.png}
  & \framedimage{figures/4_experiments/baseline_evaluation/scenario1_images/ground_truth/\setFourA.png}
  & \framedimage{figures/4_experiments/baseline_evaluation/scenario1_images/ground_truth/\setFourB.png}
  & \framedimage{figures/4_experiments/baseline_evaluation/scenario1_images/DoubleDIP/\setFourA.png}
  & \framedimage{figures/4_experiments/baseline_evaluation/scenario1_images/DoubleDIP/\setFourB.png}
  & \framedimage{figures/4_experiments/baseline_evaluation/scenario1_images/DAD/\setFourA.png}
  & \framedimage{figures/4_experiments/baseline_evaluation/scenario1_images/DAD/\setFourB.png}
  & \framedimage{figures/4_experiments/baseline_evaluation/scenario1_images/BIDeN/\setFourA.png}
  & \framedimage{figures/4_experiments/baseline_evaluation/scenario1_images/BIDeN/\setFourB.png}
  & \framedimage{figures/4_experiments/baseline_evaluation/scenario1_images/Ours/\setFourA.png}
  & \framedimage{figures/4_experiments/baseline_evaluation/scenario1_images/Ours/\setFourB.png}
  & \framedimage{figures/4_experiments/baseline_evaluation/scenario1_images/LCD/\setFourA.png}
  & \framedimage{figures/4_experiments/baseline_evaluation/scenario1_images/LCD/\setFourB.png}
  & \framedimage{figures/4_experiments/baseline_evaluation/scenario1_images/GIP/\setFourA.png}
  & \framedimage{figures/4_experiments/baseline_evaluation/scenario1_images/GIP/\setFourB.png} \\
  \noalign{\vspace{2pt}}

\vertlabel{US}
  & \framedimage{figures/4_experiments/baseline_evaluation/scenario2_images/input/\mixFour.png}
  & \framedimage{figures/4_experiments/baseline_evaluation/scenario2_images/ground_truth/\setFourA.png}
  & \framedimage{figures/4_experiments/baseline_evaluation/scenario2_images/ground_truth/\setFourB.png}
  & \framedimage{figures/4_experiments/baseline_evaluation/scenario2_images/DoubleDIP/\setFourA.png}
  & \framedimage{figures/4_experiments/baseline_evaluation/scenario2_images/DoubleDIP/\setFourB.png}
  & \framedimage{figures/4_experiments/baseline_evaluation/scenario2_images/DAD/\setFourA.png}
  & \framedimage{figures/4_experiments/baseline_evaluation/scenario2_images/DAD/\setFourB.png}
  & \framedimage{figures/4_experiments/baseline_evaluation/scenario2_images/BIDeN/\setFourA.png}
  & \framedimage{figures/4_experiments/baseline_evaluation/scenario2_images/BIDeN/\setFourB.png}
  & \framedimage{figures/4_experiments/baseline_evaluation/scenario2_images/Ours/\setFourA.png}
  & \framedimage{figures/4_experiments/baseline_evaluation/scenario2_images/Ours/\setFourB.png}
  & \redx & \redx \\
  \noalign{\vspace{2pt}}

\vertlabel{SV}
  & \framedimage{figures/4_experiments/baseline_evaluation/scenario3_images/input/\mixFour.png}
  & \framedimage{figures/4_experiments/baseline_evaluation/scenario3_images/ground_truth/\setFourA.png}
  & \framedimage{figures/4_experiments/baseline_evaluation/scenario3_images/ground_truth/\setFourB.png}
  & \framedimage{figures/4_experiments/baseline_evaluation/scenario3_images/DoubleDIP/\setFourA.png}
  & \framedimage{figures/4_experiments/baseline_evaluation/scenario3_images/DoubleDIP/\setFourB.png}
  & \framedimage{figures/4_experiments/baseline_evaluation/scenario3_images/DAD/\setFourA.png}
  & \framedimage{figures/4_experiments/baseline_evaluation/scenario3_images/DAD/\setFourB.png}
  & \framedimage{figures/4_experiments/baseline_evaluation/scenario3_images/BIDeN/\setFourA.png}
  & \framedimage{figures/4_experiments/baseline_evaluation/scenario3_images/BIDeN/\setFourB.png}
  & \framedimage{figures/4_experiments/baseline_evaluation/scenario3_images/Ours/\setFourA.png}
  & \framedimage{figures/4_experiments/baseline_evaluation/scenario3_images/Ours/\setFourB.png}
  & \redx & \redx \\

\end{tabular}

\caption{\textbf{Qualitative comparison on seen domains.} Results for \textit{Fruits vs. Animals} across mixing scenarios: Known Scalar (\textbf{KS}), Unknown Scalar (\textbf{US}), and Spatially Varying (\textbf{SV}). The four blocks represent distinct test cases: \textit{Redberries--Fox}, \textit{Strawberry--Giraffe}, \textit{Strawberry--Squirrel}, and \textit{Fruit--Duck}. Each block is one blended pair.}
\label{fig:combined-seen-domains}
\end{figure*}
\begin{figure*}[p]
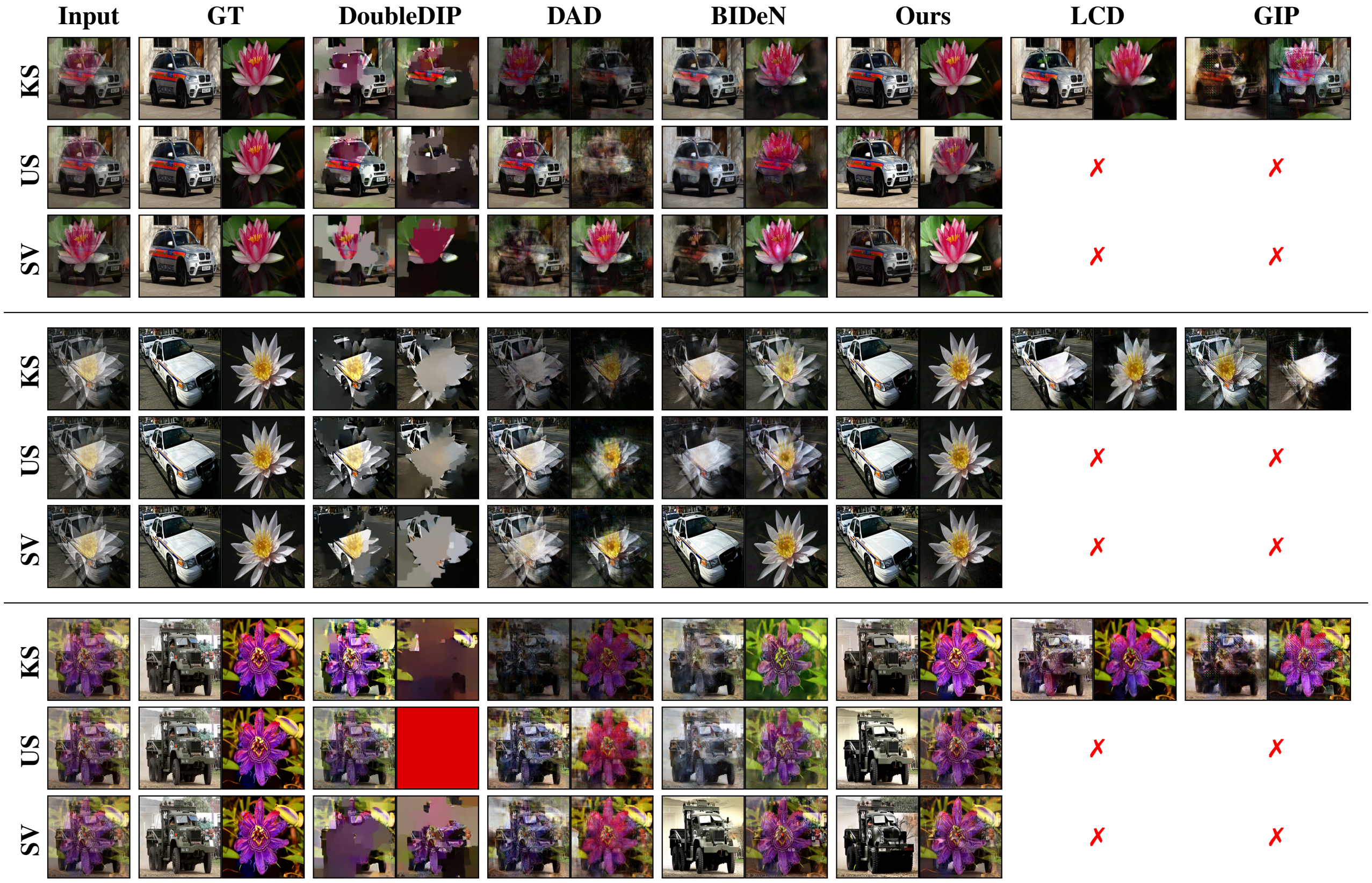
 %
\centering
\small
\renewcommand{\arraystretch}{0.5}

\newcommand{\setOneA}{gray-car}
\newcommand{\setOneB}{pink-flower}
\newcommand{\mixOne}{\setOneA-\setOneB}

\newcommand{\setTwoA}{police}
\newcommand{\setTwoB}{wflower}
\newcommand{\mixTwo}{\setTwoA-\setTwoB}

\newcommand{\setThreeA}{tank}
\newcommand{\setThreeB}{pflower}
\newcommand{\mixThree}{\setThreeA-\setThreeB}

\newcommand{\framedimage}[1]{\includegraphics[width=0.057\textwidth, valign=m, frame]{#1}}

\newcommand{\vertlabel}[1]{\rotatebox[origin=c]{90}{\textbf{#1}}}

\newcommand{\redx}{\multicolumn{2}{c}{\centering\textcolor{red}{\adjustbox{valign=m}{\ding{55}}}}}

\begin{tabular}{
  c @{\hskip 0.1cm} %
  c @{\hskip 0.1cm} %
  c@{}c @{\hskip 0.1cm} %
  c@{}c @{\hskip 0.1cm} %
  c@{}c @{\hskip 0.1cm} %
  c@{}c @{\hskip 0.1cm} %
  c@{}c @{\hskip 0.1cm} %
  c@{}c @{\hskip 0.1cm} %
  c@{}c             %
}

& \textbf{Input} & \multicolumn{2}{c}{\textbf{GT}} & \multicolumn{2}{c}{\textbf{DoubleDIP}} & \multicolumn{2}{c}{\textbf{DAD}} & \multicolumn{2}{c}{\textbf{BIDeN}} & \multicolumn{2}{c}{\textbf{Ours}} & \multicolumn{2}{c}{\textbf{LCD}} & \multicolumn{2}{c}{\textbf{GIP}} \\[0.1cm]

\vertlabel{KS}
  & \framedimage{figures/4_experiments/generalization/scenario1_images/input/\mixOne.png}
  & \framedimage{figures/4_experiments/generalization/scenario1_images/ground_truth/\setOneA.png}
  & \framedimage{figures/4_experiments/generalization/scenario1_images/ground_truth/\setOneB.png}
  & \framedimage{figures/4_experiments/generalization/scenario1_images/DoubleDIP/\setOneA.png}
  & \framedimage{figures/4_experiments/generalization/scenario1_images/DoubleDIP/\setOneB.png}
  & \framedimage{figures/4_experiments/generalization/scenario1_images/DAD/\setOneA.png}
  & \framedimage{figures/4_experiments/generalization/scenario1_images/DAD/\setOneB.png}
  & \framedimage{figures/4_experiments/generalization/scenario1_images/BIDeN/\setOneA.png}
  & \framedimage{figures/4_experiments/generalization/scenario1_images/BIDeN/\setOneB.png}
  & \framedimage{figures/4_experiments/generalization/scenario1_images/Ours/\setOneA.png}
  & \framedimage{figures/4_experiments/generalization/scenario1_images/Ours/\setOneB.png}
  & \framedimage{figures/4_experiments/generalization/scenario1_images/LCD/\setOneA.png}
  & \framedimage{figures/4_experiments/generalization/scenario1_images/LCD/\setOneB.png}
  & \framedimage{figures/4_experiments/generalization/scenario1_images/GIP/\setOneA.png}
  & \framedimage{figures/4_experiments/generalization/scenario1_images/GIP/\setOneB.png} \\
  \noalign{\vspace{2pt}}

\vertlabel{US}
  & \framedimage{figures/4_experiments/generalization/scenario2_images/input/\mixOne.png}
  & \framedimage{figures/4_experiments/generalization/scenario2_images/ground_truth/\setOneA.png}
  & \framedimage{figures/4_experiments/generalization/scenario2_images/ground_truth/\setOneB.png}
  & \framedimage{figures/4_experiments/generalization/scenario2_images/DoubleDIP/\setOneA.png}
  & \framedimage{figures/4_experiments/generalization/scenario2_images/DoubleDIP/\setOneB.png}
  & \framedimage{figures/4_experiments/generalization/scenario2_images/DAD/\setOneA.png}
  & \framedimage{figures/4_experiments/generalization/scenario2_images/DAD/\setOneB.png}
  & \framedimage{figures/4_experiments/generalization/scenario2_images/BIDeN/\setOneA.png}
  & \framedimage{figures/4_experiments/generalization/scenario2_images/BIDeN/\setOneB.png}
  & \framedimage{figures/4_experiments/generalization/scenario2_images/Ours/\setOneA.png}
  & \framedimage{figures/4_experiments/generalization/scenario2_images/Ours/\setOneB.png}
  & \redx & \redx \\
  \noalign{\vspace{2pt}}

\vertlabel{SV}
  & \framedimage{figures/4_experiments/generalization/scenario3_images/input/\mixOne.png}
  & \framedimage{figures/4_experiments/generalization/scenario3_images/ground_truth/\setOneA.png}
  & \framedimage{figures/4_experiments/generalization/scenario3_images/ground_truth/\setOneB.png}
  & \framedimage{figures/4_experiments/generalization/scenario3_images/DoubleDIP/\setOneA.png}
  & \framedimage{figures/4_experiments/generalization/scenario3_images/DoubleDIP/\setOneB.png}
  & \framedimage{figures/4_experiments/generalization/scenario3_images/DAD/\setOneA.png}
  & \framedimage{figures/4_experiments/generalization/scenario3_images/DAD/\setOneB.png}
  & \framedimage{figures/4_experiments/generalization/scenario3_images/BIDeN/\setOneA.png}
  & \framedimage{figures/4_experiments/generalization/scenario3_images/BIDeN/\setOneB.png}
  & \framedimage{figures/4_experiments/generalization/scenario3_images/Ours/\setOneA.png}
  & \framedimage{figures/4_experiments/generalization/scenario3_images/Ours/\setOneB.png}
  & \redx & \redx \\
  \noalign{\vspace{5pt}} \hline \noalign{\vspace{5pt}}

\vertlabel{KS}
  & \framedimage{figures/4_experiments/generalization/scenario1_images/input/\mixTwo.png}
  & \framedimage{figures/4_experiments/generalization/scenario1_images/ground_truth/\setTwoA.png}
  & \framedimage{figures/4_experiments/generalization/scenario1_images/ground_truth/\setTwoB.png}
  & \framedimage{figures/4_experiments/generalization/scenario1_images/DoubleDIP/\setTwoA.png}
  & \framedimage{figures/4_experiments/generalization/scenario1_images/DoubleDIP/\setTwoB.png}
  & \framedimage{figures/4_experiments/generalization/scenario1_images/DAD/\setTwoA.png}
  & \framedimage{figures/4_experiments/generalization/scenario1_images/DAD/\setTwoB.png}
  & \framedimage{figures/4_experiments/generalization/scenario1_images/BIDeN/\setTwoA.png}
  & \framedimage{figures/4_experiments/generalization/scenario1_images/BIDeN/\setTwoB.png}
  & \framedimage{figures/4_experiments/generalization/scenario1_images/Ours/\setTwoA.png}
  & \framedimage{figures/4_experiments/generalization/scenario1_images/Ours/\setTwoB.png}
  & \framedimage{figures/4_experiments/generalization/scenario1_images/LCD/\setTwoA.png}
  & \framedimage{figures/4_experiments/generalization/scenario1_images/LCD/\setTwoB.png}
  & \framedimage{figures/4_experiments/generalization/scenario1_images/GIP/\setTwoA.png}
  & \framedimage{figures/4_experiments/generalization/scenario1_images/GIP/\setTwoB.png} \\
  \noalign{\vspace{2pt}}

\vertlabel{US}
  & \framedimage{figures/4_experiments/generalization/scenario2_images/input/\mixTwo.png}
  & \framedimage{figures/4_experiments/generalization/scenario2_images/ground_truth/\setTwoA.png}
  & \framedimage{figures/4_experiments/generalization/scenario2_images/ground_truth/\setTwoB.png}
  & \framedimage{figures/4_experiments/generalization/scenario2_images/DoubleDIP/\setTwoA.png}
  & \framedimage{figures/4_experiments/generalization/scenario2_images/DoubleDIP/\setTwoB.png}
  & \framedimage{figures/4_experiments/generalization/scenario2_images/DAD/\setTwoA.png}
  & \framedimage{figures/4_experiments/generalization/scenario2_images/DAD/\setTwoB.png}
  & \framedimage{figures/4_experiments/generalization/scenario2_images/BIDeN/\setTwoA.png}
  & \framedimage{figures/4_experiments/generalization/scenario2_images/BIDeN/\setTwoB.png}
  & \framedimage{figures/4_experiments/generalization/scenario2_images/Ours/\setTwoA.png}
  & \framedimage{figures/4_experiments/generalization/scenario2_images/Ours/\setTwoB.png}
  & \redx & \redx \\
  \noalign{\vspace{2pt}}

\vertlabel{SV}
  & \framedimage{figures/4_experiments/generalization/scenario3_images/input/\mixTwo.png}
  & \framedimage{figures/4_experiments/generalization/scenario3_images/ground_truth/\setTwoA.png}
  & \framedimage{figures/4_experiments/generalization/scenario3_images/ground_truth/\setTwoB.png}
  & \framedimage{figures/4_experiments/generalization/scenario3_images/DoubleDIP/\setTwoA.png}
  & \framedimage{figures/4_experiments/generalization/scenario3_images/DoubleDIP/\setTwoB.png}
  & \framedimage{figures/4_experiments/generalization/scenario3_images/DAD/\setTwoA.png}
  & \framedimage{figures/4_experiments/generalization/scenario3_images/DAD/\setTwoB.png}
  & \framedimage{figures/4_experiments/generalization/scenario3_images/BIDeN/\setTwoA.png}
  & \framedimage{figures/4_experiments/generalization/scenario3_images/BIDeN/\setTwoB.png}
  & \framedimage{figures/4_experiments/generalization/scenario3_images/Ours/\setTwoA.png}
  & \framedimage{figures/4_experiments/generalization/scenario3_images/Ours/\setTwoB.png}
  & \redx & \redx \\
  \noalign{\vspace{5pt}} \hline \noalign{\vspace{5pt}}

\vertlabel{KS}
  & \framedimage{figures/4_experiments/generalization/scenario1_images/input/\mixThree.png}
  & \framedimage{figures/4_experiments/generalization/scenario1_images/ground_truth/\setThreeA.png}
  & \framedimage{figures/4_experiments/generalization/scenario1_images/ground_truth/\setThreeB.png}
  & \framedimage{figures/4_experiments/generalization/scenario1_images/DoubleDIP/\setThreeA.png}
  & \framedimage{figures/4_experiments/generalization/scenario1_images/DoubleDIP/\setThreeB.png}
  & \framedimage{figures/4_experiments/generalization/scenario1_images/DAD/\setThreeA.png}
  & \framedimage{figures/4_experiments/generalization/scenario1_images/DAD/\setThreeB.png}
  & \framedimage{figures/4_experiments/generalization/scenario1_images/BIDeN/\setThreeA.png}
  & \framedimage{figures/4_experiments/generalization/scenario1_images/BIDeN/\setThreeB.png}
  & \framedimage{figures/4_experiments/generalization/scenario1_images/Ours/\setThreeA.png}
  & \framedimage{figures/4_experiments/generalization/scenario1_images/Ours/\setThreeB.png}
  & \framedimage{figures/4_experiments/generalization/scenario1_images/LCD/\setThreeA.png}
  & \framedimage{figures/4_experiments/generalization/scenario1_images/LCD/\setThreeB.png}
  & \framedimage{figures/4_experiments/generalization/scenario1_images/GIP/\setThreeA.png}
  & \framedimage{figures/4_experiments/generalization/scenario1_images/GIP/\setThreeB.png} \\
  \noalign{\vspace{2pt}}

\vertlabel{US}
  & \framedimage{figures/4_experiments/generalization/scenario2_images/input/\mixThree.png}
  & \framedimage{figures/4_experiments/generalization/scenario2_images/ground_truth/\setThreeA.png}
  & \framedimage{figures/4_experiments/generalization/scenario2_images/ground_truth/\setThreeB.png}
  & \framedimage{figures/4_experiments/generalization/scenario2_images/DoubleDIP/\setThreeA.png}
  & \framedimage{figures/4_experiments/generalization/scenario2_images/DoubleDIP/\setThreeB.png}
  & \framedimage{figures/4_experiments/generalization/scenario2_images/DAD/\setThreeA.png}
  & \framedimage{figures/4_experiments/generalization/scenario2_images/DAD/\setThreeB.png}
  & \framedimage{figures/4_experiments/generalization/scenario2_images/BIDeN/\setThreeA.png}
  & \framedimage{figures/4_experiments/generalization/scenario2_images/BIDeN/\setThreeB.png}
  & \framedimage{figures/4_experiments/generalization/scenario2_images/Ours/\setThreeA.png}
  & \framedimage{figures/4_experiments/generalization/scenario2_images/Ours/\setThreeB.png}
  & \redx & \redx \\
  \noalign{\vspace{2pt}}

\vertlabel{SV}
  & \framedimage{figures/4_experiments/generalization/scenario3_images/input/\mixThree.png}
  & \framedimage{figures/4_experiments/generalization/scenario3_images/ground_truth/\setThreeA.png}
  & \framedimage{figures/4_experiments/generalization/scenario3_images/ground_truth/\setThreeB.png}
  & \framedimage{figures/4_experiments/generalization/scenario3_images/DoubleDIP/\setThreeA.png}
  & \framedimage{figures/4_experiments/generalization/scenario3_images/DoubleDIP/\setThreeB.png}
  & \framedimage{figures/4_experiments/generalization/scenario3_images/DAD/\setThreeA.png}
  & \framedimage{figures/4_experiments/generalization/scenario3_images/DAD/\setThreeB.png}
  & \framedimage{figures/4_experiments/generalization/scenario3_images/BIDeN/\setThreeA.png}
  & \framedimage{figures/4_experiments/generalization/scenario3_images/BIDeN/\setThreeB.png}
  & \framedimage{figures/4_experiments/generalization/scenario3_images/Ours/\setThreeA.png}
  & \framedimage{figures/4_experiments/generalization/scenario3_images/Ours/\setThreeB.png}
  & \redx & \redx \\

\end{tabular}

\caption{\textbf{Qualitative comparison on unseen domains.} Results for \textit{Vehicles vs. Flowers} across three mixing scenarios: Known Scalar (\textbf{KS}), Unknown Scalar (\textbf{US}), and Spatially Varying (\textbf{SV}). The three blocks represent: \textit{Gray Car--Pink Flower}, \textit{Police--White Flower}, and \textit{Tank--Purple Flower}. Each block is one blended pair; the baselines were not trained on these domains.}
\label{fig:combined-unseen-domains}
\end{figure*}

In this section, we provide additional qualitative comparisons across all three deblending scenarios evaluated in our experiments. Each figure shows results for Scenario 1 (Known Scalar), Scenario 2 (Unknown Scalar), and Scenario 3 (Spatially-Varying) side-by-side, allowing direct comparison of method performance as task difficulty increases. These examples complement the main results by demonstrating consistency across seen domains (\cref{fig:combined-seen-domains}: \textit{Redberries--Fox}, \textit{Strawberry--Giraffe}, \textit{Strawberry--Squirrel}, and \textit{Fruit--Duck}) and unseen domains (\cref{fig:combined-unseen-domains}: \textit{Gray Car--Pink Flower}, \textit{Police--White Flower}, and \textit{Tank--Purple Flower}).
\subsection{Ablation Study}
\label{appendix:ablation_study}
\subsubsection{Text-Guidance Strategies}
\label{appendix:text_guidance}
To evaluate the impact of the text prompts on the separation process, we compare our full method against two alternative configurations:
\begin{enumerate}
    \item \textbf{Ours:} Utilizes $c^k_+$ for positive guidance and the opposing source's description as the negative prompt $c^k_-=c^{j}_+$ where $j \neq k$ \citep{hong2024differ}.
    \item \textbf{Standard CFG:} Uses $c^k_+$ for positive guidance and an empty string for the negative prompt ($c^k_- = \emptyset$). This configuration evaluates whether using the opposing source as a negative prompt (as in \textbf{Ours}) contributes to the deblending process.
    \item \textbf{Class-Level:} Employs \textbf{Ours} strategy but with generic class labels as prompts (e.g., ``fruits'', ``animals'') instead of detailed descriptions. This configuration assesses the importance of prompt granularity and specificity for effective separation.
\end{enumerate}
We perform this ablation under a constrained setting compared to the main experiments due to the computational cost. Specifically, we set the number of candidates to $N_{cand}=1$ and evaluate on a subset of 100 pairs from the \textit{Fruits vs Animals} test set. We test on two of the mixing scenarios: The first is the constant scalar $\alpha=0.5$, and the second is spatially varying $\alpha$ map. We report performance across three CFG scales ($\omega \in \{1.0, 2.0, 3.0\}$) alongside an unconditional baseline, where $\omega=0.0$ and there are no positive or negative prompts.

\begin{table*}[!tp] %
\centering
\scriptsize
\setlength{\tabcolsep}{6pt} %
\caption{Ablation of text-guidance strategies on 100 Fruits/Animals pairs with $N_{\mathrm{cand}}=1$. \textbf{Baseline} is prompt-free sampling ($\omega=0$); \textbf{Ours} uses the opposing source's description as the negative prompt (cross-negative); \textbf{Standard CFG} uses an empty negative prompt; \textbf{Class-Level} uses generic class labels. Sample~1 and Sample~2 are the two recovered sources. Bold marks the best value per scenario and column. Cross-negative prompting is clearly better than standard CFG on spatially varying $\alpha$, while the two are comparable for a known scalar.}
\begin{tabular}{llc ccc ccc}
\toprule
\multicolumn{3}{c}{} & \multicolumn{3}{c}{\textbf{Sample 1}} & \multicolumn{3}{c}{\textbf{Sample 2}} \\
\cmidrule(lr){4-6} \cmidrule(lr){7-9}

\textbf{Scenario} & \textbf{Method} & \textbf{CFG} ($\omega$) & \textbf{PSNR}$\uparrow$ & \textbf{SSIM}$\uparrow$ & \textbf{LPIPS}$\downarrow$ & \textbf{PSNR}$\uparrow$ & \textbf{SSIM}$\uparrow$ & \textbf{LPIPS}$\downarrow$ \\
\midrule

\multirow{10}{*}{\makecell[l]{\textbf{Known}\\\textbf{Scalar}}}
 & \textbf{Baseline} & 0.0 & \num{17.7258} & \num{0.7207} & \num{0.2215} & \num{17.6530} & \num{0.7166} & \num{0.2666} \\
 \cmidrule(lr){2-9}
 & \multirow{3}{*}{\makecell[l]{\textbf{Cross-neg.}\\\textbf{(Ours)}}} & 1.0 & \num{20.1523} & \num{0.7842} & \num{0.1494} & \num{20.0426} & \num{0.7811} & \num{0.1890} \\
 & & 2.0 & \bfseries\num{20.5356} & \num{0.7852} & \num{0.1499} & \bfseries\num{20.3608} & \num{0.7864} & \num{0.1829} \\
 & & 3.0 & \num{19.5746} & \num{0.7800} & \num{0.1564} & \num{19.5845} & \num{0.7787} & \num{0.1946} \\
 \cmidrule(lr){2-9}
 & \multirow{3}{*}{\makecell[l]{Standard\\CFG}} & 1.0 & \num{20.3293} & \bfseries\num{0.7897} & \bfseries\num{0.1447} & \num{20.1637} & \bfseries\num{0.7871} & \bfseries\num{0.1827} \\
 & & 2.0 & \num{19.9885} & \num{0.7819} & \num{0.1581} & \num{19.9775} & \num{0.7827} & \num{0.1886} \\
 & & 3.0 & \num{19.8441} & \num{0.7674} & \num{0.1679} & \num{19.8216} & \num{0.7704} & \num{0.2003} \\
 \cmidrule(lr){2-9}
 & \multirow{3}{*}{Class-Level} & 1.0 & \num{18.5482} & \num{0.7344} & \num{0.2074} & \num{18.4971} & \num{0.7322} & \num{0.2476} \\
 & & 2.0 & \num{18.4040} & \num{0.7524} & \num{0.1788} & \num{18.3596} & \num{0.7528} & \num{0.2213} \\
 & & 3.0 & \num{17.7398} & \num{0.7326} & \num{0.2099} & \num{17.7056} & \num{0.7300} & \num{0.2483} \\
\midrule

\multirow{10}{*}{\makecell[l]{\textbf{Spatially}\\\textbf{Varying}}}
 & \textbf{Baseline} & 0.0 & \num{13.8231} & \num{0.4897} & \num{0.4272} & \num{13.9357} & \num{0.4523} & \num{0.5147} \\
 \cmidrule(lr){2-9}
 & \multirow{3}{*}{\makecell[l]{\textbf{Cross-neg.}\\\textbf{(Ours)}}} & 1.0 & \num{14.6743} & \num{0.5303} & \num{0.3672} & \num{14.9081} & \num{0.4860} & \num{0.4617} \\
 & & 2.0 & \bfseries\num{17.3130} & \bfseries\num{0.7132} & \bfseries\num{0.2180} & \bfseries\num{17.6106} & \bfseries\num{0.7180} & \bfseries\num{0.2424} \\
 & & 3.0 & \num{17.2050} & \num{0.6906} & \num{0.2375} & \num{17.4268} & \num{0.7015} & \num{0.2614} \\
 \cmidrule(lr){2-9}
 & \multirow{3}{*}{\makecell[l]{Standard\\CFG}} & 1.0 & \num{14.8372} & \num{0.5394} & \num{0.3503} & \num{15.0452} & \num{0.4949} & \num{0.4404} \\
 & & 2.0 & \num{15.0206} & \num{0.5472} & \num{0.3635} & \num{15.0007} & \num{0.4895} & \num{0.4486} \\
 & & 3.0 & \num{14.6593} & \num{0.5240} & \num{0.3906} & \num{14.7022} & \num{0.4745} & \num{0.4705} \\
 \cmidrule(lr){2-9}
 & \multirow{3}{*}{Class-Level} & 1.0 & \num{14.1698} & \num{0.5066} & \num{0.3951} & \num{14.3488} & \num{0.4708} & \num{0.4798} \\
 & & 2.0 & \num{14.2485} & \num{0.5171} & \num{0.4049} & \num{14.4379} & \num{0.4727} & \num{0.4761} \\
 & & 3.0 & \num{14.0665} & \num{0.5054} & \num{0.4224} & \num{14.0235} & \num{0.4505} & \num{0.5015} \\
\bottomrule
\end{tabular}

\label{tab:text_guidance_compare}
\end{table*}

The quantitative results in \cref{tab:text_guidance_compare} highlight several trends regarding the role of semantic guidance. First, using text prompts outperforms the unconditional \textbf{Baseline}, demonstrating that while the prior knowledge of the model provides a foundation for separation, explicit semantic direction yields a performance gain. Second, the comparison between detailed and coarse descriptions reveals that prompt detail is important: the \textbf{Class-Level} configuration consistently underperforms compared to using detailed prompts.
Finally, the distinction between strategies becomes most apparent in the complex mixing regime. While \textbf{Standard CFG} remains competitive in the simple \textit{$Known\ Scalar$} scenario, its performance degrades noticeably in the \textit{Spatially\ Varying} setting. In contrast, \textbf{Ours} strategy maintains high separation fidelity in the spatially varying regime. This indicates that explicitly negating the competing signal via $c^k_-$ is beneficial in more complex scenarios. Qualitative results are shown in \cref{fig:combined_text_guidance}.
\begin{figure*}[htbp]
\centering
\small
\renewcommand{\arraystretch}{0.5}

\newcommand{\basepath}{figures/0_appendix/text_guidance/text_images}

\newcommand{\setOneA}{apples}
\newcommand{\setOneB}{cat}
\newcommand{\mixOne}{\setOneA\_\setOneB}

\newcommand{\setTwoA}{fruits}
\newcommand{\setTwoB}{kangaro}
\newcommand{\mixTwo}{\setTwoA\_\setTwoB}

\newcommand{\framedimage}[1]{\includegraphics[width=0.075\textwidth, valign=m, frame]{#1}}

\newcommand{\vertlabel}[1]{\rotatebox[origin=c]{90}{\textbf{#1}}}

\begin{tabular}{
  c @{\hskip 0.1cm} %
  c @{\hskip 0.1cm} %
  c@{\hskip 1pt}c @{\hskip 0.1cm} %
  c@{\hskip 1pt}c @{\hskip 0.1cm} %
  c@{\hskip 1pt}c @{\hskip 0.1cm} %
  c@{\hskip 1pt}c @{\hskip 0.1cm} %
  c@{\hskip 1pt}c             %
}

& \textbf{Input} & \multicolumn{2}{c}{\textbf{GT}} & \multicolumn{2}{c}{$\bm{\omega=0.0}$} & \multicolumn{2}{c}{$\bm{\omega=1.0}$} & \multicolumn{2}{c}{$\bm{\omega=2.0}$} & \multicolumn{2}{c}{$\bm{\omega=3.0}$} \\[0.1cm]

\vertlabel{Ours}
  & \framedimage{\basepath/input/\mixOne.png}
  & \framedimage{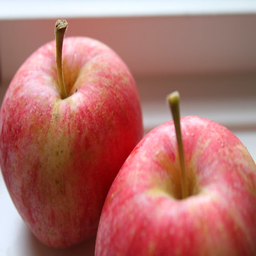}
  & \framedimage{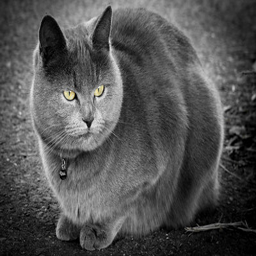}
  & \framedimage{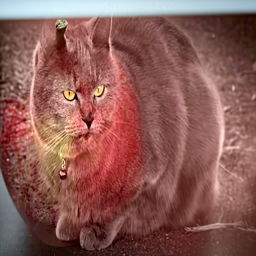}
  & \framedimage{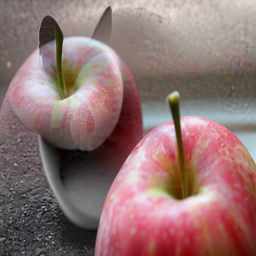}
  & \framedimage{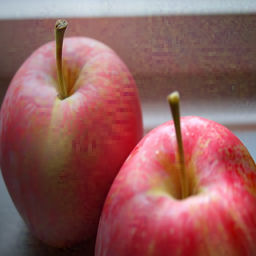}
  & \framedimage{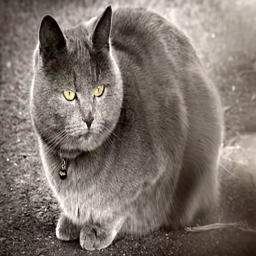}
  & \framedimage{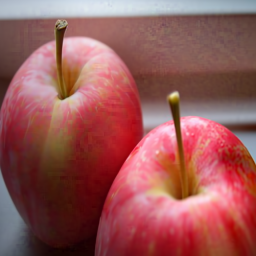}
  & \framedimage{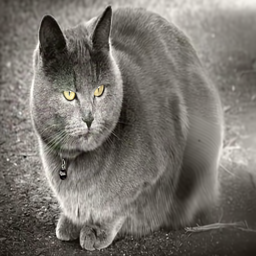}
  & \framedimage{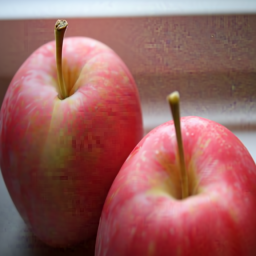}
  & \framedimage{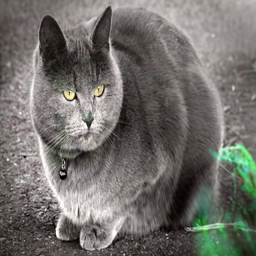} \\
  \noalign{\vspace{3pt}}

\vertlabel{\shortstack{Standard\\CFG}}
  & \framedimage{\basepath/input/\mixOne.png}
  & \framedimage{\basepath/ground_truth/\setOneA.png}
  & \framedimage{\basepath/ground_truth/\setOneB.png}
  & \framedimage{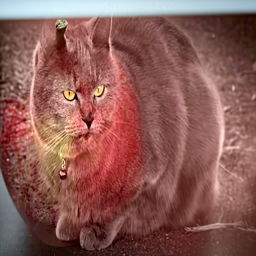}
  & \framedimage{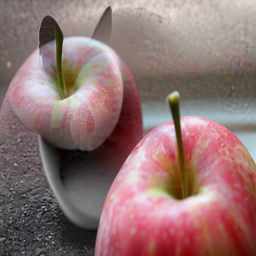}
  & \framedimage{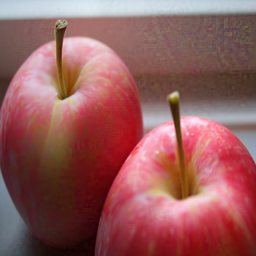}
  & \framedimage{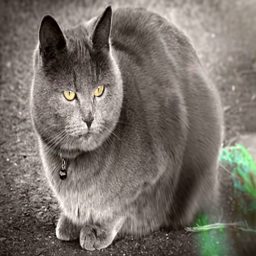}
  & \framedimage{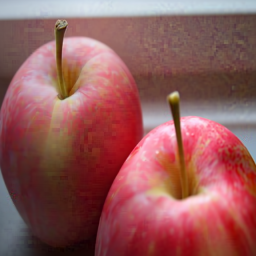}
  & \framedimage{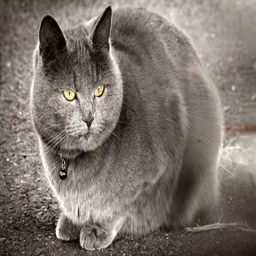}
  & \framedimage{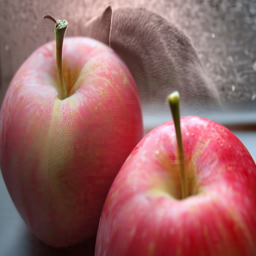}
  & \framedimage{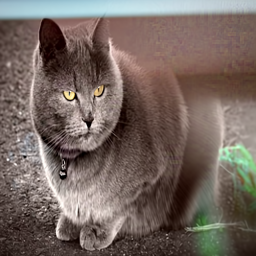} \\
  \noalign{\vspace{3pt}}

\vertlabel{\shortstack{Class-\\Level}}
  & \framedimage{\basepath/input/\mixOne.png}
  & \framedimage{\basepath/ground_truth/\setOneA.png}
  & \framedimage{\basepath/ground_truth/\setOneB.png}
  & \framedimage{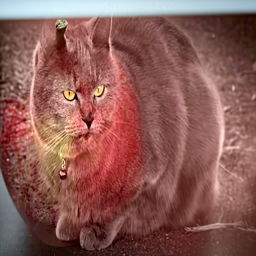}
  & \framedimage{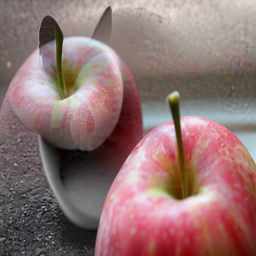}
  & \framedimage{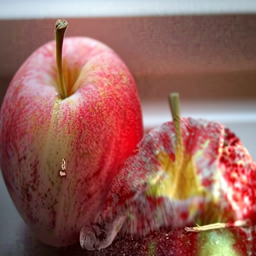}
  & \framedimage{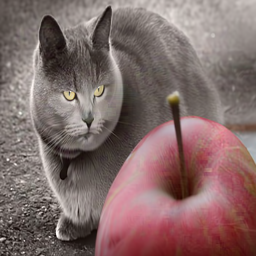}
  & \framedimage{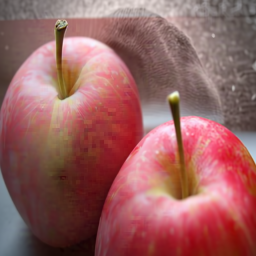}
  & \framedimage{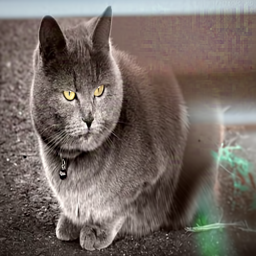}
  & \framedimage{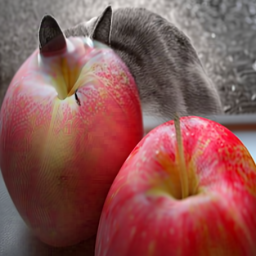}
  & \framedimage{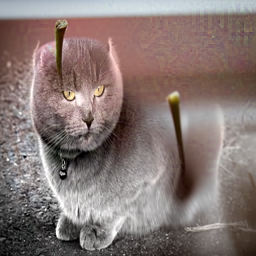} \\
  \noalign{\vspace{5pt}} \hline \noalign{\vspace{5pt}}

\vertlabel{Ours}
  & \framedimage{\basepath/input/\mixTwo.png}
  & \framedimage{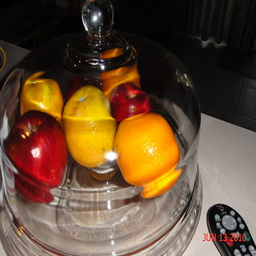}
  & \framedimage{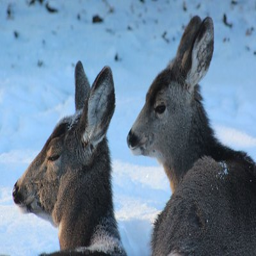}
  & \framedimage{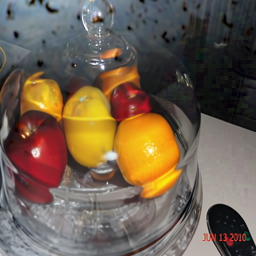}
  & \framedimage{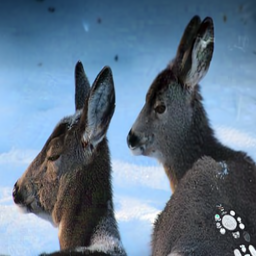}
  & \framedimage{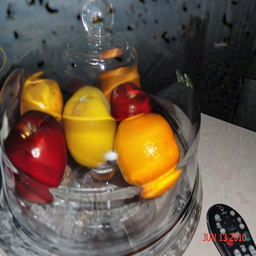}
  & \framedimage{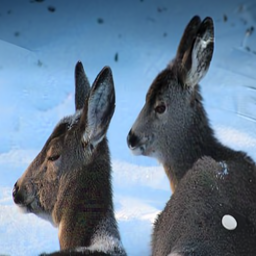}
  & \framedimage{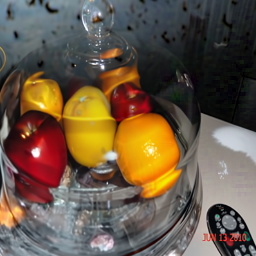}
  & \framedimage{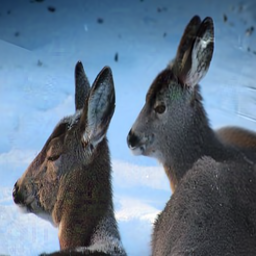}
  & \framedimage{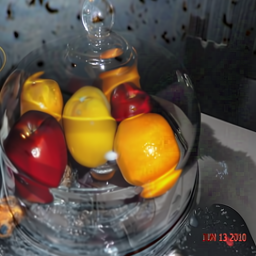}
  & \framedimage{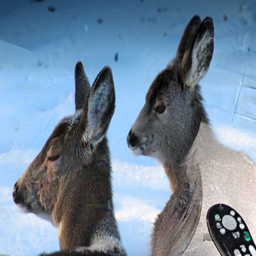} \\
  \noalign{\vspace{3pt}}

\vertlabel{\shortstack{Standard\\CFG}}
  & \framedimage{\basepath/input/\mixTwo.png}
  & \framedimage{\basepath/ground_truth/\setTwoA.png}
  & \framedimage{\basepath/ground_truth/\setTwoB.png}
  & \framedimage{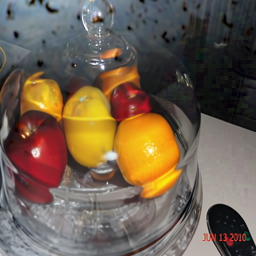}
  & \framedimage{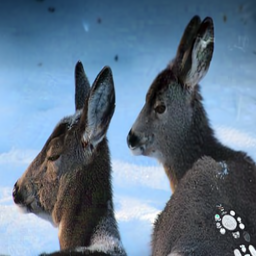}
  & \framedimage{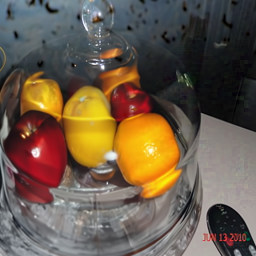}
  & \framedimage{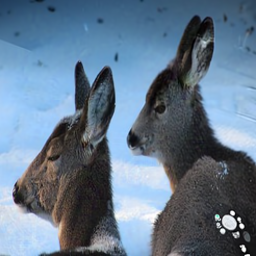}
  & \framedimage{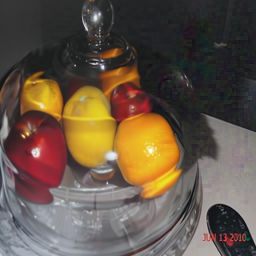}
  & \framedimage{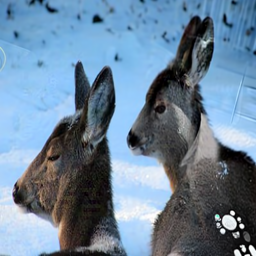}
  & \framedimage{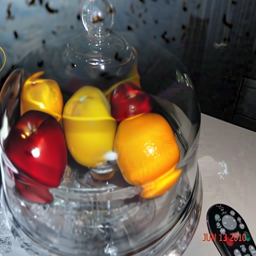}
  & \framedimage{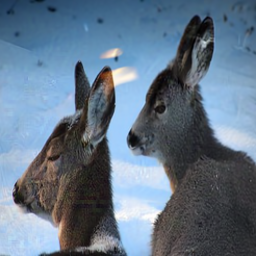} \\
  \noalign{\vspace{3pt}}

\vertlabel{\shortstack{Class-\\Level}}
  & \framedimage{\basepath/input/\mixTwo.png}
  & \framedimage{\basepath/ground_truth/\setTwoA.png}
  & \framedimage{\basepath/ground_truth/\setTwoB.png}
  & \framedimage{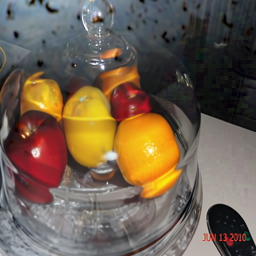}
  & \framedimage{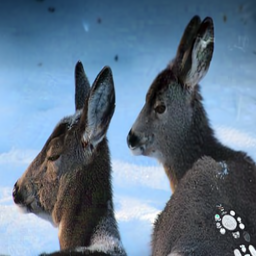}
  & \framedimage{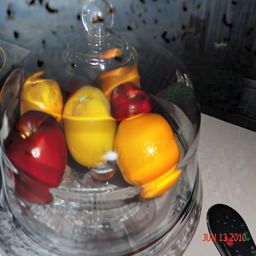}
  & \framedimage{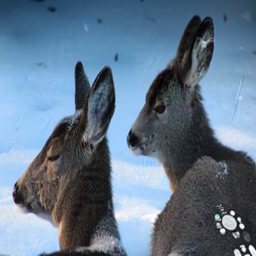}
  & \framedimage{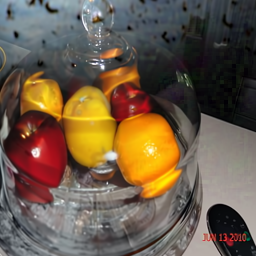}
  & \framedimage{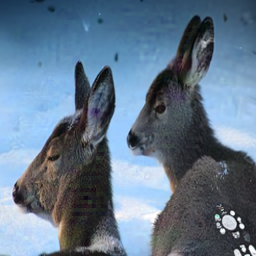}
  & \framedimage{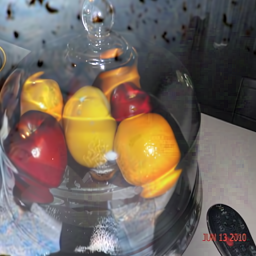}
  & \framedimage{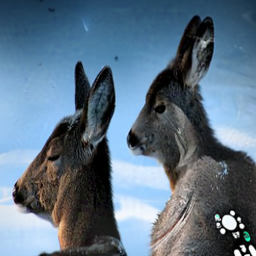} \\

\end{tabular}

\caption{%
\textbf{Qualitative comparison of text guidance methods.}
We compare three text guidance strategies across different guidance strengths ($\omega$) for two datasets: \textit{Apples--Cat} (top) and \textit{Fruits--Kangaroo} (bottom).
The methods are: \textbf{Ours} (cross-negative prompting: the other source's description is the negative prompt), \textbf{Standard CFG} (empty negative prompt), and \textbf{Class-Level} (cross-negative prompting with generic class labels instead of descriptions).
$\omega=0.0$ is prompt-free sampling and is identical across methods.
Ours is the top row of each block; quantitative results are in \cref{tab:text_guidance_compare}.
}
\label{fig:combined_text_guidance}
\end{figure*}

\subsubsection{Automated Prompt Generation}
\label{appendix:auto_prompts}

Our intended operating point permits lightweight user-provided layer descriptions, but this interaction can be inconvenient for fully automated deployment. We therefore test whether a local vision--language model (VLM) can supply the same semantic input from the blended observation alone. We use Qwen2-VL-7B-Instruct with greedy decoding and a fixed instruction that asks for exactly two one-sentence descriptions of the underlying photographs. The VLM receives only the blended image: it has no access to source images, annotations, filenames, ground truth, or manually written prompts. Its raw response and parsed descriptions are retained for audit.

We run the full pipeline on 80 spatially varying blends from the Fruits/Animals test set, held out from the pairs used to develop the VLM instruction. We compare three conditions on the same blends: \emph{None} uses $\omega=0$; \emph{Auto} uses the two VLM descriptions; and \emph{Manual} uses the human descriptions used elsewhere in the paper. Each condition generates three Stage~1 candidates with the known-scalar Stage~1 sampler ($T_s=80$, $t_s=64$) while estimating $\alpha$. Auto and Manual select a candidate with permutation-invariant PickScore; None has no prompts to score against and takes the first candidate. One Stage~2 refinement follows, with $\alpha$ kept fixed at its Stage~1 estimate. Metrics are computed as in \cref{tab:baseline_results}.

\begin{table}[t]
\centering
\caption{End-to-end automated-prompt study on 80 spatially varying blends with three Stage~1 candidates and one Stage~2 refinement. Auto prompts are generated by the VLM from the blended observation alone.}
\label{tab:auto_prompts_confirmation}
\small
\setlength{\tabcolsep}{5pt}
\begin{tabular}{lrrr}
\toprule
Prompts & PSNR $\uparrow$ & SSIM $\uparrow$ & LPIPS $\downarrow$ \\
\midrule
None ($\omega=0$) & 14.98 & 0.650 & 0.310 \\
Auto (Qwen2-VL) & 15.63 & 0.658 & 0.296 \\
Manual & \textbf{16.31} & \textbf{0.700} & \textbf{0.249} \\
\bottomrule
\end{tabular}
\end{table}

\begin{figure*}[t]
\centering
\small
\renewcommand{\arraystretch}{0.9}

\newcommand{\qualwidth}{0.12\textwidth}
\newcommand{\framedimage}[1]{\includegraphics[width=\qualwidth, valign=m, frame]{#1}}
\newcommand{\imgprompt}[2]{%
  \begin{tabular}[t]{@{}c@{}}
    \framedimage{#1} \\[2pt]
    \parbox{\qualwidth}{\centering\scriptsize\itshape ``#2''}
  \end{tabular}%
}
\newcommand{\vertlabel}[1]{\rotatebox[origin=c]{90}{\textbf{#1}}}
\newcommand{\gbase}{figures/0_appendix/auto_prompts/Gallery}

\begin{tabular}{
  c @{\hskip 0.1cm}
  c @{\hskip 0.2cm}
  c@{\hskip 0.15cm}c @{\hskip 0.3cm}
  c@{\hskip 0.15cm}c
}
 & \textbf{Input} & \multicolumn{2}{c}{\textbf{Auto (Qwen2-VL)}} & \multicolumn{2}{c}{\textbf{Manual}} \\[0.15cm]

\vertlabel{Representative}
  & \framedimage{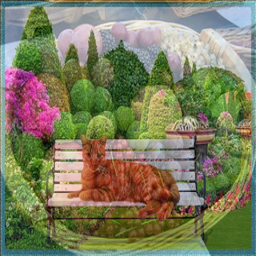}
  & \imgprompt{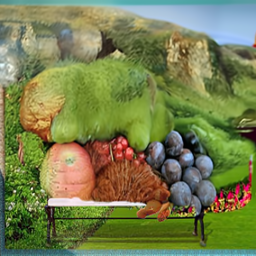}{A serene landscape with rolling hills and a clear blue sky.}
  & \imgprompt{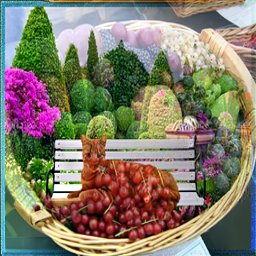}{A cat is sitting on a bench in a lush garden with vibrant flowers and greenery.}
  & \imgprompt{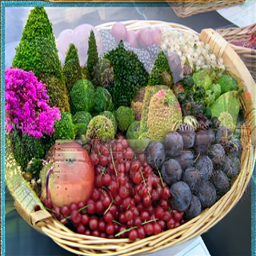}{A basket filled with colorful fruits}
  & \imgprompt{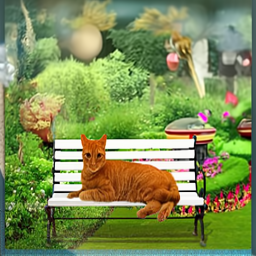}{A ginger cat on a white bench, trees in the background} \\
  \noalign{\vspace{6pt}}

\vertlabel{Failure}
  & \framedimage{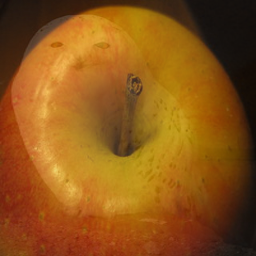}
  & \imgprompt{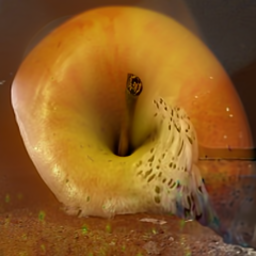}{A close-up of a ripe apple with a stem.}
  & \imgprompt{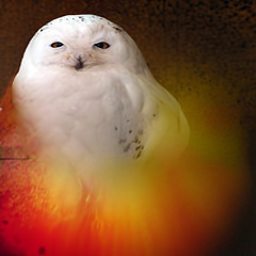}{A close-up of a person's face.}
  & \imgprompt{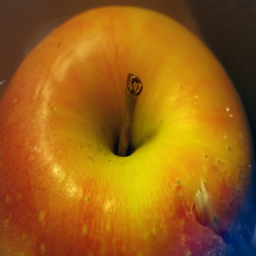}{A close-up photograph of red apple}
  & \imgprompt{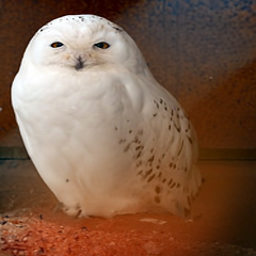}{A photograph of a white owl facing the camera} \\

\end{tabular}

\caption{Qualitative automated-prompt examples from the 80-pair end-to-end study. The VLM receives only the blended observation (\textbf{Input}) and writes both prompts shown under the \textbf{Auto} decompositions; \textbf{Manual} shows the same blend decomposed with the human-written descriptions used elsewhere in the paper. The two examples are the pairs closest to the median Auto--Manual PSNR gap (\textbf{Representative}: Auto 12.00\,dB vs.\ Manual 12.52\,dB) and to the 10th-percentile gap (\textbf{Failure}: Auto 16.10\,dB vs.\ Manual 20.55\,dB), selected by these pre-specified quantiles rather than by visual inspection. Both illustrate source misidentification: the VLM describes the fruit basket as a landscape and the owl as a person's face, and the corresponding Auto layers retain content from the other source.}
\label{fig:auto_prompts_qualitative}
\end{figure*}

The VLM produced two parseable descriptions for every input, and the pipeline runs unchanged with them, so a human is not required to provide semantic input. As shown in \cref{tab:auto_prompts_confirmation}, automatic prompts improve PSNR over prompt-free sampling by 0.65\,dB, about half of the 1.33\,dB gain obtained with manual prompts; their SSIM and LPIPS gains over prompt-free sampling are small. Automatic prompts remain below manual prompts on all three metrics. Failures mainly occur when the VLM misidentifies one source or describes a shared background instead of a constituent image (\cref{fig:auto_prompts_qualitative}), suggesting that better VLM prompting could narrow the remaining gap.

\subsubsection{Candidate Selection: CLIP vs. PickScore}
\label{appendix:selector_ablation}
The stochastic Stage~1 sampler produces three candidates. PickScore and CLIP ViT-B/32 cosine similarity are optional, ground-truth-free automated selectors, while in the intended interactive workflow a user may choose among the candidates. For the automated-selector comparison, both selectors use the same three Stage~1 candidates, score both possible assignments between the two outputs and the two prompts, and freeze their decision before Stage~2 metrics are loaded. We then evaluate the corresponding same-identifier Stage~2 refinement; all six settings have complete Stage~2 coverage for all three candidates.

\begin{table*}[t]
\centering
\scriptsize
\setlength{\tabcolsep}{5pt}
\caption{Stage~1 selector comparison on 300 seen (Fruits, Animals) and 100 unseen (Vehicles, Flowers) pairs per scenario. Metrics are computed from the matching same-identifier Stage~2 refinement and averaged over both recovered sources. Higher PSNR/SSIM and lower LPIPS are better.}
\begin{tabular}{lllccc}
\toprule
\textbf{Domain} & \textbf{Scenario} & \textbf{Selector} & \textbf{PSNR}$\uparrow$ & \textbf{SSIM}$\uparrow$ & \textbf{LPIPS}$\downarrow$ \\
\midrule
\multirow{6}{*}{Seen}
 & \multirow{2}{*}{Known scalar} & CLIP & 20.7634 & 0.8074 & 0.1493 \\
 & & \textbf{PickScore} & \textbf{21.3595} & \textbf{0.8134} & \textbf{0.1418} \\
 & \multirow{2}{*}{Unknown scalar} & CLIP & 19.1261 & 0.7440 & 0.2078 \\
 & & \textbf{PickScore} & \textbf{19.7825} & \textbf{0.7634} & \textbf{0.1902} \\
 & \multirow{2}{*}{Spatially varying} & CLIP & 17.5881 & 0.7189 & 0.2261 \\
 & & \textbf{PickScore} & \textbf{18.0521} & \textbf{0.7325} & \textbf{0.2118} \\
\midrule
\multirow{6}{*}{Unseen}
 & \multirow{2}{*}{Known scalar} & CLIP & 19.7174 & 0.7923 & 0.1406 \\
 & & \textbf{PickScore} & \textbf{20.3229} & \textbf{0.8115} & \textbf{0.1193} \\
 & \multirow{2}{*}{Unknown scalar} & CLIP & 18.8687 & 0.7577 & 0.1620 \\
 & & \textbf{PickScore} & \textbf{19.4802} & \textbf{0.7796} & \textbf{0.1403} \\
 & \multirow{2}{*}{Spatially varying} & CLIP & 17.3242 & 0.7350 & 0.1769 \\
 & & \textbf{PickScore} & \textbf{18.0788} & \textbf{0.7527} & \textbf{0.1620} \\
\bottomrule
\end{tabular}
\label{tab:selector_comparison}
\end{table*}

As reported in \cref{tab:selector_comparison}, PickScore improves the aggregate mean in all 18 setting--metric comparisons, with gains of $0.4640$--$0.7546$ dB PSNR, $0.0060$--$0.0219$ SSIM, and $0.0075$--$0.0217$ LPIPS. The selectors agree on only 41.0--54.7\% of inputs, indicating that the difference is not a rounding effect.

\subsubsection{Number of Stage~1 Candidates}
\label{appendix:candidate_budget}
The stochastic Stage~1 sampler exposes a direct quality--latency control through the number of candidates presented to the selector. We evaluate cumulative budgets $N_{\mathrm{cand}}\in\{1,2,3,4,6,8,12,16\}$ on 10 fixed pairs from each mixing scenario. For every input, we first generate an ordered pool of 16 candidates using fixed seeds. At each budget, PickScore selects from the corresponding prefix without access to ground truth. We refine every distinct selected candidate once in Stage~2 and reuse that refinement when it wins at multiple budgets. Because this study targets the candidate budget rather than $\alpha$ estimation, all three scenarios use the known-scalar Stage~1 sampler ($T_s=80$, $t_s=64$), Stage~2 keeps $\alpha$ fixed at its Stage~1 value, and in the unknown-scalar scenario the true scalar $\alpha$ is given to the method rather than estimated. Its absolute values are therefore not comparable to \cref{tab:baseline_results}. This yields 480 Stage~1 candidates and 102 distinct Stage~2 refinements; all expected outputs are present. We report means over the same 10 pairs at every budget.

\begin{table}[t]
\centering
\caption{Representative points from the candidate-budget study, averaged over 10 input pairs per scenario. Scalar rows use the given $\alpha$ (see text). Lower LPIPS is better; higher PSNR and SSIM are better.}
\label{tab:candidate_budget}
\small
\setlength{\tabcolsep}{4pt}
\begin{tabular}{lrrrr}
\toprule
Scenario & $N_{\mathrm{cand}}$ & PSNR & SSIM & LPIPS \\
\midrule
Known scalar & 1  & 21.08 & 0.769 & 0.236 \\
             & 3  & 21.57 & 0.802 & 0.198 \\
             & 16 & \textbf{22.97} & \textbf{0.829} & \textbf{0.180} \\
\midrule
Unknown scalar & 1  & 19.90 & 0.735 & 0.232 \\
               & 3  & 20.51 & 0.762 & 0.234 \\
               & 16 & \textbf{22.02} & \textbf{0.783} & \textbf{0.201} \\
\midrule
Spatially varying & 1  & 16.95 & 0.674 & 0.320 \\
                  & 3  & \textbf{17.31} & \textbf{0.683} & 0.310 \\
                  & 16 & 16.98 & 0.663 & \textbf{0.303} \\
\bottomrule
\end{tabular}
\end{table}

As shown in \cref{tab:candidate_budget}, increasing the budget from 1 to 16 improves the known-scalar setting by 1.90~dB PSNR and 0.060 SSIM while reducing LPIPS by 0.056. The unknown-scalar setting gains 2.12~dB and 0.049 SSIM while reducing LPIPS by 0.031. In contrast, the spatial endpoint changes are small (0.03~dB PSNR, $-0.011$ SSIM, and $-0.018$ LPIPS).

The trend is not monotonic in reconstruction quality because PickScore was trained for text-to-image preference rather than reference-based deblending fidelity. A newly available candidate can therefore receive a higher PickScore yet a worse PSNR, SSIM, or LPIPS. Only 2/10 known-scalar, 1/10 unknown-scalar, and 3/10 spatial inputs retain the same winner at budgets 3 and 16, and the 3-to-16 mean metric changes are mostly small at this sample size. The largest is unknown-scalar LPIPS, which improves by 0.033.

The measured Stage~1 cost is 8.07--8.38 minutes per candidate, while the single selected Stage~2 refinement costs 16.15--16.49 minutes. Thus, with this known-scalar Stage~1 sampler, the approximate sequential latency grows from 40--41 minutes at $N_{\mathrm{cand}}=3$ to 145--150 minutes at 16; the main settings, which also estimate $\alpha$, take longer (50.6--71.0 minutes at three candidates). Three candidates provide a practical default at substantially lower cost; a larger pool is an optional high-compute setting for scalar blends, rather than a universally superior choice.

\subsubsection{Best-of-\texorpdfstring{$N$}{N} Selection for the DoubleDIP Baseline}
\label{appendix:baseline_oversampling}

Our pipeline benefits from a stochastic Stage~1 sampler that produces several candidates, from which a ground-truth-free selector (CLIP or PickScore) picks the best (\cref{appendix:selector_ablation,appendix:candidate_budget}). The main comparison grants DoubleDIP the same best-of-three PickScore selection opportunity (\cref{tab:baseline_results}). Here, we quantify its effect and compare it with CLIP selection.

We rerun DoubleDIP with three samples per input, keeping all optimization settings identical ($8000$ iterations, learning rate $10^{-3}$, $256{\times}256$ resolution) and varying only the random seed across the three samples. We then score each of the three candidates with both CLIP ViT-B/32 and PickScore embeddings and choose the per-input best-of-three independently for each selector. The selector itself never sees ground-truth images, but unlike for our method it is not permutation-invariant: each candidate's two outputs were first ordered to match the ground-truth sources, and the score is the sum of the two prompt--image cosine similarities in that order. This ordering can only help DoubleDIP's selection. We evaluate on the same two benchmarks as the main experiments: the $300$-pair comparison set and the $100$-pair generalization set (unseen domains), across all three mixing scenarios.

\begin{table}[t]
\centering
\caption{Best-of-three selector comparison for the training-free DoubleDIP baseline. CLIP and PickScore independently select from the same three samples for each input; outputs are paired with prompts in ground-truth source order (see text). Bold marks the best value within each evaluation set and mixing scenario. Lower LPIPS is better; higher PSNR and SSIM are better.}
\label{tab:baseline_oversampling}
\small
\setlength{\tabcolsep}{4pt}
\begin{tabular}{llccc}
\toprule
\textbf{Scenario} & \textbf{Selector} & \textbf{PSNR}$\uparrow$ & \textbf{SSIM}$\uparrow$ & \textbf{LPIPS}$\downarrow$ \\
\midrule
\multicolumn{5}{l}{\textit{Comparison (300 pairs)}} \\
\midrule
\multirow{2}{*}{\makecell[l]{Known\\Scalar}}
                  & CLIP       & 14.42 & 0.515 & 0.537 \\
                  & PickScore  & \textbf{14.58} & \textbf{0.521} & \textbf{0.535} \\
\midrule
\multirow{2}{*}{\makecell[l]{Unknown\\Scalar}}
                  & CLIP       & 14.66 & 0.511 & \textbf{0.545} \\
                  & PickScore  & \textbf{14.69} & \textbf{0.516} & 0.552 \\
\midrule
\multirow{2}{*}{\makecell[l]{Spatially-\\Varying}}
                  & CLIP       & 13.68 & 0.457 & 0.597 \\
                  & PickScore  & \textbf{13.71} & \textbf{0.466} & \textbf{0.594} \\
\midrule
\multicolumn{5}{l}{\textit{Generalization (100 pairs, unseen domains)}} \\
\midrule
\multirow{2}{*}{\makecell[l]{Known\\Scalar}}
                  & CLIP       & \textbf{13.76} & 0.457 & \textbf{0.511} \\
                  & PickScore  & 13.70 & \textbf{0.459} & 0.518 \\
\midrule
\multirow{2}{*}{\makecell[l]{Unknown\\Scalar}}
                  & CLIP       & 13.88 & 0.447 & \textbf{0.524} \\
                  & PickScore  & \textbf{13.97} & \textbf{0.453} & 0.531 \\
\midrule
\multirow{2}{*}{\makecell[l]{Spatially-\\Varying}}
                  & CLIP       & 13.01 & \textbf{0.397} & \textbf{0.570} \\
                  & PickScore  & \textbf{13.05} & \textbf{0.397} & 0.576 \\
\bottomrule
\end{tabular}
\end{table}

As shown in \cref{tab:baseline_oversampling}, the two selectors behave comparably, with PickScore slightly ahead on PSNR and SSIM in most settings.

Crucially, the choice of selector does not change the qualitative picture: even with the same selection opportunity, DoubleDIP remains in the $13$--$15$~dB PSNR range, far below our method (\cref{tab:baseline_results}) and its own candidate-budget curve (\cref{tab:candidate_budget}). The separation quality gap therefore stems from the strength of the underlying prior rather than from the selection rule.

\subsubsection{Number of Recurrence Iterations}
\label{appendix:recurrence_ablation}
We sweep Stage~1 recurrence count $n_1\in\{4,6,8,10\}$ and Stage~2 count $n_2\in\{2,4,6,8,10\}$ on the same 20 spatially varying blends used for the timestep ablation. Performance generally improves with recurrence count (\cref{tab:recurrence_ablation}), but the gains saturate. At $n_1=8$, increasing $n_2$ from 8 to 10 improves PSNR by only 0.13\,dB, SSIM by 0.005, and LPIPS by 0.002, while increasing recorded Stage~2 sampling time from 17.3 to 20.8 minutes. Increasing both counts to 10 slightly improves SSIM and LPIPS but reduces PSNR relative to $(8,10)$. We therefore use eight recurrences in both stages as a quality--runtime compromise.

\begin{table*}[t]
    \centering
    \caption{Recurrence-count ablation on 20 spatially varying Fruits/Animals blends with $N_{\mathrm{cand}}=1$. Rows vary Stage~1 recurrences $n_1$ and columns vary Stage~2 recurrences $n_2$; each cell reports PSNR$\uparrow$/SSIM$\uparrow$/LPIPS$\downarrow$. Bold denotes the best value and ties in the displayed digits are bold in every tied cell, and underlining marks the adopted setting $(n_1,n_2)=(8,8)$. Metrics are recomputed from saved outputs at $256\times256$ after permutation-invariant source matching.}
    \label{tab:recurrence_ablation}
    \scriptsize
    \setlength{\tabcolsep}{5pt}
    \begin{tabular}{c@{\hspace{6pt}}ccccc}
        \toprule
        $n_1\backslash n_2$ & 2 & 4 & 6 & 8 & 10 \\
        \midrule
        4  & 14.520/.565/.357 & 14.923/.597/.335 & 15.131/.631/.312 & 15.426/.637/.312 & 15.529/.651/.290 \\
        6  & 14.816/.587/.325 & 15.252/.618/.301 & 15.515/.647/.279 & 15.765/.658/.267 & 15.794/.661/.266 \\
        8  & 15.560/.625/.295 & 16.000/.656/.270 & 16.282/.671/.259 & \underline{16.737/.690/.241} & \textbf{16.868}/\textbf{.695}/.239 \\
        10 & 15.326/.631/.280 & 15.705/.659/.259 & 15.964/.674/.250 & 16.259/.689/.236 & 16.445/\textbf{.695/.232} \\
        \bottomrule
    \end{tabular}
\end{table*}

\subsubsection{Diffusion Timestep Selection}
\label{appendix:timestep_ablation}

We leverage the CCDF principle \cite{chung2022come}, which enables structure preserving by forward-diffusing to an intermediate timestep $t < T$ before reversing. To determine the optimal starting timesteps for both stages, we conduct a grid search ablation.
 
We evaluate five starting timesteps for Stage 1 (with maximum timestep $T_s=100$): $t_s \in \{20, 40, 60, 80, 100\}$, and four starting timesteps for Stage 2 (with $T_r=600$): $t_r \in \{50, 200, 350, 500\}$. This results in 20 configurations tested on a subset of 20 pairs from the \textit{Fruits vs Animals} dataset with spatially-varying $\alpha$-maps and $N_{cand}=1$.

\begin{figure*}[t]
    \centering
    \begin{subfigure}[b]{0.42\textwidth}
        \centering
        \includegraphics[width=\textwidth]{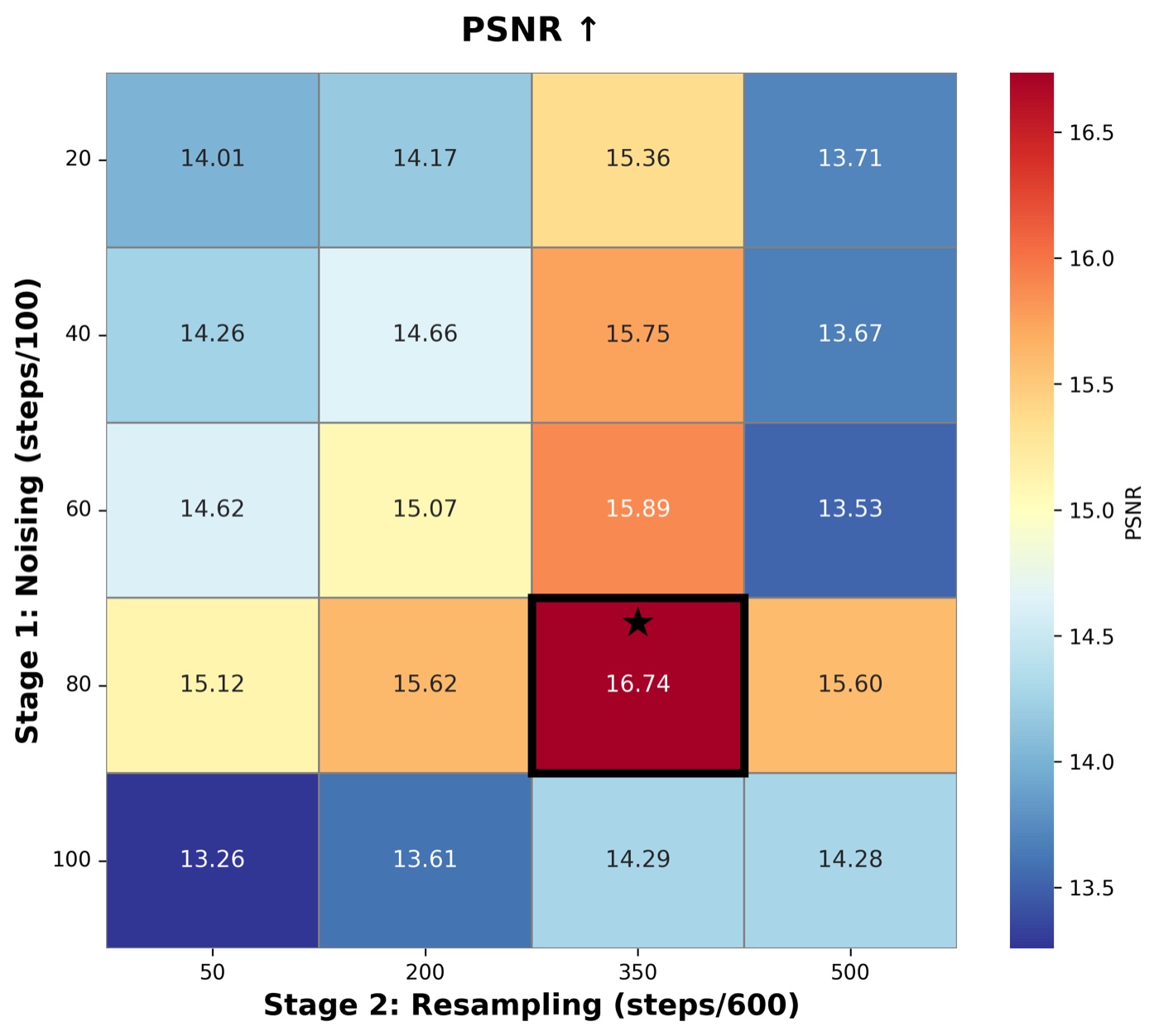}
        \caption{PSNR}
        \label{fig:heatmap_psnr}
    \end{subfigure}%
    \hfill
    \begin{subfigure}[b]{0.42\textwidth}
        \centering
        \includegraphics[width=\textwidth]{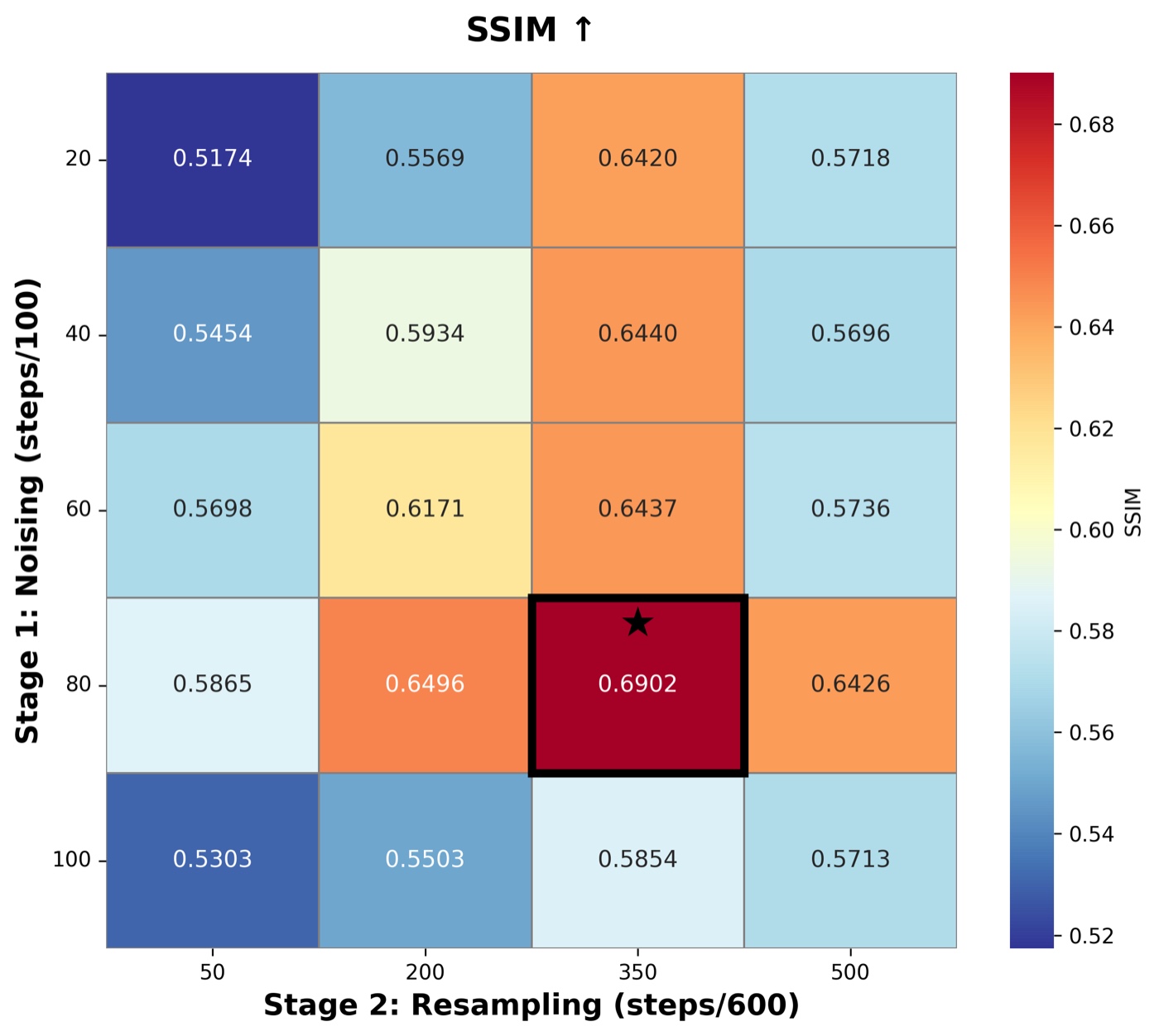}
        \caption{SSIM}
        \label{fig:heatmap_ssim}
    \end{subfigure}

    \vspace{0.15cm}

    \begin{subfigure}[b]{0.42\textwidth}
        \centering
        \includegraphics[width=\textwidth]{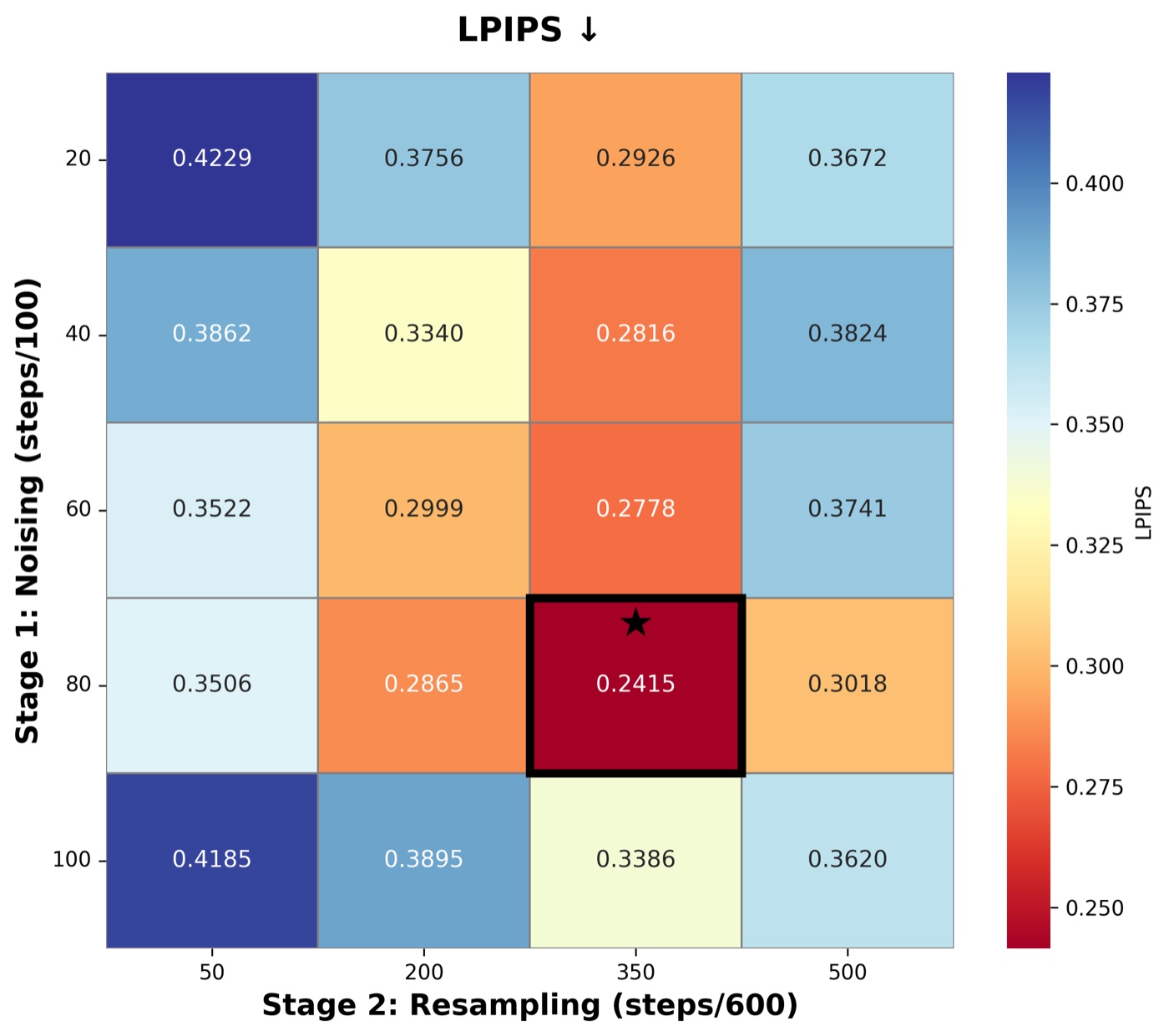}
        \caption{LPIPS}
        \label{fig:heatmap_lpips}
    \end{subfigure}
    
    \caption{Performance across Stage~1 starting timestep $t_s$ (rows, out of $T_s=100$) and Stage~2 starting timestep $t_r$ (columns, out of $T_r=600$) on 20 spatially varying Fruits/Animals blends with $N_{\mathrm{cand}}=1$. The starred box marks the adopted setting $(t_s,t_r)=(80,350)$. Each panel has its own colour scale.}
    \label{fig:sampling_steps_ablation}
\end{figure*}

The results in \cref{fig:sampling_steps_ablation} show that increasing the Stage~1 starting timestep improves all three metrics from $t_s=20$ through $t_s=80$, but performance degrades at the fully noised setting $t_s=100$. For Stage~2, $t_r=350$ gives the best PSNR, SSIM, and LPIPS for every tested Stage~1 setting, while increasing to $t_r=500$ degrades performance. We therefore use the intermediate settings $t_s=80$ and $t_r=350$, which provide enough noise for separation and refinement without the drift observed at the largest values.

\subsubsection{Necessity of Two-Stage Refinement}
\label{appendix:two_stage_ablation}
To validate the necessity of the two stages, we compare decomposition with and without the pixel-space refinement stage on 300 inputs from each mixing scenario: \textit{Known Scalar}, \textit{Unknown Scalar}, and \textit{Spatially-Varying} (\cref{subsec:exp_setup}). For each input, the Stage 1 candidate selected by PickScore is compared with its corresponding same-identifier Stage 2 refinement. Since Stage 1 produces outputs at 512x512 while Stage 2 operates at 256x256, we downsample Stage 1 outputs before evaluation.

\begin{table}[H]
\centering
\small
\caption{Effect of Stage~2 refinement on 300 Fruits/Animals inputs per mixing scenario. Stage~1 is the PickScore-selected candidate (downsampled to $256\times256$) and Stage~2 its refinement. Metrics are averaged over both recovered sources.}
\begin{tabular}{lcccc}
\toprule
\textbf{Scenario} & \textbf{Stage} & \textbf{PSNR}$\uparrow$ & \textbf{SSIM}$\uparrow$ & \textbf{LPIPS}$\downarrow$ \\
\midrule
\makecell[l]{Known\\Scalar} & 1 & 17.3872 & 0.5910 & 0.3175 \\
               & 2 & \textbf{21.3595} & \textbf{0.8134} & \textbf{0.1418} \\
\makecell[l]{Unknown\\Scalar} & 1 & 16.7536 & 0.5668 & 0.3450 \\
               & 2 & \textbf{19.7825} & \textbf{0.7634} & \textbf{0.1902} \\
\makecell[l]{Spatially-\\Varying} & 1 & 15.6474 & 0.5453 & 0.3634 \\
               & 2 & \textbf{18.0521} & \textbf{0.7325} & \textbf{0.2118} \\
\bottomrule
\end{tabular}
\label{tab:two_stage_ablation}
\end{table}

As shown in \cref{tab:two_stage_ablation}, pixel-space refinement substantially improves PSNR, SSIM, and LPIPS in all three scenarios. The gains are largest in the known-scalar setting, but remain substantial for unknown-scalar and spatially-varying blends, supporting the role of Stage 2 in restoring fidelity and perceptual quality beyond the coarse latent-space decomposition. Qualitative comparisons can be found in \cref{fig:two_stage_comparison}.

\begin{figure*}[!b] %
\centering
\footnotesize
\renewcommand{\arraystretch}{0.5}

\newcommand{\basepath}{figures/0_appendix/two_stage/Gallery}
\newcommand{\setOneA}{fruits}
\newcommand{\setOneB}{chicken}
\newcommand{\mixOne}{\setOneA\_\setOneB}
\newcommand{\setTwoA}{market}
\newcommand{\setTwoB}{bull}
\newcommand{\mixTwo}{\setTwoA\_\setTwoB}

\newcommand{\framedimage}[1]{\includegraphics[width=0.125\textwidth, valign=m, frame]{#1}}

\newcommand{\doubleimage}[2]{%
  \setlength{\fboxsep}{0pt}\setlength{\fboxrule}{0.5pt}%
  \fbox{\includegraphics[width=0.125\textwidth, valign=m]{#1}}%
  \hspace{1pt}%
  \fbox{\includegraphics[width=0.125\textwidth, valign=m]{#2}}%
}

\newcommand{\vertlabel}[1]{\rotatebox[origin=c]{90}{\textbf{#1}}}

\begin{tabular}{
  c @{\hskip 0.1cm}   %
  c @{\hskip 0.2cm}   %
  c @{\hskip 0.2cm}   %
  c @{\hskip 0.2cm}   %
  c                   %
}

 & \textbf{Input} & \textbf{Stage 1} & \textbf{Stage 2} & \textbf{Ground Truth} \\[0.1cm]

\vertlabel{\shortstack{Fruits\\Chicken}}
  & \framedimage{\basepath/input/\mixOne.png}
  & \doubleimage{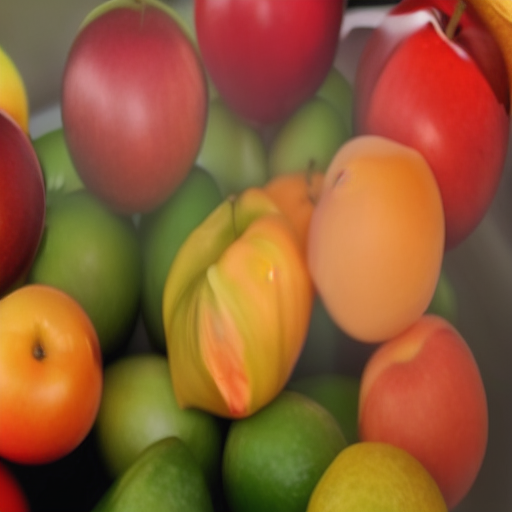}{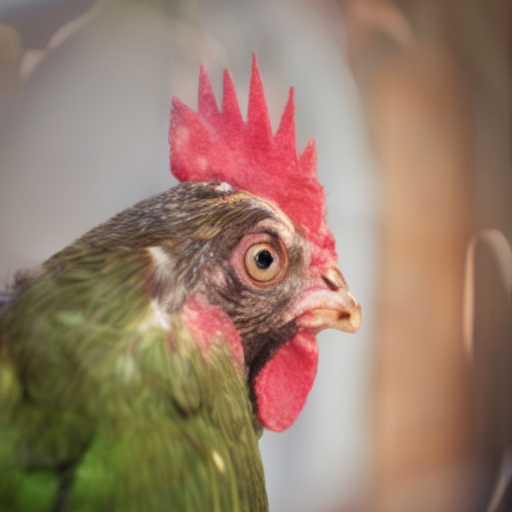}
  & \doubleimage{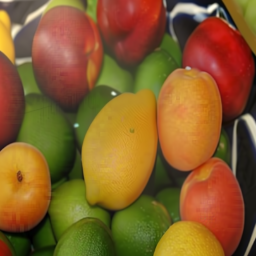}{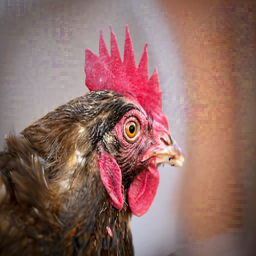}
  & \doubleimage{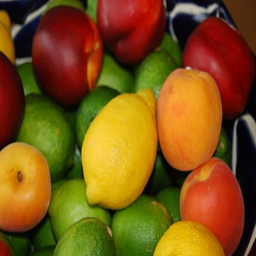}{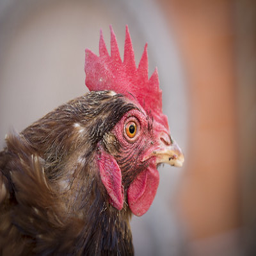} \\
  
  \noalign{\vspace{3pt}} \hline \noalign{\vspace{3pt}} %

\vertlabel{\shortstack{Market\\Bull}}
  & \framedimage{\basepath/input/\mixTwo.png}
  & \doubleimage{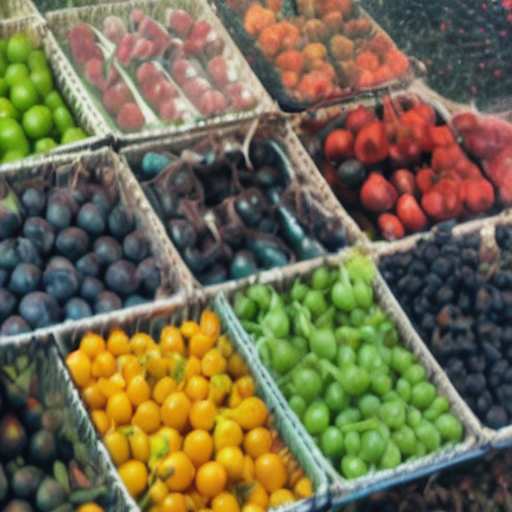}{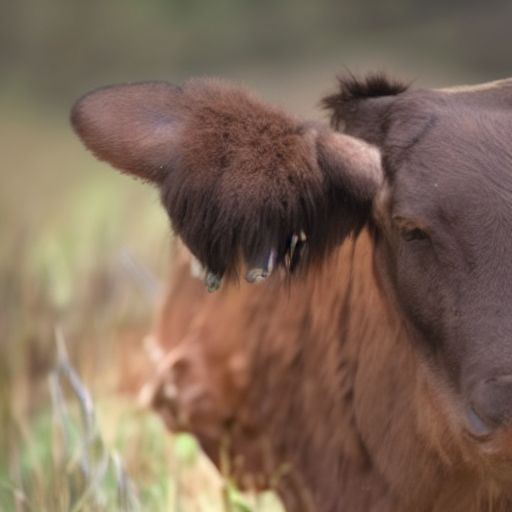}
  & \doubleimage{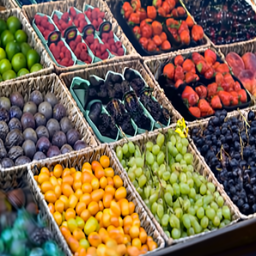}{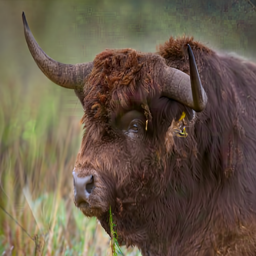}
  & \doubleimage{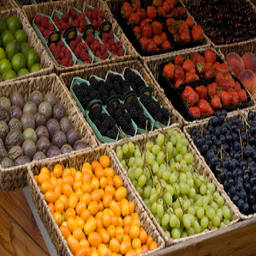}{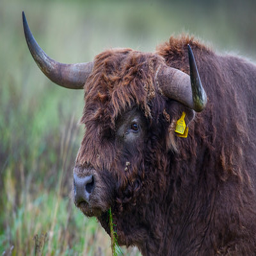} \\

\end{tabular}

\caption{\textbf{Qualitative results of the two-stage pipeline.} Stage~1 performs coarse semantic separation; Stage~2 restores fine details and better matches the ground truth.}
\label{fig:two_stage_comparison}
\end{figure*}

\subsubsection{Alpha Maps Outside the Polynomial Family}
\label{appendix:alpha_out_of_family}
The spatially varying test blends in \cref{tab:baseline_results} are drawn from the same second-order polynomial family as our $\alpha$ estimator. To test this assumption, we blend 20 Fruits/Animals pairs with two smooth matte families that a quadratic cannot represent exactly: a \emph{Gaussian} blob with random center, width, and amplitude, and a \emph{reflection-like} diagonal band with a radial vignette. We compare our quadratic estimator with a more expressive low-frequency alternative: a sigmoid-bounded $4\times4$ control grid upsampled bicubically to full resolution (16 parameters). Because of computational cost, this study runs Stage~1 only, with one candidate per input and the Stage~1 sampler of the known-scalar setting ($T_s=80$, $t_s=64$); absolute values are therefore not comparable to \cref{tab:baseline_results}.

\begin{table}[t]
\centering
\caption{Stage~1 results (one candidate per input) on 20 Fruits/Animals pairs per ground-truth matte family. Bold marks the better estimator for each ground-truth matte; the paired 95\% confidence intervals of the PSNR differences include zero. Alpha MAE is the mean absolute error between the recovered and ground-truth $\alpha$-maps.}
\label{tab:alpha_out_of_family}
\small
\setlength{\tabcolsep}{3pt}
\begin{tabular}{llrrrr}
\toprule
GT matte & Estimator & PSNR$\uparrow$ & SSIM$\uparrow$ & LPIPS$\downarrow$ & $\alpha$ MAE$\downarrow$ \\
\midrule
Polynomial & Quadratic & 14.18 & 0.504 & 0.586 & 0.180 \\
           & Grid $4\times4$ & \textbf{14.52} & \textbf{0.518} & \textbf{0.565} & \textbf{0.142} \\
\midrule
Gaussian & Quadratic & 13.91 & 0.500 & \textbf{0.599} & 0.265 \\
         & Grid $4\times4$ & \textbf{14.11} & \textbf{0.507} & 0.603 & \textbf{0.237} \\
\midrule
Reflection-like & Quadratic & 13.52 & 0.482 & 0.625 & 0.242 \\
                & Grid $4\times4$ & \textbf{13.66} & \textbf{0.496} & \textbf{0.616} & \textbf{0.216} \\
\bottomrule
\end{tabular}
\end{table}

As shown in \cref{tab:alpha_out_of_family}, the quadratic estimator degrades gradually outside its own family: PSNR drops by 0.27--0.66\,dB and $\alpha$ MAE increases from 0.180 to 0.242--0.265. The control grid lowers $\alpha$ MAE in all three families, but its source-image gains are small (+0.14 to +0.33\,dB PSNR), and their paired 95\% confidence intervals include zero at this sample size. We therefore keep the quadratic estimator, and regard more expressive low-frequency $\alpha$ representations as a promising extension for mattes far from the polynomial family.

\fi

\end{document}